\documentclass[12pt,english]{article}
\usepackage[T1]{fontenc}

\usepackage{geometry}
\usepackage{amssymb}
\usepackage{babel}
\usepackage{caption}
\usepackage{hyperref}

\usepackage[utf8]{inputenc}

\usepackage{amsthm}
\usepackage{amssymb}

\newcommand{\Halmos}{$\blacksquare$}

\usepackage{hyperref}
\usepackage[capitalise]{cleveref}

\usepackage[normalem]{ulem}

\newtheorem{lemma}{Lemma}

\newtheorem{theorem}{Theorem}

\newtheorem{proposition}{Proposition}

\Crefname{table}{Table}{Tables}

\crefname{lemma}{Lemma}{Lemmas}
\crefname{theorem}{Theorem}{Theorems}
\crefname{proposition}{Proposition}{Propositions}
\usepackage{endnotes}
\let\footnote=\endnote

\usepackage{natbib}
 \bibpunct[, ]{(}{)}{,}{a}{}{,}%

\newcommand{\w}{w}
\newcommand{\z}{z}
\newcommand{\dimension}{d}

\newcommand{\sigmatilde}{\tilde{\sigma}}
\newcommand\numberthis{\addtocounter{equation}{1}\tag{\theequation}}

\usepackage{algorithm}
\usepackage{algorithmic}
\usepackage{graphicx}
\usepackage{caption}
\usepackage{mathptmx}
\usepackage{amsmath}
\usepackage{amssymb}

\usepackage{subcaption}
\usepackage{breqn}

\usepackage{bibunits}
\defaultbibliographystyle{natbib}
\defaultbibliography{or}

\date{}

\begin{document}
\begin{bibunit}

\title{Bayesian Optimization with Expensive Integrands}

\author{Saul Toscano-Palmerin, Peter I. Frazier\\
st684@cornell.edu, pf98@cornell.edu\\
School of Operations Research \& Information Engineering\\
Cornell University, Ithaca, NY 14853\\
}

\maketitle

\begin{abstract} 
We propose a Bayesian optimization algorithm for objective functions that are 
% \pfedit{expectations of expensive-to-evaluate functions with respect to a known density or probability mass functions.}
sums or integrals of expensive-to-evaluate functions, allowing noisy evaluations.
These objective functions arise in multi-task Bayesian optimization for tuning machine learning hyperparameters, optimization via simulation, and sequential design of experiments with random environmental conditions. 
Our method is average-case optimal by construction when a single evaluation of the integrand remains within our evaluation budget.  Achieving this one-step optimality requires solving a challenging value of information optimization problem, for which we provide a novel efficient discretization-free computational method.  We also provide consistency proofs for our method in both continuum and discrete finite domains for objective functions that are sums.
In numerical experiments comparing against previous state-of-the-art methods, including those that also leverage sum or integral structure, our method performs as well or better across a wide range of problems and offers
significant improvements when evaluations are noisy or the integrand varies smoothly in the integrated variables.
\end{abstract}

\section{Introduction}

We consider two closely-related derivative-free black-box global optimization problems with expensive-to-evaluate objective functions,
\begin{equation}
\max_{x\in A\subset\mathbb{R}^{\dimension}} G(x) := \max_{x\in A\subset\mathbb{R}^{\dimension}}\sum_{w=1}^{n}F\left(x,w\right)p(w)
\label{eq:goal1}
\end{equation}
and
\begin{equation}
\max_{x\in A\subset\mathbb{R}^{\dimension}} G(x) := \max_{x\in A\subset\mathbb{R}^{\dimension}}\int_{}F\left(x,w\right)p(w)dw,
\label{eq:goal2}
\end{equation}
where $A$ is a simple compact set (e.g., a hyperrectangle, simplex, or finite collection of points); $w$ is a vector belonging to a set $W$; $p$ is finite and inexpensive to evaluate with a known analytic form; and $F$ is expensive to evaluate, does not provide derivatives with its evaluations, and may be observable either directly or with independent normally distributed noise. We also assume in \eqref{eq:goal2} that $F(x,\cdot)p(\cdot)$ is integrable for each $x$.
Here, ``expensive-to-evaluate'' functions are ones that consume a great deal of time per evaluation, e.g., minutes or hours each, or whose number is otherwise severely restricted \citep[see, e.g.,][]{sacks1989, booker1998}.
We treat $F$ as a black box and assume that it is continuous in $x$ and also in $w$ in problem \eqref{eq:goal2} and well-represented by a Gaussian process prior \citep{RaWi06} as described below, but make no other assumptions on its structure.

This pair of closely related problems arises in three settings:
\begin{enumerate}

\item First, both \eqref{eq:goal1} and \eqref{eq:goal2} arise when optimizing average-case performance of an engineering system or business process across environmental conditions,
where $F\left(x,\w\right)$ is the performance of system design $x$ under environmental condition $\w$, and $p(\w)$ represents the fraction of time that condition $\w$ occurs.
This arises, for example, when choosing 
the shape of an aircraft's wing \citep{MLA},
the configuration of a cardiovascular bypass graft \citep{marsden2010},
or the parameters of an algorithm that dispatches cars in a ride-sharing service.

\item Second, both \eqref{eq:goal1} and \eqref{eq:goal2} arise when we wish to optimize the expected value of a system modeled by a discrete-event simulation $f(x,\omega)$, where $\omega$ is random. In this setting, we may choose a random variable $\w$ whose distribution $p(\w)$ we know, and for which we can simulate $\omega$ given $\w$.  We may then define $F\left(x,\w\right) = E[f(x,\omega) | \w]$. Our objective $E[f(x,\omega)]$ becomes either $\sum_{\w=1}^n F(x,\w) p(w)$ if $\w$ is discrete or $\int F(x,\w) p(\w) d\w$ if $\w$ is continuous, and we can obtain noisy observations of $F(x,\w)$ by simulating $f(x,\omega)$ with $\omega$ drawn from its conditional distribution given $\w$. This arises, for example, when building a transportation system to maximize expected service quality subject to stochastic patterns of arrivals through the day $\omega$, which we can simulate given the total number of arrivals in a day $\w$. When used for variance reduction in simulation rather than optimization, this technique is known as stratification \citep{glasserman2003monte}.

\item Third, \eqref{eq:goal1} arises when tuning
a machine learning algorithm's hyperparameters by using 
k-fold cross-validation.  In this application, 
$F\left(x,\w\right)$ is the error on fold $\w$ using hyperparameters $x$ and our goal is to minimize $\sum_\w F\left(x,\w\right)$.  This problem arises more generally when optimizing average performance across multiple prediction tasks  \citep{bardenet2013collaborative,hutter2011sequential,swersky2013multi} and is called multi-task Bayesian optimization   \citep{swersky2013multi}.
%and the function $F(x,w)$ for $w$ fixed is called a task.  

% One prominent example is minimizing cross-validation error in k-fold cross validation \citep{swersky2013multi}, where $F\left(x,\w\right)$ is the error on fold $\w$ performed using hyperparameters $x$.  This class of applications is called multi-task Bayesian optimization \citep{swersky2013multi} and the function $F(x,w)$ for $w$ fixed is called a task.
\end{enumerate}

% Because $F$ is a black-box derivative-free expensive function, so is $G$.  
% The expensive derivative-free black box nature of $F$ creates similar properties for $G$: when evaluated using a sum, numerical quadrature, or Monte Carlo, it is an expensive-to-evaluate derivative-free black box function that may be noisy.  

Although \eqref{eq:goal1} and \eqref{eq:goal2} can be considered jointly as maximization of an objective
$\int F\left(x,w\right)d\mu\left(w\right)$ where $\mu$ is a measure,
and our theoretical analysis will at times take this view, \eqref{eq:goal1} and \eqref{eq:goal2} have very different properties computationally and have been considered separately in the literature so we refer to them separately here. 

\paragraph{Potential Solution Approaches:}
Problems \eqref{eq:goal1} and \eqref{eq:goal2} may be solved by optimizing $G(x)$ directly with a method designed for derivative-free black box global optimization of expensive and possibly noisy functions.  These methods include Bayesian optimization methods \citep{jones1998efficient, forrester2008engineering,brochu2010tutorial} and other surrogate-based optimization methods \citep{barthelemy1993,torczon1997, shoemaker2014}. 
Indeed, $G(x)$ can be evaluated using multiple evaluations of $F(x,\w)$ by summing in \eqref{eq:goal1} or with numerical quadrature in \eqref{eq:goal2}. However, the expense of evaluating $G(x)$ is many times larger than for $F(x,\w)$, especially if $n$ in \eqref{eq:goal1} is large or numerical quadrature in \eqref{eq:goal2} is performed accurately.
This approach is inefficient because it is unable to adjust the computational effort spent on evaluating $G(x)$: it either evaluates it fully or not at all. This inefficiency is most apparent when the first evaluations of $F(x,\w)$ at $x$ indicate $G(x)$ is substantially sub-optimal: the extra expense of fully evaluating $G(x)$ is wasted.
% shorter version of previous 2 sentences
%This approach is inefficient because the first few evaluations of $F(x,\w)$ at some $x$ may already indicate $G(x)$ is substantially below optimal: the extra expense of a full evaluation of $G(x)$ is wasted.

These problems may also be solved by applying a black box global optimization method to noisy observations of $G(x)$ obtained via Monte Carlo sampling. One may sample $w_{1},\ldots,w_{m}$ from $p$ and use $\frac{1}{m}\sum_{i=1}^{m}F(x,w_{i})$ as a noisy estimate of $G(x)$. This approach is inefficient because it ignores information about $\w$ when building its surrogate for $G$.  This inefficiency is most apparent when the first two evaluations of $F(x,\w)$ are at the same or very similar $\w$. If $F$ is free from noise and varies slowly with $\w$, the second such evaluation provides little information beyond the first.
This inefficiency could perhaps be mitigated by using Quasi Monte Carlo  \citep[QMC; see, e.g.,][]{glasserman2003monte}  together with an optimization method (ideally a multifidelity one, e.g., \citealt{forrester2007multi}) that is tolerant to bias in its noisy observations, but even this would become inefficient when most of $G(x)$'s variability is driven by values of $\w$ not sampled until later in the QMC sequence.

These inefficiencies in optimization based on surrogate models of $G$ suggest one may create a more efficient method through surrogate models of $F$, coupled with intelligent selection of points $x$ and $w$ at which to evaluate $F$.  
\citet{williams2000sequential} and \citet{swersky2013multi}
developed methods of this type for solving \eqref{eq:goal1},
and \citet{groot2010bayesian,Xie:2012} for solving \eqref{eq:goal2}. 
While these approaches can improve performance over modeling $G$ alone, we show in this article that they leave substantial room for improvement.
Indeed, all of these previous approaches except \citet{Xie:2012} use heuristic two-step rules that choose $x$ first without considering $\w$, and then choose $\w$ with $x$ fixed.  As we show below, not considering $x$ and $\w$ together causes these methods to perform poorly in certain settings, even sometimes failing to be asymptotically consistent.
Moreover, these previous methods are insufficiently general: \cite{Xie:2012} requires $p$ and the kernel of the covariance of the Gaussian process to be Gaussian, and all previous methods require evaluations of $F$ to be free from noise, significantly restricting their applicability. 

\paragraph{Contributions:}
In this paper, we significantly generalize and improve over this previous work by developing a novel method, Bayesian Quadrature Optimization (BQO), that uses a one-step value of information analysis to select the pair of points $x,\w$ at which to evaluate $F$.
This method is general and supports solving
either \eqref{eq:goal1} or \eqref{eq:goal2} with noisy or noise-free evaluations of $F$, and this support for noisy observations significantly expands the applicability of our approach within optimization via simulation.
This algorithm is Bayes-optimal by construction when only a single evaluation of $F$ may be made. We also prove that it provides a consistent estimator of the global optimum of $G$ as the number of samples allowed extends to infinity in both the finite and continuum domain settings for the finite sum problem \eqref{eq:goal1}.
Performing the one-step value of information analysis at the heart of BQO requires solving a challenging optimization problem, and we present novel computational methods that solve this problem efficiently,
including a novel discretization-free method for estimating the gradient of the value of information, a new convergence analysis of a different and less efficient discretized scheme more closely related to past work, and a novel transformation that provides a more computationally convenient form of $F$. 
%In our examples ($\mathsection$\cref{sec:citibike}, $\mathsection$\cref{sec:IPexample}) we show how to transform any simulation optimization problem into a formulation as \eqref{eq:goal1} and \eqref{eq:goal2}, which means that IBO obtains the best results on simulation optimization when there are noisy observations. 
We demonstrate that our algorithm substantially outperforms state-of-the-art Bayesian optimization methods when observations are noisy or the integrand varies smoothly in the integrated variables, and performs as well as state-of-the-art methods in the remaining settings.  Our demonstrations use a variety of problems from optimization via simulation and hyperparameter tuning in machine learning. 
We also provide a robust implementation of our method at \url{https://github.com/toscanosaul/bayesian_quadrature_optimization}.

% We show that IBO performs much better than classical BO methods such as expected improvement and knowledge gradient in the noisy case, and quite well in the deterministic case. Moreover, we show how to apply our method to realistic examples in different areas such as simulation optimization, machine learning, and operations research. 
% We demonstrate that our algorithm substantially outperforms state-of-the-art BGO methods in the noisy setting, and performs as good as state-of-the-art methods in the noise-free setting. We demonstrate the effectiveness of our method in optimization problems based on both a queuing and inventory problem simulation, and cross-validation for choosing the optimal architecture of a convolutional neural network and collaborative filtering. 

Our method improves over the previous literature in three ways: 
First, it is more \textit{general}, as it is the first to allow noise in the evaluation of $F$, the first to simultaneously support solving both \eqref{eq:goal1} and \eqref{eq:goal2}, and allows general $p$ in contrast with \citet{groot2010bayesian} and \citet{Xie:2012}'s requirement that $p$ be a normal density.
Second, it is more \textit{well-supported theoretically}, as its one-step optimality justification contrasts with the heuristic justification offered in \citet{williams2000sequential,groot2010bayesian} and \citet{swersky2013multi}. (\citealt{Xie:2012} is one-step Bayes-optimal for the special case of \eqref{eq:goal2} that it considers.)  Also none of these previous methods come with a proof of consistency, and \citet{williams2000sequential} may fail to be consistent if a poor tie-breaking rule is chosen as we note below. Third, it provides \textit{better empirical performance} in problems with noisy evaluations or when the integrand varies smoothly in the integrated variables, and performs comparably in other problems. We discuss this previous literature in more detail below.

This paper significantly extends the conference paper \cite{MCQMC}, where an early version of the BQO method was referred to as Stratified Bayesian Optimization (SBO).  We have re-named our method to reflect its more general ability to solve problems beyond the second use-case based on stratification described at the start of this section. Beyond that conference paper, the current paper includes proofs of consistency for both finite and continuum domains for the finite sum problem \eqref{eq:goal1}, a proof of convergence of the discretized computational method used in that paper, a new discretization-free computational method that is substantially more efficient in higher dimensions, and additional numerical experiments on new problems with new benchmark algorithms. 

% Our method IBO uses a prior Gaussian process on $F$, and chooses $(x_{n}, \w_{n})$ using a one-step Bayes-optimal acquisition function based on a value-of-information (\citet{Ho66}) analysis. It then uses the resulting observation within a Bayesian quadrature framework to update its Gaussian process posterior on both $(x,\w)\mapsto F(x,\w)$, and the integral/sum. By using more information, we make our statistical model more powerful, and provide better answers with fewer samples than standard Bayesian optimization techniques. 

\paragraph{Detailed Discussion of Related Work:}
\citet{williams2000sequential} considers the problem \eqref{eq:goal1} when $F$ is noiseless, and uses a small modification of the well-known expected improvement acquisition function \citep{Mockus1989,jones1998efficient}. Their acquisition function is a two step procedure which first uses expected improvement to choose $x\in A$ by maximizing the conditional expectation of $\mbox{max}\left\{ 0,G\left(x\right)-\mbox{max}_{1\leq i\leq n}G\left(x_{i}\right)\right\} $ given the past $n$ observations, and then chooses $w\in W$ by minimizing the posterior mean squared prediction error. This algorithm is not consistent for finite $A$ for the following reason: After $F(x,w)$ has been evaluated for all $x\in A$, (but not necessarily all $w \in W$), 
$G\left(x\right)-\mbox{max}_{x\in A}G\left(x\right)\leq0$ almost surely.  This implies that the conditional expectation of $\mbox{max}\left\{ 0,G\left(x\right)-\mbox{max}_{x\in A}G\left(x\right)\right\} $ is $0$ for all $x$.  If the tie-breaking rule used chooses the same $x$ on each iteration, then this method will fail to evaluate each $x$ infinitely often.  
\citet{lehman2004} considers minor modifications of the previous algorithm, and their M-robust algorithm can also fail to be consistent with a poor tie-breaking rule. 

\citet{groot2010bayesian} considers problem \eqref{eq:goal2} when $F$ is noiseless, $p(\w)$ is Gaussian, and the kernel of the Gaussian process on $F$ is the squared exponential kernel. Its acquisition function is a minor modification of the active learning method ALC \citep{cohn1996}.   Numerical experiments in that paper do not demonstrate an improvement over evaluating $G$ directly. While the method proposed for choosing $x,w$ is motivated by minimizing the expected variance of the objective $G$ after one evaluation, it does not do so optimally.  Instead, like \citet{williams2000sequential}, it chooses $x$ ignoring what $w$ will be chosen, and then chooses $w$ with $x$ fixed.  This is in contrast with our approach, which chooses $x$ and $w$ jointly in a one-step optimal way.  

\citet{swersky2013multi} considers problem \eqref{eq:goal1} when $F$ is noiseless, and uses a small modification of the expected improvement acquisition function \citep{jones1998efficient}. Like \citet{williams2000sequential} and \citet{groot2010bayesian}, and in contrast with our joint optimization approach, it first chooses $x$ ignoring $w$, and then in a second step it chooses $w$ with $x$ fixed.  
Although one should typically choose $w$ to reduce uncertainty about $G(x)$, \citet{swersky2013multi} instead chooses $w$ using an expected improvement criterion over $F(x,w)$ even though we are not maximizing over $w$.  This can select points whose posterior mean of $F(x,w)$ is high but posterior variance is extremely low, essentially wasting a measurement.
This leads in turn to examples where the policy repeatedly samples the same $x$ and under-explores, as we discuss in the appendix ($\mathsection$\ref{bad_example_mt}).
Our numerical experiments show this method can perform well in problems with a small number of homogeneous tasks, but tends to underperform significantly as the number of tasks increase.

\citet{Xie:2012} considers problem \eqref{eq:goal2} when $F$ is noiseless, $p$ is Gaussian and independent, and the Gaussian process on $F$ has a squared exponential kernel. BQO generalizes the method in that paper to the significantly more applicable setting where $F$ can be noisy (required for application to optimization via simulation), with any $p$ (required for applications to cross-validation in machine learning) and any kernel (required for good performance on a wider variety of problems).  These generalizations significantly increase the difficulty of the problem, because they preclude closed-form expressions used in \citet{Xie:2012}.
We also provide significantly improved computational methods: 
the discretized method used in \citet{Xie:2012} to optimize the acquisition function requires computation that scales exponentially in the dimension, preventing its use for more than $3$ dimensions, while our discretization-free method has sub-exponential scaling and numerical experiments demonstrate excellent performance on problems in up to $7$ dimensions.  Although \citet{Xie:2012} does not provide theoretical analysis, one can see our convergence proof for the discretized method as addressing theoretical questions left unanswered by that previous work. We also go beyond \citet{Xie:2012} in extending our methodology to be fully Bayesian by sampling Gaussian process hyperparameters from their posterior distribution using slice sampling.

Other related work includes \citet{marzat2013}, which considers a related but different formulation of \eqref{eq:goal1} based on maximizing worst-case performance over a discrete set of environmental conditions. \citet{lam2008} considers a modification of \citet{williams2000sequential} where the criterion used is for response surface model fit instead of global optimization.

Our consistency proof for the finite sum problem \eqref{eq:goal1} is the first for any algorithm that evaluates $F$ instead of $G$. Consistency of some Bayesian optimization algorithms that evaluate $G$ have, however, been shown in the literature. \citet{frazier2009knowledge} proved consistency of the knowledge gradient algorithm for any Gaussian process for finite domains. Later \citet{bull2011} proved consistency of expected improvement for functions that belong to the reproducing kernel Hilbert space (RKHS) of the covariance function. The recent working paper \citet{bect2016} also contains consistency results for knowledge gradient and expected improvement over any Gaussian process with continuous paths.

Our BQO algorithm can be considered to be within the class of knowledge gradient policies \citep{PowellFrazier2008}, because it selects the ($x,\w$) to sample that maximizes the expected utility of the final solution, under the assumption, made for tractability, that we may take only one additional sample. Work on knowledge gradient algorithms in other settings includes \citet{frazier2009knowledge,poloczek2017multi,wugradientsbo}.  Our discretization-free approach leverages ideas in particular from \cite{wugradientsbo}.
Our algorithm also leverages Bayesian quadrature techniques \citep{o1991bayes}, which build a Gaussian process model of the function $F(x,\w)$, and then use the relationships given by the sum or integral to imply a second Gaussian process model on the objective $G$.

%The main difference between them is that \citet{swersky2013multi} considers w discrete, and \citet{williams2000sequential} considers only w continuous.

% \citet{williams2000sequential} only applies its method to analytic test functions, and \citet{swersky2013multi} only to one cross-validation problem. In both cases, the authors did not compare their method with standard BO methods such as expected improvement.

The rest of this paper is organized as follows: $\mathsection$\ref{model} presents our statistical model. $\mathsection$\ref{SBO} presents the conceptual value of information analysis underlying the BQO algorithm. $\mathsection$\ref{sec:VOI} describes computation of the value of information and its derivative, and presents the BQO algorithm in a practically implementable form. $\mathsection$\ref{sec:asymptotic} presents theoretical results on consistency of BQO. $\mathsection$\ref{experiments} presents  simulation experiments. $\mathsection$\ref{conclusion} concludes.

\section{Statistical Model}
\label{model}
Our BQO algorithm relies on a Gaussian process (GP) model of the underlying function $F$, which then implies 
% (because both integration and sum are linear functions)
a Gaussian process model over $G$. Before presenting BQO in $\mathsection$\ref{SBO}, we present this statistical model to provide notation used through the rest of the paper. The first part of our development is standard in Bayesian optimization \citep{jones1998efficient} and Bayesian quadrature \citep{o1991bayes}, while the second part, in which a Gaussian process on the function's integral or sum is obtained, is only standard in Bayesian quadrature.

% This statistical approach is standard (Gaussian process modeling of an unknown function is used throughout Bayesian optimization, and its implication of a Gaussian process model on the function's integral is known in Bayesian quadrature, e.g., \citealt{o1991bayes})  
% but we summarize it here to define notation used later.

We suppose that observing $F$ at $x,w$ provides an observation
$y\left(x,w\right)$ equal to $F\left(x,w\right)$ optionally perturbed by 
additive independent normally distributed noise with mean $0$ and variance
$\lambda_{\left(x,w\right)}$. 
To permit estimation, we require one of two additional assumptions on this noise:
either that $\lambda_{\left(x,w\right)}$ is constant across the domain; 
or that observing at $x,w$ also provides an observation of $\lambda_{\left(x,w\right)}$.
The first assumption has been shown to be effective in a wide range of applications in the Bayesian optimization literature \citep{snoek2012practical}. The second is reasonable in discrete-event simulation applications in which $y(x,w)$ is the average of a large batch of independent replications. In such applications, the difference between $y(x,w)$ and its mean $F(x,w)$ converges to a normal distribution by the central limit theorem as the batch size grows large, and $\lambda_{\left(x,w\right)}$ can be estimated by dividing the sample variance of these samples by their number \citep{kim:2007}.

We assume that the function $F$ follows a Gaussian process prior distribution:
\[
F\left(\cdot,\cdot\right)\mid \theta\sim GP\left(\mu_{0}\left(\cdot,\cdot;\theta\right),\Sigma_{0}\left(\cdot,\cdot,\cdot,\cdot; \theta\right)\right),
\]
where $\mu_{0}$ is a real-valued function taking arguments $x,\w$ (the {\it mean function}), $\Sigma_{0}$ is a positive semi-definite function taking arguments $x,\w,x',\w'$ (the {\it kernel}), and $\theta$ are the hyperparameters of the mean function and kernel.  $\theta$ contains $\lambda_{(x,w)}$ when the variance of the observational noise is assumed to be unknown and constant. Common choices for $\mu_0$ and $\Sigma_0$ from the Gaussian process regression literature  \citep{RaWi06,murphy2012machine,goovaerts1997geostatistics,seeger2005semiparametric,bonilla2007multi}
appropriate for problem \eqref{eq:goal2}
include setting $\mu_0$ to a constant and letting $\Sigma_0$ be the squared exponential or Mat\'{e}rn $5/2$ kernel. In the case of the finite sum \eqref{eq:goal1}, kernels from the intrinsic model of coregionalization are appropriate \citep{seeger2005semiparametric, goovaerts1997geostatistics, bonilla2007multi} and will be discussed in $\mathsection$\ref{experiments}.

Following work on fully Bayesian inference in GP regression \citep{Neal:GPBayesian}, we additionally place a Bayesian prior distribution $\pi$ on $\theta$.  This prior can regularize values of $\theta$ used in inference, pushing them toward regions of the space of hyperparameters believed to best correspond to the data.  The prior can also be set constant if there is enough data to obviate such regularization.  

We now discuss inference supposing that we have $n$ points in the historical data $H_{n}:= \left(y_{1:n},\w_{1:n},x_{1:n}\right)$, where $y_{i} = y(x_i,\w_i)$ is a (possibly noisy) observation of $F(x_{i},\w_{i})$ with the conditional distribution given $x_i$,$\w_i$ described above.
Within our inference procedure we sample $\theta$ from its posterior distribution given $H_n$ via slice sampling \citep{radford2003Slice}.  One may also replace this sampling-based fully Bayesian treatment of $\theta$ by using the maximum {\it a posteriori} estimate (MAP), which sets $\theta$ to its posterior mode  \citep{murphy2012machine}.  This is less computationally intensive, but tends to be less accurate.  The maximum likelihood estimate (MLE) of $\theta$ is a particular case of the MAP when the prior distribution on $\theta$ is flat. 

Using this procedure to sample $\theta$,
we now describe computation of the posterior distribution on both $F$ and $G$ given $\theta$.  The posterior distribution on $F$ given $\theta$ at time $n$ is 
\[
F\left(\cdot,\cdot;\theta\right)\mid H_{n}, \theta\sim GP\left(\mu_{n}\left(\cdot,\cdot; \theta\right),\Sigma_{n}\left(\cdot,\cdot,\cdot,\cdot;\theta\right)\right),
\]
where the parameters $\mu_{n}$, $\Sigma_{n}$ can be computed using standard
results from Gaussian process regression \citep{RaWi06}.  
To support later analysis, we provide these expressions here, suppressing dependence on $\theta$ in our notation: 
% for $\mu_n$ and $\Sigma_n$:
%where $\mathcal{F}^{n}$ be the sigma-algebra generated by $H_{n}:= \left( y_{1:n}
\begin{eqnarray}
\mu_{n}\left(x,w\right) & = & \mu_{0}\left(x,w\right)+\left[\Sigma_{0}\left(x,w,x_{1},w_{1}\right)\mbox{ }\cdots\mbox{ }\Sigma_{0}\left(x,w,x_{n},w_{n}\right)\right]A_{n}^{-1}\left(\begin{array}{c}
y_{1}-\mu_{0}\left(x_{1},w_{1}\right)\\
\vdots\\
y_{n}-\mu_{0}\left(x_{n},w_{n}\right)
\end{array}\right)\label{post_mean_F}\\
\Sigma_{n}\left(x,w,x',w'\right) & = & \Sigma_{0}\left(x,w,x',w'\right)-\left[\Sigma_{0}\left(x,w,x_{1},w_{1}\right)\mbox{ }\cdots\mbox{ }\Sigma_{0}\left(x,w,x_{n},w_{n}\right)\right]A_{n}^{-1}\left(\begin{array}{c}
\Sigma_{0}\left(x',w',x_{1},w_{1}\right)\\
\vdots\\
\Sigma_{0}\left(x',w',x_{n},w_{n}\right)
\end{array}\right) \label{post_cov_F}
\end{eqnarray}

where 
\[
A_{n}=\left[\begin{array}{ccc}
\Sigma_{0}\left(x_{1},w_{1},x_{1},w_{1}\right) & \cdots & \Sigma_{0}\left(x_{1},w_{1},x_{n},w_{n}\right)\\
\vdots & \ddots & \vdots\\
\Sigma_{0}\left(x_{n},w_{n},x_{1},w_{n}\right) & \cdots & \Sigma_{0}\left(x_{n},w_{n},x_{n},w_{n}\right)
\end{array}\right]+\mbox{diag}\left(\lambda_{(x_{1},w_{1})},\ldots,\lambda_{(x_{n},w_{n})}\right).
\]

We now describe the posterior distribution on the objective function $G$ given $\theta$.
We assume that $G$ is written in its integral form \eqref{eq:goal2}.  Results for \eqref{eq:goal1} are similar, where the resulting expressions are obtained by replacing integration over $w$ by a sum over $w$ (or equivalently Lebesgue integration with respect to a counting measure).
We denote by $E_{n}$, $\mathrm{Cov}_{n}$, and $\mathrm{Var}_n$ the conditional expectation, conditional covariance, and conditional variance with respect to the Gaussian process posterior given $H_{n}$ and $\theta$.
By results from Bayesian quadrature \citep{o1991bayes}, for
$G(x) := \int F(x,\w) p(\w)\,d\w$, we have that
\begin{align}
E_{n}\left[G(x) \right] 
&= \int\mu_{n}(x,w)p\left(w\right)d\w := a_n(x;\theta), \label{eq:a_n} \\
\mbox{Cov}_{n}\left(G(x),G(x')\right)
& = \int\int\Sigma_{n}\left(x,\w,x',\w'\right)p\left(\w\right)p\left(\w'\right)d\w\,d\w'. \label{eq:cov_n_g}
\end{align}

Ignoring some technical details, the first line is derived using interchange of integral and expectation, as in 
$E_{n}\left[G(x) \right] 
= E_{n}\left[\int F(x,\w) p(\w)\,d\w \right] 
= \int E_{n}\left[F(x,\w) p(\w) \right] \,d\w
= \int\mu_{n}(x,w)p\left(w\right)d\w$.
The second line is derived similarly, though with more effort, by writing the covariance in terms of the expectation and interchanging expectation and integration.

The posterior distributions of $F$ and $G$ given $H_n$ and marginalizing over $\theta$ are infinite mixtures of Gaussian processes. Means of these posterior distributions can be obtained by averaging \eqref{post_mean_F} or \eqref{eq:a_n} with respect to the posterior on $\theta$.

% Is it possible to fill this in in a simple way?
%$\mbox{Cov}_{n}\left(G(x),G(x')\right)
%=\mathbb{E}_{n}\left[(G(x)-\int \mu_n(x,\w) p(\w)\,d\w)(G(x')-\int \mu_n(x',\w') p(\w')\,d\w') \right]
%=\mathbb{E}_{n}\left[(\int ((F(x,\w)-\mu_n(x,\w)) p(\w)\,d\w)(\int (F(x',\w')-\mu_n(x',\w')) p(\w')\,d\w') \right]
%=\mathbb{E}_{n}\left[\int F(x,\w)-\mu_n(x,\w) p(\w), d\w- 
%G(x)-\mu_n(x),G(x')-\mu_n(x')\right]

\section{Conceptual Description of the BQO Algorithm}
\label{SBO}

%%%%%%%%%%%%%%%%%%%%%%%%%%%%%%%%%

Our BQO algorithm uses the statistical model described in $\mathsection$\ref{model} and samples $F$ sequentially. 
It chooses where to sample $F$ using a value of information analysis \citep{Ho66}.
This analysis measures the expected quality of the best solution we can provide to \eqref{eq:goal1} or \eqref{eq:goal2} after $n$ samples, and how this quality improves with an additional sample of $F(x,\w)$.
In this section we describe this value of information analysis from a conceptual perspective in preparation for describing in $\mathsection$\ref{sec:VOI} the novel computational methodology we create to make its implementation possible.
This conceptual value of information analysis echos the use of value of information in Bayesian optimization in works on knowledge-gradient methods \citep{frazier2009knowledge} and in problem (2) by \citet{Xie:2012}. 
The value of information analysis we describe here generalizes \citet{Xie:2012} to support noisy observations, integrals with dependent normally distributed densities, non-normally distributed densities, and sums, and general kernels.  
While the generalization of the {\it conceptual} form of the value of information analysis from \citet{Xie:2012} to handle this richer class of problems is straightforward, it presents a host of new computational challenges that requires new methodology, as fully described in $\mathsection$\ref{sec:VOI}.

We conduct our value of information analysis assuming the hyperparameters $\theta$ are given, as is common practice in Bayesian optimization \citep{swersky2013multi,reviewBO}.  Then, in the implementation of the BQO algorithm, because $\theta$ is unknown, we average this $\theta$-dependent value of information over the posterior on $\theta$. While in principle one could instead conduct the full value of information analysis acknowledging that $\theta$ is unknown, proceeding as we do provides substantial computational benefits.

To conduct this analysis,
we first study the expected quality of the best solution we can provide. 
% current best expected solution after $n$ samples, so that we can analyze how it improves in the next iteration when a new sample is taken.
Given $n$ samples, $\theta$, and a risk-neutral utility function, we would choose as our solution to \eqref{eq:goal1} or \eqref{eq:goal2},
\begin{equation*}
x_{n,\theta}^{*} \in \mbox{argmax}_{x \in A} E_{n}\left[G(x)\mid \theta\right]=\mbox{argmax}_{x\in A} a_n(x;\theta),
\end{equation*}
where $a_n(x;\theta):=E_{n}[G(x)\mid \theta]$.
%This is the Bayes-optimal solution when we are risk neutral.
This solution has expected value (again, with respect to the posterior after $n$ samples given $\theta$),
\begin{equation*}
a_{n,\theta}^* := E_{n}\left[G\left(x_{n,\theta}^{*}\right)\mid \theta\right] = \max_{x}E_{n}\left[G(x)\mid \theta\right]=\max_{x} a_n(x; \theta).
\end{equation*}

Consequently, the improvement in expected solution quality resulting from a sample at $(x,\w)$ at time $n$ is 
\begin{equation}
    V_n(x,\w; \theta) = E_n\left[ a_{n+1,\theta}^* - a_{n,\theta}^* \mid x_{n+1}=x, \w_{n+1}=\w\right],
    \label{eq:VOI}
\end{equation}
and we refer to this quantity as the {\it value of information}.  Our Bayesian Quadrature Optimization (BQO) algorithm 
then is defined as the algorithm that samples where this value of information (marginalized over $\theta$) is maximized,
\begin{equation}
    \left(x_{n+1},\w_{n+1}\right)\in\mbox{argmax}_{x,\w} E\left[V_{n}\left(x,\w; \theta\right) \mid H_n\right].  \label{eq:max_VOI}
\end{equation}
Here, the expectation is over the posterior on $\theta$, as indicated by the subscript.

This policy is one-step Bayes optimal in the known-hyperparameter case (i.e., the prior on $\theta$ is concentrated on a single value), in the sense that if we can take one more sample before reporting a final solution then its sampling decision maximizes the expected value of $G$ at this reported final solution. 
It is not necessarily Bayes-optimal if we can take more than one sample, but we argue that it remains a reasonable heuristic, and our numerical experiments in $\mathsection$\ref{experiments} support this. It is also Bayes-optimal for the problem \eqref{eq:goal1} in the known-hyperparameter case when the number of iterations converge to infinity, as we show in $\mathsection$\ref{sec:asymptotic}.

% Detailed computation of the value of information $V_{n}$, and its gradient with respect to $x$ and $\w$ when the domain space is continuum, is discussed below in $\mathsection$\ref{sec:VOI}.  When the domain space is continuum, we use the gradient of $V_{n}$ to solve \eqref{eq:max_VOI} using multi-start gradient ascent, multi-start stochastic gradient ascent (we recommend to use Adam method \citep{adam}), or multi-start sequential least squares programming (\citet{kraft1988software}). If the domain space is finite, we can just evaluate $V_{n}$ at all possible alternatives, and choose the best one.

% The IBO algorithm estimating the hyperparameters of the model with MLE or MAP is summarized below.  We can also place a prior on the hyperparameters. The difference in this case is that we solve $\left(x_{n+1},\w_{n+1}\right)\in\mbox{argmax}_{x,\w}V_{n}\left(x,\w\right)$ using  multi-start stochastic gradient ascent to optimize $V_{n}$, and $\theta$ is sampled from  its posterior using slice sampling.

% In  $\mathsection$\ref{sec:VOI}, we describe two ways of approximating the IBO policy: one using a finite discretization of the domain of the points $x$, and other one using Monte Carlo methods. The complexity of the IBO algorithm under the first one is $O(LN^2+N^4)$ if it is run during $N$ iterations, and $L$ is the number of points in the discretization of the domain of the points $x$. If we use a Monte Carlo estimator instead, the complexity of the algorithm is $O(N^4)$ if it is run during $N$ iterations (see $\mathsection$\ref{sec:VOI}). 

\begin{figure}[!htbp]
  \centering
  \subcaptionbox{The contours of $F\left(x,\w\right)$.  $G$ is determined \\ from $F$ by $G(x) = \int F(x,\w) p(w)\, dw$.}[0.45\linewidth]{
      \includegraphics[width=0.45\linewidth,height=2.1in]{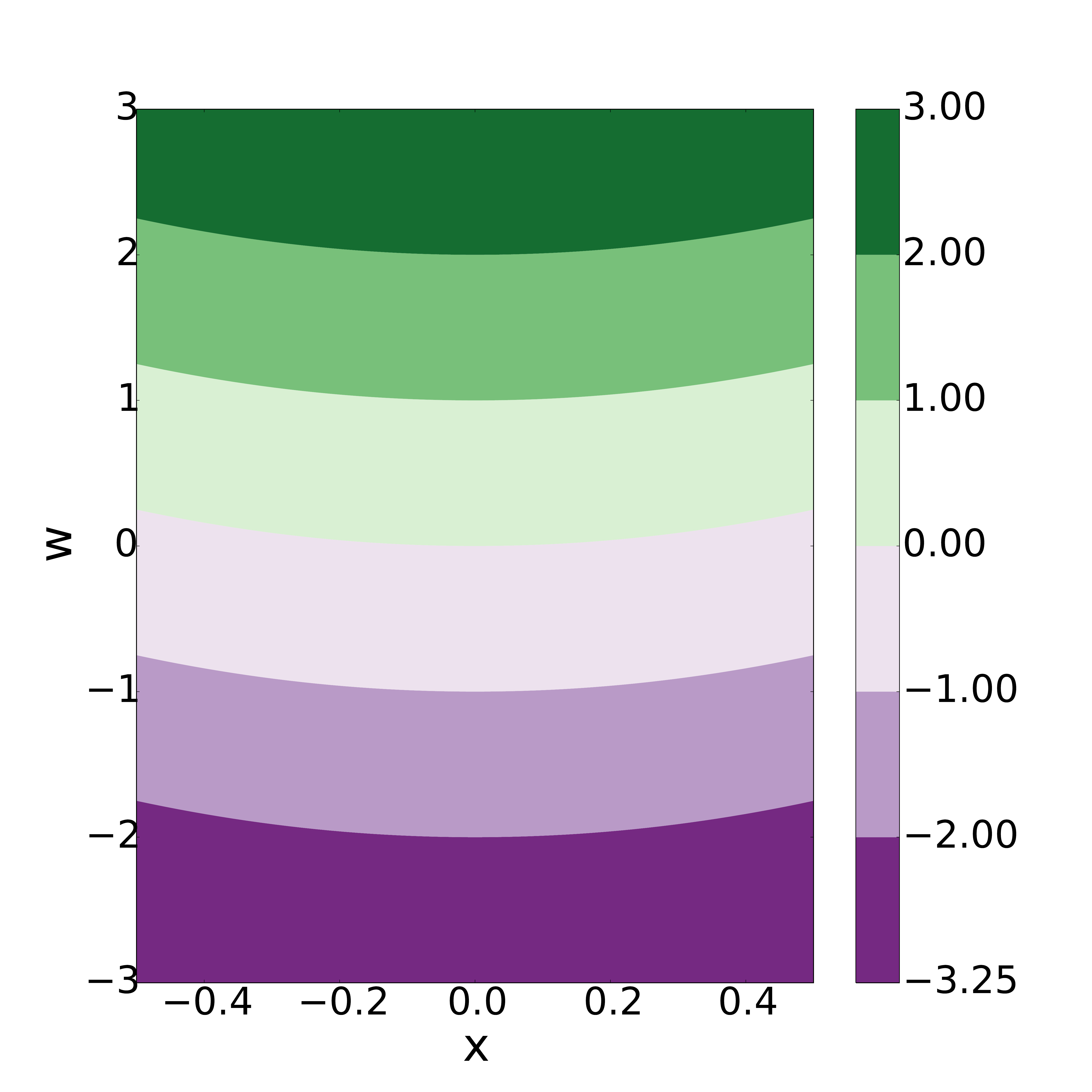}}
   %   \quad
  \subcaptionbox{The contours of BQO's estimate $\mu_n(x,\w; \theta)$ of $F\left(x,\w\right)$ after $n=9$ evaluations of $F$ by BQO.}[0.45\linewidth]{
      \includegraphics[width=0.45\linewidth,height=2.1in]{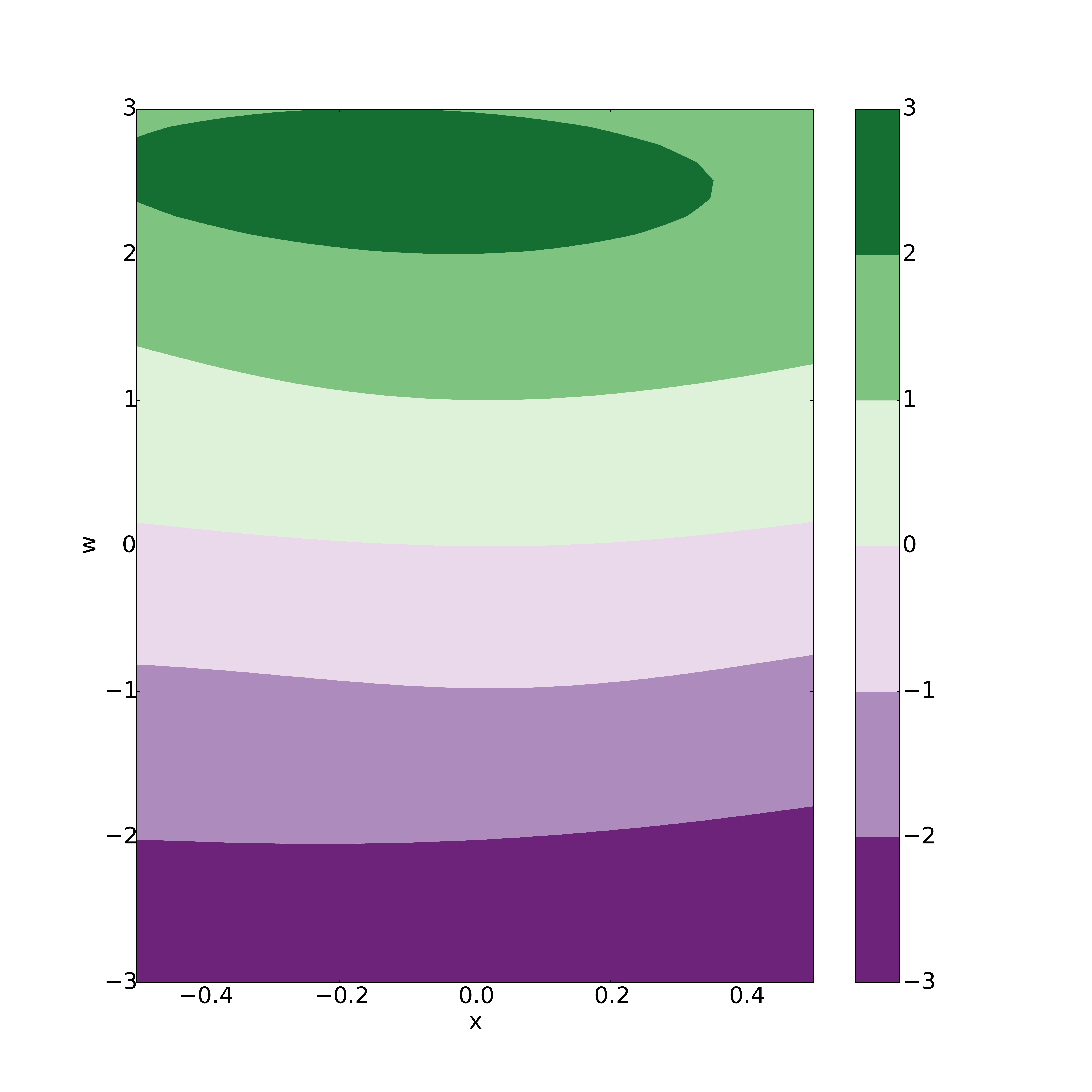}}
  \subcaptionbox{
      The contours of BQO's value of information $V_n(x,\w; \theta)$ versus $x$ and $\w$. $F$ was evaluated previously at the red points, chosen according to a random uniform design in an initial phase of training, and at $n=9$ black points, chosen by BQO.}[0.45\linewidth]{
      \includegraphics[width=0.45\linewidth,height=2.1in]{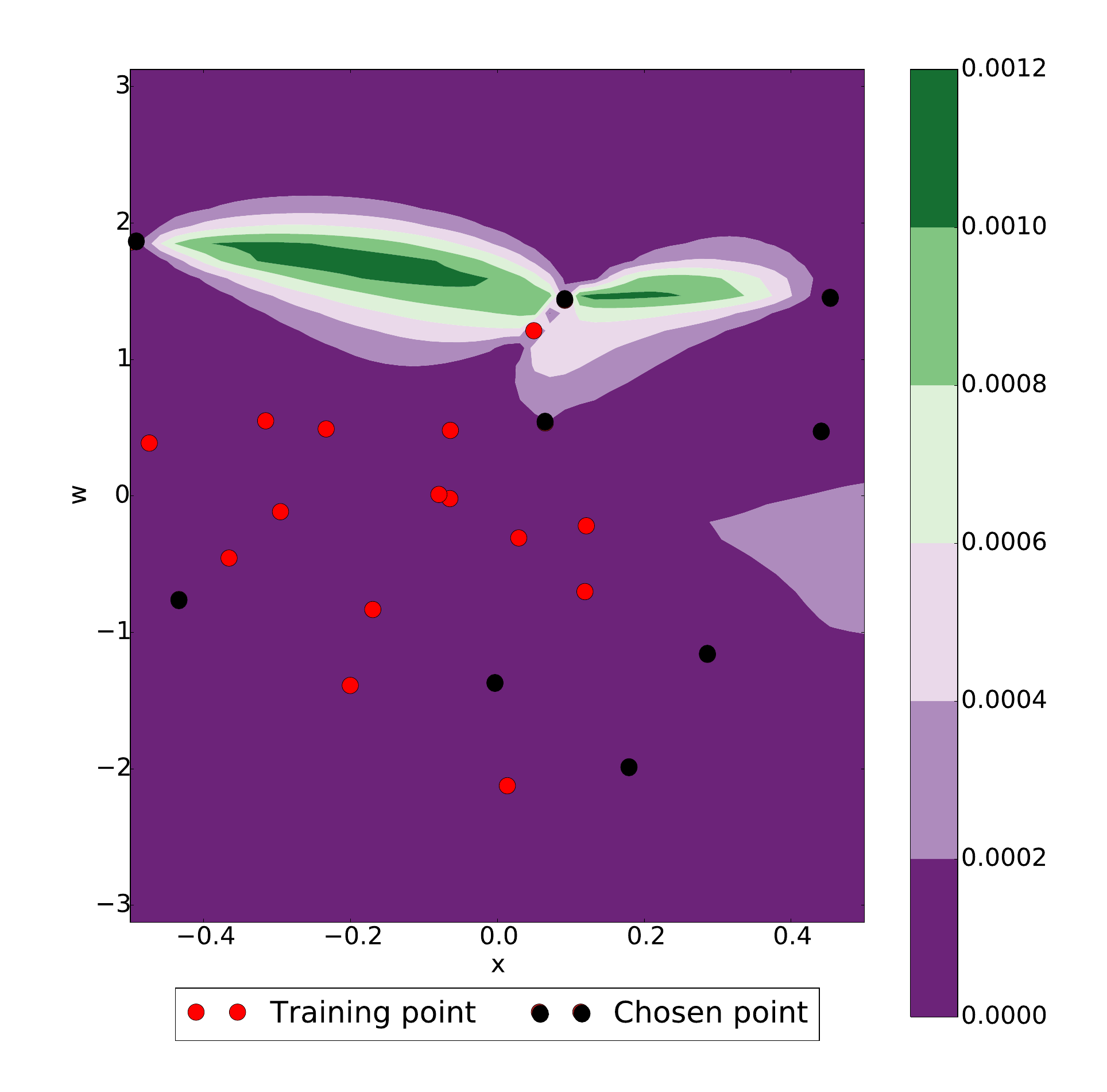}
  }
  \quad
  \subcaptionbox{
  The objective $G(x)$, BQO's estimate $a_n(x;\theta)$ of $G(x)$, and BQO's $95\%$ credible interval for $G(x)$ after  $n=9$ evaluations of $F$ by BQO.
  The estimate of $G$ is extremely close to its true value, especially near its maximum.
  }[0.45\linewidth]{
      \includegraphics[width=0.37\linewidth,height=2.0in]{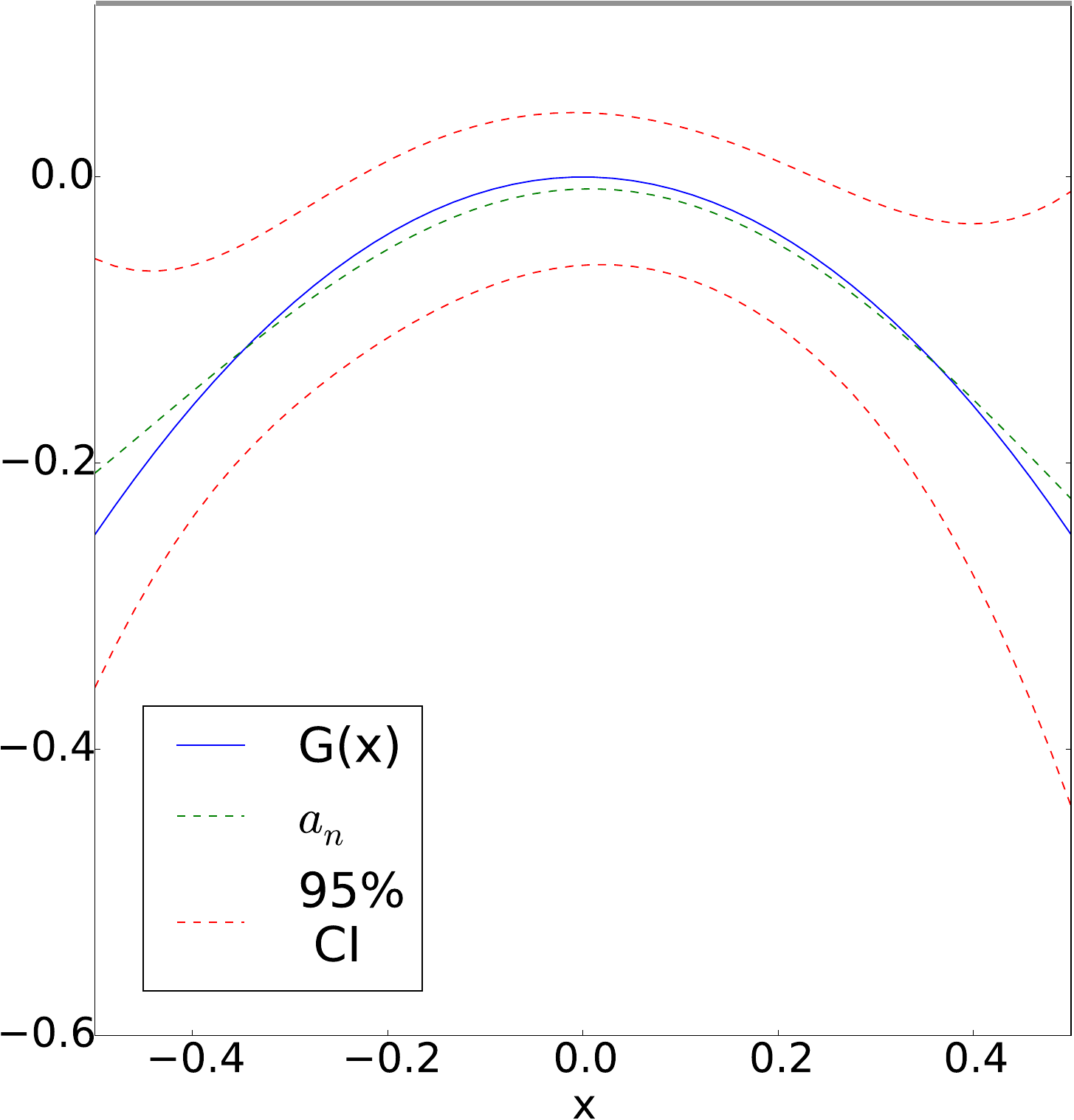}\hfill{}
  }
  \smallskip
\caption{
Illustration of the BQO algorithm on an analytic test problem after evaluating $F$ at points chosen uniformly at random in an initial phase of training and $n=9$ points chosen by BQO.   
\label{fig:tahi10-IBO}}
\end{figure}

Figure~\ref{fig:tahi10-IBO} illustrates BQO, showing one step in the algorithm applied to a simple analytic test problem 
\begin{equation}
\label{eq:test}
\max_{x\in\left[-\frac12,\frac12\right]}E\left[\z x^{2}+\w\right],
\end{equation}
where $\w\sim N\left(0,1\right)$, $\z\sim N\left(-1,1\right)$, and $F(x,\w)=E[zx^2+w\,|\,w] = -x^2+w$.   Direct computation shows $G(x)=-x^{2}$. In the figure, we fix $\theta$ to a maximum likelihood estimate obtained using 15 training points.

The figure's first row shows the contours of $F(x,\w)$ (left panel) and BQO's estimate (right panel) after evaluating $F$ at points chosen uniformly at random in an initial training phase, and at an additional $n=9$ points chosen by BQO.
The second row's left panel shows the value of information $V_n(x,\w;\theta)$.  
The value of information is small near where BQO has already sampled, because it has less uncertainty about $F(x,\w)$ in this region. 
BQO's value of information is also small for extreme values of $x$, because its posterior on $G$ suggests that these $x$ are far from its maximum, and small for extreme values of $w$ because $p(x,w)$ is small there. BQO's value of information is thus largest for points that are far from previous samples, relatively close to the maximizer of $G$'s posterior mean, and have moderate values of $w$. 
BQO samples next at the point with the largest value of information, near $x=-0.2$ and $\w=1.8$.
% IBO will sample at the $(x,\w)$ pair with largest VOI, which is to decide where to sample.  IBO will sample near $x=-0.2$ and $\w=1.8$.  This choice of $(x,\w)$ is reasonable because it is far from previous evaluations (causing us to have high uncertainty about $F$), has a value of $p(w)$ large enough that $F$ in this vicinity contributes significantly to $G$, and the posterior mean on $G(x)$ is relatively close to $a^*_n$.  
The second row's right panel shows the posterior on $G$.
%The posterior mean of $G$ is close to its true value, and the point with maximum posterior mean is close to the true maximizer of $G$.
This posterior is accurate and almost perfectly estimates $G$'s maximizer. 

\begin{figure}[!htbp]
\centering
  \subcaptionbox{
  The value of information versus $x$ under a traditional Bayesian optimization method (KG).  The value pictured is after noisy evaluations of $G$ at the red points, chosen in an initial phase of training, and the $n=9$ black points chosen by KG.}[0.45\linewidth]{
      \includegraphics[width=0.45\linewidth,height=2.1in]{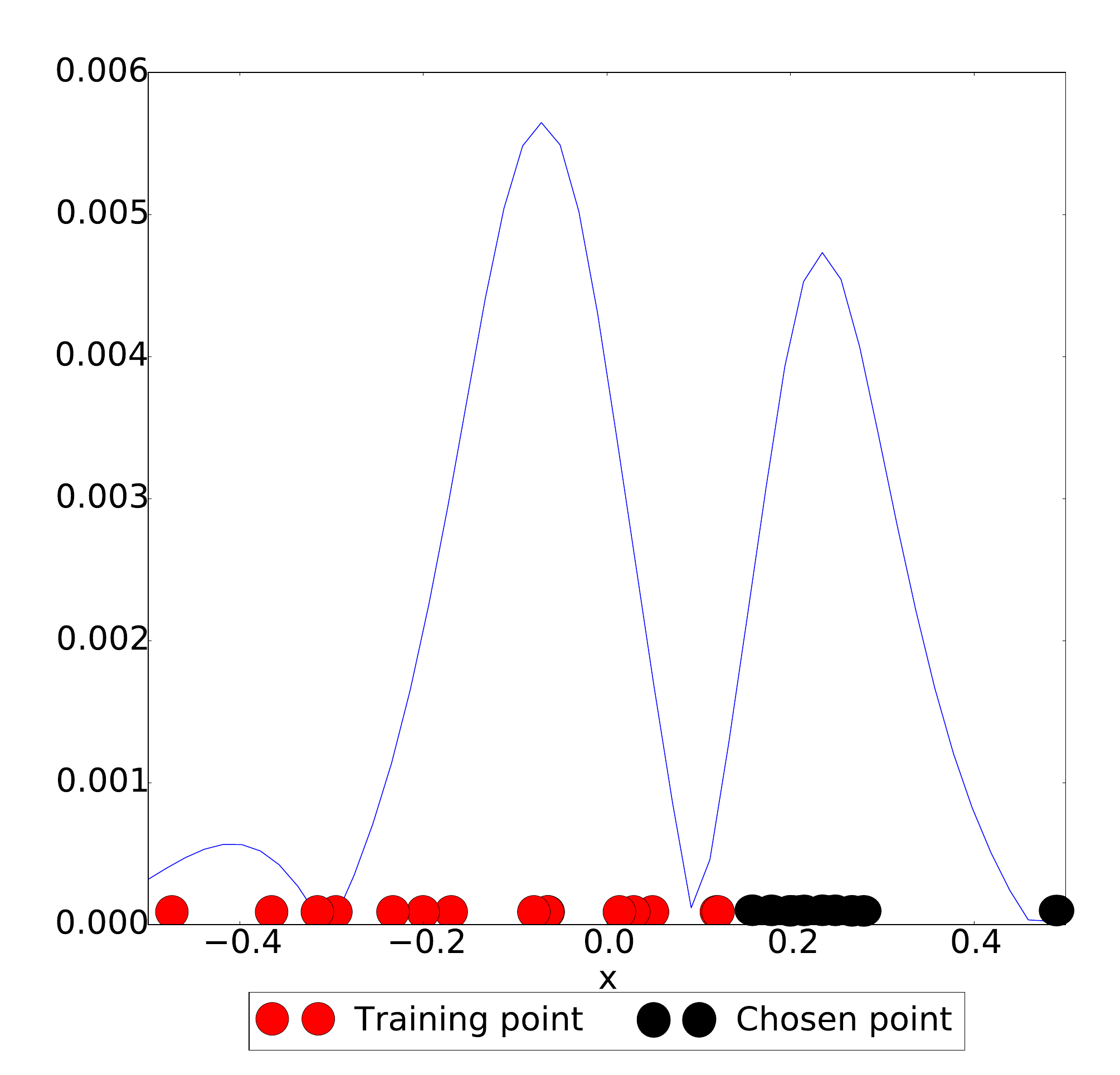}
  }
  \quad
  \subcaptionbox{
  The objective $G(x)$, KG's estimate $\mu_n(x; \theta)$ of $G(x)$; and KG's $95\%$ credible interval for $G(x)$, after $n=9$ noisy evaluations of $G$. This estimate is of lower quality than BQO's because they do not use the observed values of $\w$.
  }[0.45\linewidth]{
      \includegraphics[width=0.34\linewidth,height=2.0in]{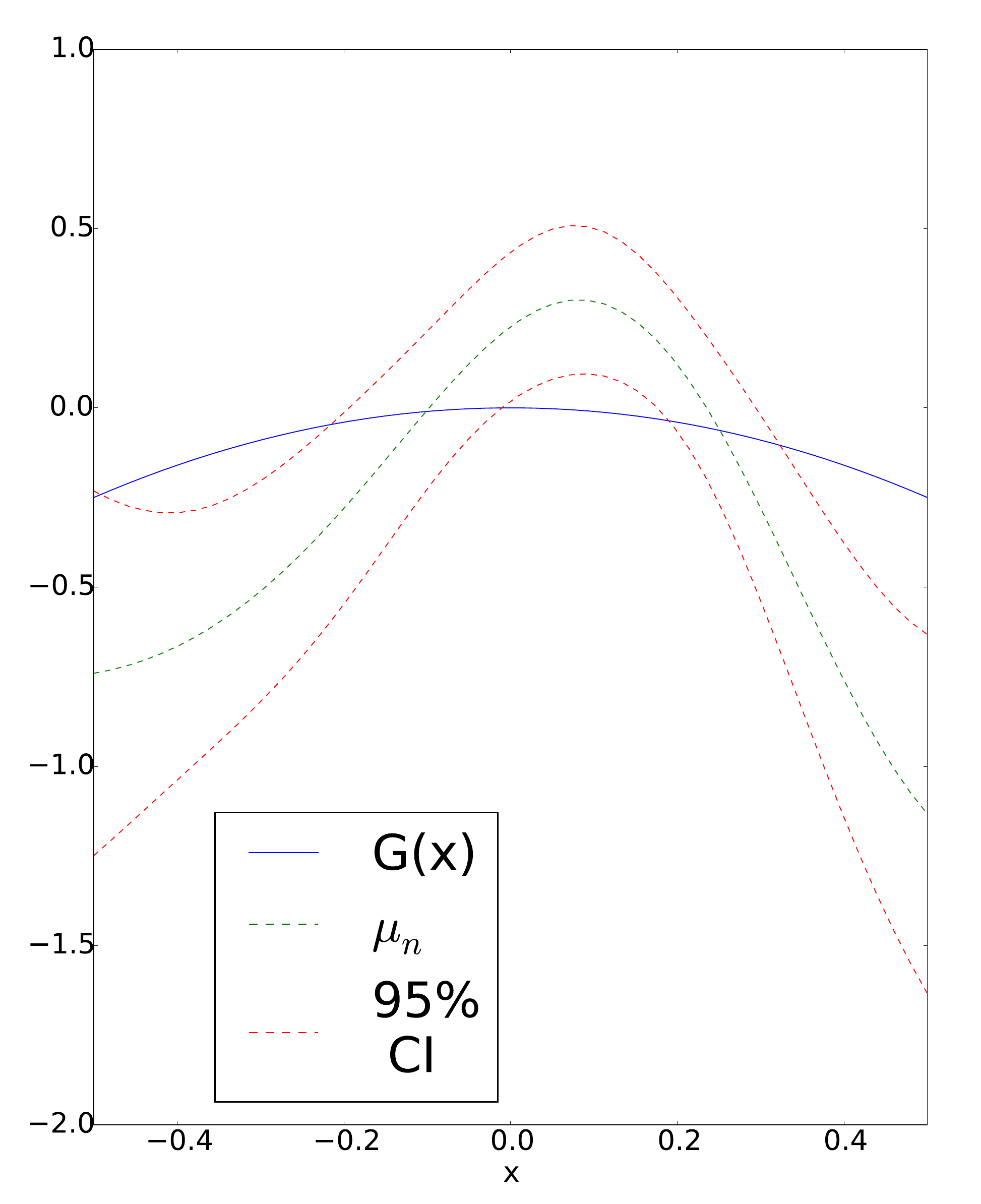}
  }
  \smallskip
\caption{
Illustration of a traditional Bayesian optimization algorithm in the same problem setting as Figure~\ref{fig:tahi10-IBO}.  The algorithm pictured is the knowledge gradient (KG) method \citep{frazier2009knowledge}.  This algorithm evaluates $G$, unlike BQO's evaluations of $F$.  As a consequence, it tends to provide lower-quality estimates of $G$ within a given sampling budget.
\label{fig:tahi10-KG}}
\end{figure}

Figure~\ref{fig:tahi10-KG} shows equivalent quantities for the knowledge-gradient (KG) method \citep{frazier2009knowledge}, 
after noisy evaluations of $G$  
at points chosen uniformly at random in an initial training phase, and at an additional $n=9$ points chosen by KG. Like other traditional Bayesian optimization methods, KG models $G(x)$ directly, ignoring valuable information from $w$, and computes a value of information as a function of $x$ only while leaving the choice of $\w$ to chance.  As a consequence, KG's estimates of $G$ and its maximizer have significantly more error than BQO's estimates.

% considered below in more detail in $\mathsection$\ref{sec:test}, comparing the posterior and value of information that it computes to that computed by a benchmark Bayesian optimization algorithm from the literature, the knowledge-gradient algorithm of \citet{frazier2009knowledge}, maintains a posterior only on $G$, and is also similar to the expected improvement algorithm of \citet{jones1998efficient}. SBO builds a statistical model of $F(x,\w)$, which implies a statistical model on $G(x)$, rather than building a model on $G(x)$ directly like other Bayesian optimization methods.

% the progress of SBO and benchmark KG algorithm on this problem, and shows that modeling $F$, as does SBO, rather than modeling $G$ directly, as does KG, provides a higher quality estimate of $G$.

\section{Computation of the BQO Algorithm}
\label{sec:VOI}

In this section we develop methods to compute the value of information \eqref{eq:VOI} and its gradient, to support implementation of the BQO algorithm. 
We introduce a new and powerful method in $\mathsection$\ref{sec:monte_carlo} for computing 
unbiased stochastic estimators of the gradient of the value of information, which we refer to more briefly as stochastic gradients.
These stochastic gradients are used within a stochastic gradient ascent method to optimize the value of information. 

We also show in $\mathsection$\ref{sec:discretization} how a deterministic discretized method for approximating the value of information and its gradient, first developed in \citet{Xie:2012} for the setting without noise and independent Gaussian $p(\w)$ and kernel, can be extended to our more general setting.  When it was first proposed in \citet{Xie:2012} it lacked a theoretical analysis of its discretization error.  To address this shortcoming, we demonstrate that this discretization error vanishes asymptotically when the discretizations are sufficiently well-designed.
We refer to this method as the ``discretized method,'' and refer to the first method (which does not rely on discretization) as the ``discretization-free method.''

We provide an analysis of the computational complexity of each method, showing that the time and space complexity of the discretization-free method scale better in the dimension of $x$.  In concert with this theoretical observation, empirical observations show that the discretized method is fastest when $x$ has one or two dimensions but is too slow to be practical in higher dimensions.  In contrast, our numerical experiments ($\mathsection$\ref{experiments}) show that our novel discretization-free method is practical in dimensions as large as $7$.

% EDITIONS:
% Search and replace VOI by value of information
% check useful in dimensions as large as $10$.

To simplify proofs, we assume that $G$ has the integral form defined in \eqref{eq:goal2}. As we mentioned in $\mathsection$\ref{model}, results for \eqref{eq:goal1} are similar, where the resulting expressions are obtained by replacing integration over $w$ by a sum.
We also assume in our computation of the value of information that $\theta$ is given, as discussed in $\mathsection$\ref{SBO}, and drop the dependence on $\theta$ from our notation (except in $\mathsection$\ref{sec:IBO-alg} where we write it explicitly to support a high-level summary of the BQO algorithm).  
% To simplify the expressions, we assume that the observations of $F(x,\w)$ are noiseless, however, the equations are trivially generalizable to the case with noise.
%Due to space considerations, we keep our descriptions brief, especially in $\mathsection$\ref{sec:BO1} and $\mathsection$\ref{sec:BO2}, and detailed derivations may be found in the supplement.
Table~\ref{table:notation} summarizes notation used in this section, including both notation introduced in previous sections and new notation defined later in this section.
\begin{table}[htbp]
\caption{Table of Notation.}
\label{table:notation}
\centering
%\vskip 0.2i
\begin{center}
%\begin{small}
\begin{tabular}{rclcc}
%\begin{tabular}{lrp{6.3cm}}
\hline
$G(x)$ & $\triangleq$ & $\sum_{w=1}^{m}F\left(x,w\right)p(w)$ or $\int_{}F\left(x,w\right)p(w)dw$ \\
$V_{n}$ & $\triangleq$ & Value of information at time $n$\\
$a_{n}\left(x\right)$ & $\triangleq$ & $E_{n}\left[G(x) \right]$\\
$H_{n}$ & $\triangleq$ & History observed by time $n$\\
$\Sigma_{0}$ & $\triangleq$ & Kernel of the  Gaussian process prior distribution over the function $F$\\
$B\left(x,i\right)$ & $\triangleq$ &  $\int\Sigma_{0}\left(x,\w,x_{i},\w_{i}\right)p(\w)d\w$ if $G(x)=\int_{}F\left(x,w\right)p(w)dw$, or \\
& & $\sum_{w=1}^{m}\Sigma_{0}\left(x,\w,x_{i},\w_{i}\right)p(\w)$ if $G(x)=\sum_{w=1}^{m}F\left(x,w\right)p(w)$ for $i=1,\ldots,n+1$\\
$\gamma$ & $\triangleq$ & $(\Sigma_{0}(x_{n+1},w_{n+1},x_{1},w_{1}),\ldots,\Sigma_{0}(x_{n+1},w_{n+1},x_{n},w_{n}))^T$ \\
$\lambda_{\left(x,w\right)}$ & $\triangleq$ & Variance of the noise in evaluations of $F\left(x,w\right)$
\\
$A_{n}$ & $\triangleq$ & $\left(\Sigma_{0}\left(x_{i},w_{i},x_{j},w_{j}\right)\right)_{i,j=1}^{n} + \mbox{diag}\left(\left(\lambda_{\left(x_{i},w_{i}\right)}\right)_{i=1}^{n}\right) $ \\
%&$+\mbox{diag}\left(\sigma^{2}\left(x_{1},w_{1}\right),\ldots,\sigma^{2}\left(x_{n},w_{n}\right)\right)$
$\sigmatilde^2_n(x,x_{n+1},\w_{n+1})$ & $\triangleq$ & $\mbox{Var}_{n}\left[G\left(x\right)\right]-E_{n}\left[\mbox{Var}_{n+1}\left[G\left(x\right)\mid x_{n+1},\w_{n+1}\right]\right]$\\
$\mbox{Var}_{n}$ & $\triangleq$ & conditional variance given $H_{n}$\\
\hline
\end{tabular}
%\end{small}
\end{center}
\vskip -0.1in
\end{table}

\subsection{Preliminary Representation of the  Value of Information}
\label{sec:BO1}

In this section, we find a useful representation of the value of information \eqref{eq:VOI} that will allow us to develop the discretization-free ($\mathsection$\ref{sec:monte_carlo}) and discretized ($\mathsection$\ref{sec:discretization}) methods to approximate it and its gradient.
We first observe that we can rewrite the value of information \eqref{eq:VOI} as
\begin{align*}
V_{n}\left(x_{n+1},\w_{n+1}\right) =&E_{n}\left[\mbox{max}_{x'\in A}a_{n+1}\left(x'\right) \mid x_{n+1},\w_{n+1}\right] -\mbox{max}_{x' \in A} a_{n}\left(x'\right). \numberthis 
\label{eq:VOI_a}
\end{align*}
This expression is not directly useful from a computational perspective, so we take one step further and find the joint distribution of $a_{n+1}\left(x\right)$ across all $x$ conditioned on $x_{n+1},\w_{n+1}$ and $H_{n}$ for any $x$. This is provided by the following lemma.
% , which, along with the previous expression, will be the key element in methods for approximating the value of information and its gradient. 
The lemma is a generalization of Section 2.1 in \citet{frazier2009knowledge}, and we include the proof in the appendix $\mathsection$\ref{proofs_monte_carlo_sect}.

%we include for clarity and the integral change the equations,introduces notation, and we are going to use it in almost all the paper.

\begin{lemma}
\label{postdist}
There exists a random variable $Z_{n+1}$, whose conditional distribution given $H_{n}$ is standard normal, such that
    $a_{n+1}\left(x\right) =  a_{n}\left(x\right)+\sigmatilde_n(x,x_{n+1},\w_{n+1})Z_{n+1}$
% a_{n+1}\left(x\right) =  a_{n}\left(x\right)+\sqrt{\mbox{Var}_{n}\left[G\left(x\right)\right]-\mathbb{E}_{n}\left[\mbox{Var}_{n+1}\left[G\left(x\right)\right]\mid x_{n+1},\w_{n+1}\right]}Z_{n+1}.
for all $x$,
with 
\begin{align*}
\sigmatilde^2_n(x,x_{n+1},\w_{n+1}) :=& \mbox{Var}_{n}\left[G\left(x\right)\right]-E_{n}\left[\mbox{Var}_{n+1}\left[G\left(x\right)\right]\mid x_{n+1},\w_{n+1}\right].
\end{align*}
The posterior mean $a_{n}(x)$ of $G(x)$ can be represented by
\begin{eqnarray}
a_{n}\left(x\right)&=&\int\mu_{0}\left(x,w\right)p(w)dw+\left[B\left(x,1\right)\mbox{ }\cdots\mbox{ }B\left(x,n\right)\right]A_{n+1}^{-1}\left(\begin{array}{c}
y_{1}-\mu_{0}\left(x_{1},w_{1}\right)\\
\vdots\\
y_{n}-\mu_{0}\left(x_{n},w_{n}\right)\end{array}\right),\label{eq:post_mean_formula}
\end{eqnarray}
where $B(x,i):=\int\Sigma_{0}\left(x,w,x_{i},w_{i}\right)p\left(w\right)dw$ for
$1\leq i\leq n$.
We also have that
\begin{equation}
\begin{split}
&\sigmatilde_{n}\left(x,x_{n+1},w_{n+1}\right) = \\
&\frac{B\left(x,n+1\right)-\left[B\left(x,1\right)\mbox{ }\cdots\mbox{ }B\left(x,n\right)\right]A_{n}^{-1}\gamma}{\sqrt{\Sigma_{0}\left(\!x_{n+1}\!,\!w_{n+1}\!,\!x_{n+1}\!,\!w_{n+1}\!\right)-\gamma^{T}A_{n}^{-1}\gamma+\lambda_{(x_{n+1},w_{n+1})}}}
\times 1{\left\{ \lambda_{\left(x_{n+1},w_{n+1}\right)}\!>\!0 \text{ or } \left(x_{n+1},w_{n+1}\right)\!\notin\!\{\left(x_{i},w_{i}) :  i\!\leq\!n\right\} \right\}},
\end{split}
\label{eq:post_var_formula}
\end{equation}
where $\gamma^{T}:=(\Sigma_{0}(x_{n+1},w_{n+1},x_{1},w_{1}),\ldots,\Sigma_{0}(x_{n+1},w_{n+1},x_{n},w_{n}))$.
\end{lemma}

The expressions in this lemma require that $\lambda_{(x_{n+1},w_{n+1})}$ be known to compute the value of information.
This is seldom true in practice, but this quantity can be estimated and the estimate used in its place.
If the noise is homogeneous then it can be estimated by including it as a hyperparameter in our Gaussian-process-based inference.
If each observation is an average of many i.i.d. replications, allowing the variance of the noise in each observation to be estimated with high accuracy,
and we believe that the noise does not change abruptly in the domain,
then we can use the mean of the variance estimates from previous observations as our estimator of $\lambda_{(x_{n+1},w_{n+1})}$.
Finally, if we are in neither of these situations, then we can use the approach developed in \citet{Kersting2007} in which a Gaussian process is used to estimate the variance of heteroscedastic noise across the domain.

We now use \Cref{postdist} to estimate the value of information and its gradient in the next subsection.

% As we already mentioned, there is an alternative approach with severe computational limitations, which depends on a discretization of the feasible set $A$, over which we take the maximum in \eqref{eq:VOI_a}, into $L<\infty$ points. Although the method is computationally infeasible to use when the dimension of $A$ is not small, we present it in $\mathsection$\ref{sec:discretization}, because we analyze the complexity of the algorithm and give theoretical guarantees for its use.
% Although this approach can only be used when the dimension of the domain is very small, we present it in $\mathsection$\ref{sec:discretization}, because we believe that it gives insights of the IBO policy, and it can be very fast if the dimension of the domain is one or two. Furthermore, instead of having unbiased stochastic approximations obtained using Monte Carlo estimators, they are deterministic and biased, and they too converge to the real values as $L$ goes to infinity if we use uniform discretizations. 

% ADD: AS SUBSECTION (i think that it's done)

\subsection{Discretization-Free Computation of the Value of Information and its Gradient}
\label{sec:monte_carlo}

In this subsection, we provide unbiased and strongly consistent Monte Carlo estimators of the value of information and its gradient. Our techniques use the envelope theorem \citep{milgrom} and were inspired by \citet{wugradientsbo}, which uses this theorem to build an unbiased estimator of the gradient of the knowledge-gradient in a different setting.
%We first introduce a novel technique ($\mathsection$\ref{sec:monte_carlo}), which is based on the envelope theorem \citep{milgrom} and a Monte Carlo approximation, used to obtain unbiased estimators of the value of information and its gradient, such that they converge almost surely to the real value of information and its gradient when the number of samples, used in the Monte Carlo approximation, tends to infinity. 

First, \Cref{postdist}, equation (\ref{eq:VOI_a}), and the strong law of large numbers show that if $\left\{ Z_{i}\right\} _{i=1}^{\infty}$ are independent standard
normal random variables, then 
\[
\widehat{V}_{n,m}\left(x,w\right):=\frac{1}{m}\sum_{i=1}^{m}\left[\mbox{max}_{x'\in A}\left(a_{n}\left(x'\right)+\sigmatilde_{n}\left(x',x,w\right)Z_{i}\right)-\mbox{max}_{x'\in A}a_{n}\left(x'\right)\right]
\]
is an unbiased and strongly consistent estimator of the value of information $V_{n}(x,w)$.  
The inner optimization problems $\mbox{max}_{x'\in A}\left(a_{n}\left(x'\right)+\sigmatilde_{n}\left(x',x,w\right)Z_{i}\right)$ can be solved using standard optimization methods such as LBFGS \citep{byrd:1997} or
Newton methods (if the Hessian of the kernel exists). Gradients of the inner optimization problem's objective can
be computed using (\ref{eq:200}). 

We now build an unbiased and strongly consistent estimator of $\frac{\partial}{\partial r}V_{n}\left(x,w\right)$ where $x=\left(x_{1},\ldots,x_{d}\right),w=\left(w_{1},\ldots,w_{p}\right)$, $r\in\left\{ x_{i}:1\leq i\leq d\right\} \bigcup\left\{ w_{i}:1\leq i\leq p\right\} $, and give sufficient conditions for existence of $\frac{\partial}{\partial r}V_{n}\left(x,w\right)$. We use the envelope theorem \citep{milgrom}, along with the following lemma, which shows some smoothness properties of $a_{n}$ and $\sigmatilde_{n}$. The proof of this lemma may be found in $\mathsection$\ref{proofs_monte_carlo_sect}.

\begin{lemma}
\label{smooth_ibo}

We assume $\mu_{0}$ is constant, and the kernel $\Sigma_{0}$ of the prior distribution on $F$ is continuously differentiable and bounded. We also suppose there is a non-negative function $h$ such that $\int h(x,w',x')p(w')dw'$ is finite for all $x,x'\in A$, and $\left|\frac{\partial\Sigma_{0}\left(x,w',x',w\right)}{\partial w}\right|<h(x,w',x')$ for all $x,x'\in A$ and $w,w'\in W$. Then:
\begin{enumerate}
\item $a_{n}$ and $\sigmatilde_{n}\left(\cdot,x,w\right)$ are both continuously differentiable for any $x,w$ if $\Sigma_{n}(x,w,x,w) > 0$.
\item For any $x'$, $\sigmatilde_{n}\left(x',x,w\right)$ is continuously differentiable with respect to  $x,w$ if $\Sigma_{n}(x,w,x,w) > 0$ and $\lambda_{(x,w)}$ is continuously differentiable.
\end{enumerate}
 
\end{lemma}

The condition $\Sigma_{n}(x,w,x,w) > 0$ in the previous lemma is always true in the noisy case, as shown by \eqref{post_cov_F}. In the noiseless case,  $\Sigma_{n}(x,w,x,w)$ can be zero only if $x,w$ is a previously measured point.

The following lemma shows how to compute stochastic gradients of $V_{n}\left(x,w\right)$ and allows us to optimize $V_{n}$ with a stochastic gradient ascent method.

\begin{lemma}
\label{monte_carlo_voi}

Suppose that the hypotheses of \cref{smooth_ibo} on $\Sigma_0$ and $p$ are satisfied. Also assume that for a given $\left(x_{n+1},w_{n+1}\right)$, $\mbox{argmax}_{x'\in A}\left(a_{n}\left(x'\right)+\sigmatilde_{n}\left(x',x_{n+1},w_{n+1}\right)Z\right)$
is  almost surely a singleton, where $Z$ is a standard normal random variable.
Also assume
$\Sigma_{n}(x_{n+1},w_{n+1},x_{n+1},w_{n+1}) > 0$ and 
$\lambda_{(x,w)}$ is continuously differentiable at $x=x_{n+1}$ and $w=w_{n+1}$.
Let $\left\{ Z_{i}\right\} _{i=1}^{\infty}$ be independent standard normal random variables. Then
\[
\nabla V_{n}\left(x_{n+1},w_{n+1}\right)=\lim_{m\rightarrow\infty}\frac{1}{m}\sum_{i=1}^{m}\left(\nabla_{\left(x_{n+1},w_{n+1}\right)}\sigmatilde_{n}\left(y_{i},x_{n+1},w_{n+1}\right)Z_{i}\right) \mbox{ a.s.,}
\]
where $y_{i}=\mbox{argmax}_{x'\in A}\left(a_{n}\left(x'\right)+\sigmatilde_{n}\left(x',x_{n+1},w_{n+1}\right)Z_{i}\right)$. Furthermore, $E\left[\left(\nabla_{\left(x_{n+1},w_{n+1}\right)}\sigmatilde_{n}\left(y_{i},x_{n+1},w_{n+1}\right)Z_{i}\right)\right]=V_{n}\left(x_{n+1},w_{n+1}\right)$ for all $i$.

\end{lemma}

\proof{}

Let $Z$ be a standard normal random variable, and $f\left(x',\left(x,w\right)\right):=a_{n}\left(x'\right)+\sigmatilde_{n}\left(x',x,w\right)Z$ where $x',x\in A$ and $w\in W$. By \cref{smooth_ibo}, $f$ is continuously differentiable, and so
by the envelope theorem \citep[Corollary 4]{milgrom}, 
\begin{equation*}
\nabla_{\left(x_{n+1},w_{n+1}\right)}f\left(y,\left(x_{n+1},w_{n+1}\right)\right)=\nabla_{\left(x_{n+1},w_{n+1}\right)}\left(a_{n}\left(y\right)+\sigmatilde_{n}\left(y,x_{n+1},w_{n+1}\right)Z\right)=\nabla_{\left(x_{n+1},w_{n+1}\right)}\sigmatilde_{n}\left(y,x_{n+1},w_{n+1}\right)Z\mbox{ a.s.}
\end{equation*} 
where  $y=\mbox{argmax}_{x'\in A}\left(a_{n}\left(x'\right)+\sigmatilde_{n}\left(x',x_{n+1},w_{n+1}\right)Z\right)$ and the dependence of $y$ on $x_{n+1}$,$w_{n+1}$ is ignored when taking the gradient.

We now show that $\nabla V_{n}\left(x_{n+1},w_{n+1}\right)=E_{n}\left[\nabla_{\left(x_{n+1},w_{n+1}\right)}\mbox{max}_{x'\in A}\left(a_{n}\left(x'\right)+\sigmatilde_{n}\left(x',x_{n+1},w_{n+1}\right)Z\right)\right]$. First observe that if $z$ is a real number such that $\mbox{argmax}_{x'\in A}\left(a_{n}\left(x'\right)+\sigmatilde_{n}\left(x',x_{n+1},w_{n+1}\right)z\right)=\left\{y_{z}\right\}$, by the envelope theorem, we then have that,
\[
\nabla_{\left(x_{n+1},w_{n+1}\right)}\mbox{max}_{x'\in A}\left(a_{n}\left(x'\right)+\sigmatilde_{n}\left(x',x_{n+1},w_{n+1}\right)z\right)=\nabla_{\left(x_{n+1},w_{n+1}\right)}\left(a_{n}\left(y_{z}\right)+\sigmatilde_{n}\left(y_{z},x_{n+1},w_{n+1}\right)z\right).
\]
Furthermore, we have that 
\begin{eqnarray*}
\left|\nabla_{\left(x_{n+1},w_{n+1}\right)}\left(a_{n}\left(y_{z}\right)+\sigmatilde_{n}\left(y_{z},x_{n+1},w_{n+1}\right)z\right)\right| & = & \left|\nabla_{\left(x_{n+1},w_{n+1}\right)}\left(\sigmatilde_{n}\left(y_{z},x_{n+1},w_{n+1}\right)z\right)\right| \leq L_{\left(x_{n+1},w_{n+1}\right)}\left|z\right|
\end{eqnarray*}
where $L_{\left(x_{n+1},w_{n+1}\right)}=\mbox{sup}_{y\in A}\left|\nabla_{\left(x_{n+1},w_{n+1}\right)}\sigmatilde_{n}\left(y,x_{n+1},w_{n+1}\right)\right|$, which is finite because $\sigmatilde_{n}\left(\cdot,x_{n+1},w_{n+1}\right)$ is continuously differentiable
and $A$ is a compact set. Consequently, $E_{n}\left[L_{\left(x_{n+1},w_{n+1}\right)}\left|Z\right|\right]<\infty$.

By Corollary 5.9  of \citet{bartle}, $\nabla V_{n}\left(x,w\right)=E_{n}\left[\nabla_{\left(x_{n+1},w_{n+1}\right)}\mbox{max}_{x'\in A}\left(a_{n}\left(x'\right)+\sigmatilde_{n}\left(x',x_{n+1},w_{n+1}\right)Z\right)\right]$. Finally, the first claim of the lemma follows from the strong law of large numbers.

\endproof
% That stochastic gradient is an unbiased estimator of $\nabla V_{n}$, and it is variance goes to zero as the number of samples $M$ goes to infinity. 

As a reminder, we assumed that the objective function $G$ has the integrated form at the beginning of the section. In the case of the finite sum \eqref{eq:goal1}, the assumptions we have made in \cref{monte_carlo_voi} and \cref{smooth_ibo} may no longer hold.  
In particular, $\frac{\partial}{\partial w_{j}}\Sigma_{0}\left(x,w,x',w\right)$ 
will typically not exist for $1\leq j\leq p$ where $w=(w_{1},\ldots,w_{p})$, and so $\frac{\partial}{\partial w_{j}}V_{n}(x,w)$ may not exist either (see \cref{postdist}). 
However, our approach remains applicable in this setting:
we use $\nabla_{x} V_{n}(x,w)$ to maximize $V_{n,x}(x,w)$ for each $w\in W$, and then easily solve $\mbox{max}_{w\in W}V_{n}^{*}\left(w\right)=\mbox{max}_{w\in W}\left[\mbox{max}_{x\in A}V_{n}\left(x,w\right)\right]$ observing that $W$ is a finite set.
Using $\nabla_{x} V_{n}(x,w)$ in this way requires showing similar results to the ones presented in this section to compute stochastic gradients of $V_{n}(x,w)$ with respect to $x$ for any fixed $w$, under the assumption that $\Sigma_{0}(\cdot, w, \cdot, w')$ and $\lambda_{(\cdot,w)}$ are sufficiently smooth for any $w,w'\in W$.  We do not include these results here, because their proofs follow the same ideas already presented.

\subsection{Computation and Complexity of the BQO Algorithm}
\label{sec:IBO-alg}

In this section, we summarize computation of the BQO algorithm, which combines the tools developed in previous sections, and discuss its complexity.
First, recall we previously described methods for obtaining unbiased samples of $V_n$ and $\nabla V_n$ using a fixed value of $\theta$.
Because $\theta$ was fixed, we suppressed it in our notation, but here we indicate it explicitly, writing these values as $V_n(\theta)$ and $\nabla V_n(\theta)$ and their estimators as $\widehat{V}_n(\theta)$ and $\widehat{\nabla V}_n(\theta)$ respectively.  We will use these within a stochastic gradient 
algorithm, ADAM \citep{adam}, for solving problem \eqref{eq:max_VOI}, i.e., for maximizing $E[V_n(\theta) \mid H_n]$. 
Within this stochastic gradient algorithm,
each stochastic gradient is obtained by first taking $J$ independent samples $\widehat{\theta}_j : j=1,\ldots,J$ from the posterior distribution on $\theta$ given $H_n$ using slice sampling, and then computing $\frac{1}{J}\sum_{j=1}^J \widehat{\nabla V}_n(\widehat{\theta}_j)$, where each $\widehat{\nabla V}_n(\theta)$ uses a single independent standard normal random variable (so $m=1$ as defined \Cref{monte_carlo_voi}).
We use a similar approach in the final step of our algorithm to select a point with maximal posterior mean $E[a_N(x;\theta) | H_N]$, except that the only source of randomness in our stochastic gradient estimator is $\theta$.

The BQO algorithm using this approach is summarized in  \cref{alg:SBO}.

\begin{algorithm}[!htb]
   \caption{IBO Algorithm}
   \label{alg:SBO}
\begin{algorithmic}[1]
	\STATE Evaluate $F$ at $n_0$ points, chosen uniformly at random from $A\times W$.
% 	Use maximum likelihood or maximum a posteriori estimation to fit the parameters of the GP prior on $F$, conditioned on these $n_0$ samples. Let $\mu_0$ and $\Sigma_0$ be the mean function and covariance kernel of the resulting GP posterior on $F$.
	\FOR{$n=0$ {\bfseries to} $N-1$}
	\STATE Use the multi-start ADAM algorithm to maximize $E[V_n(\theta) \mid H_n]$,
	using the stochastic gradient estimator $\frac{1}{J} \sum_{j=1}^J \widehat{\nabla V}_n(\widehat{\theta}_j)$ on each iteration using independent samples $\widehat{\theta}_j$ from the posterior on $\theta$.
	Include all $n_0 + n$ samples of $F$ when computing the posterior.
	% The computations of the stochastic gradient $\nabla V_{n}$ and $V_{n}$ \stedit{given $\theta$} are described in $\mathsection$\ref{sec:VOI}. Those computations use all samples from the first stage, and samples $x_{1:n}$,$\w_{1:n}$, $y_{1:n}$.  
	Let $\left(x_{n+1},\w_{n+1}\right)$ be the resulting maximizer. 
% 	in  Update our Gaussian process posterior on $F$ using all samples from the first stage, and samples $x_{1:n}$,$\w_{1:n}$,						$y_{1:n}$.  This allows computation of $\mu_n$ and $\Sigma_n$ as described above, computation of $a_{n}$ through 			\eqref{eq:a_n}, and computation of $V_{n}$ and $\nabla V_{n}$ as described in $\mathsection$\ref{sec:VOI}.
	\STATE Sample $F\left(x_{n+1},\w_{n+1}\right)$ to obtain $y_{n+1}$.
	\ENDFOR

\STATE 
Use the multi-start ADAM algorithm to maximize $E[a_N(x;\theta) | H_N]$ using the stochastic gradient estimator 
$\frac{1}{J'}\sum_{j=1}^{J'} \nabla a_N(x;\widehat{\theta}_j)$, using independent samples 
$\hat{\theta}_j$ from the posterior on $\theta$. Return $x^{*}\in\mbox{argmax}_{x\in A} E[a_{N}\left(x;\theta\right) \mid H_N]$.
\end{algorithmic}
\end{algorithm}

Observe that in the noise-free case, $V_{n}\left(x,w\right)$ is not differentiable at any previously evaluated point $x,w$, as shown by the last equation of \cref{postdist}.
The set of previously evaluated points, however, is finite and so we can still use ADAM by perturbing the algorithm's current iterate whenever it resides at a non-differentiable point. A similar idea can be found in \citealt{jordan:saddle}.

Finally we discuss BQO's time and space complexity, assuming we use it to select $N$ points $(x,\w)$ to sample. To select each point $(x,\w)$ to sample, we use \Cref{alg:SBO}, which runs the ADAM algorithm for $T$ iterations.  Each iteration requires a stochastic gradient computed using $J$ independent standard Gaussian random variables, $J$ independent samples from the posterior on $\theta$ (let $Q$ be the number of iterations of slice sampling used for each), and $J$ runs of LBFGS to maximize 
$\left(a_{n}\left(x'\right)+\sigmatilde_{n}\left(x',x_{n+1},w_{n+1}\right)Z_{i}\right)$. 
Let $K$ be the number of steps in a single run of LBFGS, where each step requires an evaluation of $a_n$, $\sigmatilde_n$, and their gradients. Let $O(L)$ be the complexity of computing the kernel and its gradient, and let $O(S)$ be the complexity of computing (\ref{eq:200}),$\nabla_{n+1}B\left(x,n+1\right)$, and $B(x,i)$ for all $i\leq n$.

With this notation, we show in the appendix ($\mathsection$\ref{bqo_complexity}) that the BQO algorithm has time complexity $O(JQN^4 + JQLN^3 + JTK(SN^2 + N^3)+JTLN^2)$ and space complexity $O(N^2)$.

The integrals in (\ref{eq:200}), $\nabla_{n+1}B\left(x,n+1\right)$, and $B(x,i)$ do not necessarily have closed-form expressions.
While we might estimate them via Monte Carlo or numerical integration, this can be inconvenient and increase computational cost.
Consequently it may be better to first perform a change of variables from $w$ to another $w'$ for which integrals may be evaluated in closed form.
One such transformation, to the Gaussian distribution, is discussed in the appendix $\mathsection$\ref{gaussian_case}.
In addition, a change of variables from $w$ to $w'$ induces a change from $F(x,w)$ to some other $F'(x,w')$, which might change more slowly with $w'$ (requiring fewer samples to model it) or be better modeled by a Gaussian process. We illustrate this change of variable technique in our numerical experiments, in the inventory ($\mathsection$\ref{sec:IPexample}) and Citi Bike ($\mathsection$\ref{sec:citibike}) problems. 
%Furthermore, in the e-companion $\mathsection$\ref{gaussian_case}, we discuss transformation to a problem in which $w'$ follows a Gaussian distribution.  This Gaussian transformation is especially useful because it allows performing the integrals in (\ref{eq:200}), $\nabla_{n+1}B\left(x,n+1\right)$, and $B(x,i)$ in closed form when using a squared exponential kernel.

\subsection{Discretized Computation of the Value of Information and its Gradient}
\label{sec:discretization}

In this section we describe an alternate approach to that in $\mathsection$\ref{sec:monte_carlo} for estimating the value of information and its gradient.  It uses a discretization of $A$.
Although this approach was already considered in \citet{Xie:2012}, which presents a particular case of our method for the integral objective (\ref{eq:goal2}) in the noise-free setting with an independent Gaussian $p$ and Gaussian kernel, we extend this analysis by generalizing it to our setting and showing that a sequence of increasingly fine discretizations produces a sequence of estimators whose estimates converge to the value of information. Thus, these estimators, while biased, are strongly consistent. In practice, computational intractability limits this approach when $A$ has more than $3$ dimensions. This lack of scalability is also demonstrated by a complexity analysis we present at the end of this subsection. 

% We let $A'$ denote a discretization of $A$, so $A' \subseteq A$ and $|A'|=L$. For example, if $A$ is a hyperrectangle, then we may discretize it using a uniform mesh.

\begin{lemma}
\label{discretization_converge_past}
We assume that $\Sigma_{0}(\cdot,w,x',w
')$ is continuous for all $w,w'\in W$, $x'\in A$, $\Sigma_{0}$ is bounded, and $\mu_{0}$ is a constant. Suppose that we have an increasing sequence of finite discretizations $\left\{ A'_{L}\right\} _{L=1}^{\infty}$ of $A$, such that $\bigcup_{L=1}^{\infty}A'_{L}$ is dense in $A$. Then
\begin{align*}
V_n(x_{n+1},\w_{n+1})=\lim_{L\rightarrow\infty}\left(E_{n}\left[\max_{x\in A'_{L}} \left(a_{n}\left(x\right) + \sigmatilde(x, x_{n+1},\w_{n+1})Z_{n+1}\right)\right]-\max_{x \in A'_{L}} a_{n}\left(x\right)\right).
\end{align*}
\end{lemma}

The proof of this lemma may be found in $\mathsection$\ref{proofs_monte_carlo_sect}.  A
sequence of discretizations that satisfy the properties of the lemma can be built by considering the rationals, as we do in the proof of \cref{continuum_th}.

Using the previous lemma, we have that $V_n(x_{n+1},\w_{n+1})=\mbox{lim}_{L\rightarrow\infty}h(a_n(A'_{L}),\sigmatilde_n(A'_{L},x_{n+1},\w_{n+1}))$,
% It is easy to observe that if we consider an increasing sequence of uniform discretizations $\left\{ A'_{L}\right\} _{L=1}^{\infty}$ of $A$, such that $\bigcup_{L=1}^{\infty}A'_{L}$ is dense in $A$, and $a_{n},\sigmatilde_{n}(\ldot,x_{n+1},\w_{n+1})$ are both continuous in $A$, then
% \begin{align*}
% V_n(x_{n+1},\w_{n+1}) 
% &= E_{n}\left[\mbox{max}_{x\in A} a_{n}\left(x\right) + \sigmatilde(x, x_{n+1},\w_{n+1})Z_{n+1}\right]-\mbox{max}_{x \in A} a_{n}\left(x\right)\\
% &= \mbox{lim}_{L\rightarrow\infty}\left(E_{n}\left[\mbox{max}_{x\in A'_{L}} a_{n}\left(x\right) + \sigmatilde(x, x_{n+1},\w_{n+1})Z_{n+1}\right]-\mbox{max}_{x \in A'_{L}} a_{n}\left(x\right)\right)\\
% % &=h(a^n,\sigmatilde_n(x_{n+1},\w_{n+1}))\\
% &=\mbox{lim}_{L\rightarrow\infty}h(a_n(A'_{L}),\sigmatilde_n(A'_{L},x_{n+1},\w_{n+1})),
% \end{align*}
where 
$a_{n}(A'_{L})=\left(a_{n}\left(x_{i}\right)\right)_{i=1}^{L}$,
$\tilde{\sigma}_{n}\left(A'_{L},x,\w\right)=\left(\tilde{\sigma}_{n}\left(x_{i},x,\w\right)\right)_{i=1}^{L}$, and 
$h:\mathbb{R}^{L}\times\mathbb{R}^{L}\rightarrow\mathbb{R}$ is defined
by 
\begin{equation*}
h\left(a,b\right)=E\left[\mbox{max}_{i}a_{i}+b_{i}Z\right]-\mbox{max}_{i}a_{i},
\end{equation*}
where $a$ and $b$ are any deterministic vectors, and $Z$ is a one-dimensional
standard normal random variable. We can then approximate the value of information by $h(a_n(A'_{L}),\sigmatilde_n(A'_{L},x_{n+1},\w_{n+1}))$ for some $L$. By convenience, we denote $a_{n}\left(x_{i}\right)$ by $q_{i}$ and 
$\tilde{\sigma}_{n}\left(x_{i},x,\w\right)$ by $r_{i}$ for each $i$ in $\{1,\ldots,L\}$. If $A = A'_{L}$, which is possible if $A$ is a finite set, then the approximation is exact.

Algorithm 1 of \citet{frazier2009knowledge} applied to $h$, gives a subset of indexes
$\left\{ j_{1},\ldots,j_{\ell}\right\}$ from $\left\{ 1,\ldots,L\right\}$, such that
$V_n(x_{n+1},\w_{n+1})=h(a_n(A'_{L}),\sigmatilde_n(A'_{L},x_{n+1},\w_{n+1}))=\sum_{i=1}^{\ell-1}\left(r_{j_{i+1}}-r_{j_{i}}\right)f\left(-\left|c_{i}\right|\right)$, where $f\left(z\right):=\varphi\left(z\right)+z\Phi\left(z\right)$, $c_{i} :=  \frac{q_{j_{i+1}}-q_{j_{i}}}{r_{j_{i+1}}-r_{j_{i}}}$ for $1\leq i \leq \ell -1$, and $\varphi,\Phi$ are the standard normal cdf and pdf, respectively. This shows how to approximate the value of information $V_n$ using the discretization $A'_{L}$ of $A$.

We now show how to approximate the gradient of the value of information $V_n$ using the discretization $A'_{L}$ of $A$. Observe that if $\ell=1$, $V_{n}\left(x,\w\right)=0$ and so $\nabla V_{n}\left(x,\w\right)=0$. On
the other hand, if $\ell>1$, one can show via direct computation that $\nabla V_{n}\left(x,\w\right) = \sum_{i=1}^{\ell-1}\left(-\nabla r_{j_{i+1}}+\nabla r_{j_{i}}\right)\varphi\left(\left|c_{i}\right|\right)$.
Consequently, we only need to compute $\nabla r_{j_{i}}$ for each $i$ in $\{1,\ldots,\ell\}$ . Another direct computation shows that $\nabla_{(x_{n+1},\w_{n+1})}\tilde{\sigma}_{n}\left(x,x_{n+1},\w_{n+1}\right) =
\beta_{1} \beta_{3}-\frac{1}{2}\beta_{1}^{3}\beta_{2}\left[\beta_{5}-\beta_{4}\right]$, where 
\begin{eqnarray*}
\beta_{1} & =& \left[\Sigma_{0}\left(x_{n+1},\w_{n+1},x_{n+1},\w_{n+1}\right)-\gamma^{T}A_{n}^{-1}\gamma\right]^{-1/2},\\
\beta_{2} & = & B\left(x,n+1\right)-[B\left(x,1\right)\mbox{ }\cdots\mbox{ }B\left(x,n\right)]A_{n}^{-1}\gamma,\\
\beta_{3} & = & \left(\nabla B\left(x,n+1\right)-\nabla\left(\gamma^{T}\right)A_{n}^{-1}\left[B\left(x,1\right),\cdots,B\left(x,n\right)\right]^{T}\right),\\
\beta_{4} & = & 2\nabla\left(\gamma^{T}\right)A_{n}^{-1}\gamma,\\
\beta_{5} & = & \nabla\Sigma_{0}\left(x_{n+1},\w_{n+1},x_{n+1},\w_{n+1}\right).\\
\end{eqnarray*}

\paragraph*{Complexity of the Discretized Version of the BQO Algorithm.}
Here we discuss the time and space complexity of a version of BQO based on discretized computation of the value of information and its gradient.
We use the same notation and a similar analysis to that in the previous section. As a reminder, $O(L)$ is the complexity of the computation of the kernel and its gradient, and $O(S)$ is the complexity of computing $\nabla_{n+1}B\left(x,n+1\right)$, and $B(x,i)$ for all $i\leq n$. We sample $J$ parameters $(\widehat{\theta}_{1},\cdots,\widehat{\theta}_{J})$ from the posterior on $\theta$, running slice sampling for at most $Q$ iterations for each, and we optimize the value of information with the ADAM algorithm for at most $T$ iterations. If we use a uniform discretization of size $E^\dimension$ where $\dimension$ is the dimension of $A$, the time complexity of the BQO algorithm run for $N$ iterations is $O(E^\dimension TJ(SN^2 + N^3) + JTLN^2 + JQLN^3 + JQN^4)$ , and the space complexity is $O(N^2+E^\dimension)$. Thus if $E$ increases, the time complexity and space complexity increase exponentially.
This makes this method impractical when $d>3$.
% which is a severe problem when $x$ is high-dimensional because $E$ needs to be very large to provide a useful approximation to the value of information. 

\section{Asymptotic Analysis for BQO}
\label{sec:asymptotic}

In this section, we show consistency of BQO. We show that if $p$ and $W$ are finite, $A$ is finite or a closed box in $\mathbb{R}^{\dimension}$, the integrand function $F$ follows a Gaussian process prior with continuous paths for a fixed $w$, and the prior on the hyperparameters of the kernel is concentrated on a single value, then as the number of iterations of the algorithm tends to infinity, the optimal solution given by the BQO algorithm converges in expectation to an optimal solution $\mbox{argmax}_{x\in A}G\left(x\right)$. We omit the explicit dependence of the expressions on $\theta$ since the prior on $\theta$ is assumed concentrated on a single value.

We state two consistency results, one for continuum $A$ (Theorem~\ref{continuum_th_intuitive}) and the other for finite $A$ (Theorem~\ref{main_theorem_simplified}).  The proofs for both results may be found in the appendix.
% The proof for the finite case is easier to understand, and is presented in this section.   The proof for the continuum case is substantially more technical and may be found in the appendix.  
The proof for finite $A$ has a similar structure to the proof of consistency for the knowledge-gradient method for finite domains in problems without integrated objectives from \citet{frazier2009knowledge}.  We present a proof for the finite case partly because finite $A$ arises in practice, and partly because it is substantially simpler than the continuum case and provides a starting point for understanding the continuum proof.  
% After the appendix we comment on why the proof techniques of \citet{frazier2009knowledge} cannot be directly translated to the continuum setting, necessitating a new proof.
Our proof for the continuum goes substantially beyond the techniques required for the finite case, and develops techniques that may also be useful for proving consistency of other Bayesian optimization methods in continuum settings. Consistency of Bayesian optimization in continuum settings has been largely unexplored, with the authors being aware of only two other papers on this topic:  The working paper \citet{bect2016} contains consistency results for Bayesian optimization algorithms over Gaussian processes with continuous paths in the continuum setting; and \citet{bull2011} proved consistency of expected improvement for functions that belong to the reproducing kernel Hilbert space (RKHS) of the covariance function in the continuum setting, though Driscoll's Theorem \citep{beder2001} shows that, under some regularity conditions, sample paths of the Gaussian process almost surely do not belong to the RKHS.

% We denote the probability space by $\left(\Omega,\mathcal{F},P\right)$, and assume it is complete.

% \pfcomment{Does it make sense to discuss other papers that have proved consistency of BO algorithms in continuum settings, to point out that there are very few such papers?}

We first introduce notation needed for the theorems. Define $A'=A\times W$. Define the set $\mathcal{H}:=D\left(A'\right)\times D_{\mbox{kernel}}\left(A'\times A'\right)$,
where $D\left(A'\right)$ is the set of functions defined on $A'$, and $D_{\mbox{kernel}}\left(A'\times A'\right)$ is the set of positive semidefinite functions defined on $A'\times A'$. We define the set $\mathcal{H_{0}} \subset \mathcal{H}$ as

\[
\left\{ \left(\mu,\Sigma\right):\mu\equiv0, \Sigma_{w,w'}\left(x,y\right):=\Sigma\left(x,w,y,w'\right)\mbox{ is in }C^{1}\left(\mathbb{R}^{\dimension}\times\mathbb{R}^{\dimension}\right) \mbox{ and is isotropic for
all } w,w'\in W\right\}.
\]

We first state our result for continuum $A$ and prove it in the appendix $\mathsection$\ref{proof_continuum_case},
\begin{theorem}
\label{continuum_th_intuitive}
Suppose that $A=\left[a_{1},b_{1}\right]\times\cdots\times\left[a_{\dimension},b_{\dimension}\right]\subset\mathbb{R}^{\dimension}$,
$a_{i}<b_{i}$ for all $i$, $W$ is a finite set, and the probability space is complete. 
Assume $(\mu_0,\Sigma_0) \in \mathcal{H_{0}}$.
We assume that the function
$g_{w}\left(x\right):=\lambda_{\left(w,x\right)}$ is continuous in
$A$ for all $w\in W$, and there exists $k_{\lambda},K_{\lambda}>0$
such that $k_{\lambda}<\lambda_{\left(x,w\right)}<K_{\lambda}$ for
all $w\in W$ and $x\in A$. Then 
\[
\lim_{N}\mbox{ }E^{BQO}\left[\max_{x\in A}a_{N}\left(x\right)\right] = 
E\left[\max_{x\in A}G\left(x\right)  \right],
\]
where $E^{BQO}$ indicates expectation with respect to the distribution over sampling decisions induced by BQO.
\end{theorem}

We also state our result for finite $A$ and prove it in the appendix $\mathsection$\ref{proof_finite_case},

\begin{theorem}
\label{main_theorem_simplified}
Suppose that $A$ and $W$ are finite. 
Assume $(\mu_0,\Sigma_0) \in \mathcal{H}$.
We have that
\[
\lim_{N}\mbox{ }E^{BQO}\left[\max_{x\in A}a_{N}\left(x\right)\right] = 
E\left[\max_{x\in A}G\left(x\right)\right].
\]
\end{theorem}

\section{Numerical Experiments}
\label{experiments}

In this section we present numerical experiments motivated by applications in operations research and machine learning. We compare the BQO algorithm against baseline Bayesian optimization algorithms and algorithms from the literature designed for the specific problems considered. These experiments illustrate how the BQO algorithm can be applied in practice, and demonstrate it performs at least as well as these benchmarks on all problems considered, and often much better. 
%We can observe in practice that the IBO acquisition function has many local optimums, and consequently we need to use several different starting points to optimize the acquisition function, which can be problem, due to computational resources, when $W$ is a finite set without a metric, because we need to optimize the acquisition function for each possible value of $w$. We believe that is the reason that our algorithm did not outperform one of the baselines only in the cross validation problems. \stcomment{I'm running IBO again on the CNN problem again to see if we get better results, and that statement is true}

We compare on seven test problems: a test problem with a simple analytic form ($\mathsection$\ref{sec:test}); a composition of Branin functions ($\mathsection$\ref{sec:branin}) used in \citet{williams2000sequential}; a realistic problem arising in the design of the New York City's Citi Bike system ($\mathsection$\ref{sec:citibike}); cross-validation of convolutional neural networks ($\mathsection$\ref{sec:CVexample_cnn}) and recommendation engines ($\mathsection$\ref{sec:CVexample_filtering}); an inventory problem with substitution ($\mathsection$\ref{sec:IPexample}); and a collection of problems simulated from Gaussian process priors ($\mathsection$\ref{sec:GPexample}) that provide insight into how the benefit provided by BQO is determined by problem characteristics, and that identify problems where BQO is most helpful.

As benchmark algorithms we consider the multi-task algorithm in Section 3.2 of \citet{swersky2013multi} and the algorithm of \citet{williams2000sequential}, which both place a Gaussian process prior on $F(x,\w)$, as BQO does. In addition, we consider two baseline Bayesian optimization algorithms: the knowledge-gradient (KG) policy of \citet{frazier2009knowledge} and  the Expected Improvement criterion of \citet{jones1998efficient}, which both place the Gaussian process prior directly on $G(x)$. The KG policy is equivalent to BQO in problems where all components of $\w$ are moved into $x$. We also solved the problems from $\mathsection$\ref{sec:test} and $\mathsection$\ref{sec:citibike} with Probability of Improvement (PI) \citep{brochu2010tutorial}, but do not include these results because both KG and EI significantly outperform PI.

We now discuss the kernels used in these experiments.
When implementing BQO and the benchmark algorithms in the Branin ($\mathsection$\ref{sec:branin}) and inventory ($\mathsection$\ref{sec:IPexample}) problems, we use the $5/2$-Mat\'ern kernel
$\Sigma_{0}\left(x,w,x',w'\right)=\sigma^{2}\left(1+\sqrt{5}r+\frac{5}{3}r^{2}\right)\mbox{exp}\left(-\sqrt{5}r\right)$,
where $r=\sqrt{\sum_{i=1}^{n}\alpha_{1}^{\left(i\right)}\left(x_{i}-x'_{i}\right)^{2}+\sum_{i=1}^{d_{1}}\alpha_{2}^{\left(i\right)}\left(w_{i}-w'_{i}\right)^{2}}$. 

% In order to keep comparisons fair, we use the $5/2-Mat\'ern$ kernel in the sequential design of experiments algorithm (SDE) of \citet{williams2000sequential}, and expected improvement. 

In the cross-validation problems ($\mathsection$\ref{sec:CVexample_filtering}, $\mathsection$\ref{sec:CVexample_cnn}), we use the expected improvement algorithm with the $5/2$-Mat\'ern kernel, and the BQO algorithm (\cref{alg:SBO} in $\mathsection$\ref{SBO})  and multi-task Bayesian algorithm with the task kernel \citep{swersky2013multi}, which is the Kronecker product of a  $5/2$-Mat\'ern kernel and a kernel defined only over the finite set $W$. Specifically this kernel is defined by 
$\Sigma_{0}\left(x,t,x',t'\right)=\sigma_{t,,t'}\left(1+\sqrt{5}r+\frac{5}{3}r^{2}\right)\mbox{exp}\left(-\sqrt{5}r\right)$
where $r=\sqrt{\sum_{k=1}^{n}\alpha^{\left(k\right)}\left[x_{k}-x'_{k}\right]^{2}}$, $n$ is the number of tasks, and $\left\{ \sigma_{t,t'}\right\} _{t,t'\in\left\{ 1,\ldots,n\right\} }$ are real numbers, such that $\sigma_{t,t'} = \sigma_{t_{1},t'_{1}}$ whenever $t_{1} \neq t'_{1}$ and $t\neq t'$, and
%\pfdelete{$\left(\begin{array}{ccc}
%\sigma_{1,1} & \cdots & \sigma_{1,n}\\
%\vdots & \ddots & \vdots\\
%\sigma_{n,1} & \cdots & \sigma_{n,n}
%\end{array}\right)$}
the matrix $(\sigma_{t,t'} : t,t' \in \{1,\ldots,n\})$
is symmetric and positive definite.

% In those problems, we comparate our method against the fully Bayesian multi-task Bayesian optimization algorithm with the task kernel, and fully Bayesian expected improvement with the $5/2-Mat\'ern$ kernel.

In the analytic test problem ($\mathsection$\ref{sec:test}), Citi Bike problem ($\mathsection$\ref{sec:citibike}), and the problems simulated from Gaussian process priors ($\mathsection$\ref{sec:GPexample}), 
we implemented BGO
and the benchmark algorithms with the squared exponential kernel 
$\Sigma_{0}\left(x,\w,x',\w'\right)=
\sigma_{0}^{2}\mbox{exp}\left(-\sum_{k=1}^{n}\alpha_{1}^{\left(k\right)}\left[x_{k}-x'_{k}\right]^{2}-\sum_{k=1}^{d_{1}}\alpha_{2}^{\left(k\right)}\left[\w_{k}-\w'_{k}\right]^{2}\right)$,
where $\sigma_{0}^{2}$ is the common prior variance and $\alpha_{1}^{\left(1\right)},\ldots,\alpha_{1}^{\left(n\right)},\alpha_{2}^{\left(1\right)},\ldots,\alpha_{2}^{\left(d_{1}\right)}\in\mathbb{R}_{+}$
are the length scales parameters.

In the majority of our experiments 
($\mathsection$\ref{sec:branin} and 
$\mathsection$\ref{sec:CVexample_filtering}-$\mathsection$\ref{sec:IPexample})
we implement BGQO using the discretization-free approach with fully Bayesian inference over hyperparameters  (\cref{alg:SBO} in $\mathsection$\ref{SBO}).
In the analytic test problem ($\mathsection$\ref{sec:test}), the Citi Bike problem ($\mathsection$\ref{sec:citibike}), and the problems simulated from Gaussian process priors ($\mathsection$\ref{sec:GPexample}), 
we use the discretized version ($\mathsection$\ref{sec:discretization}) of the BQO algorithm.  In these problems we also calculate the hyperparameters of the kernels, $\sigma^{2}$ and $\mu_{0}$, using maximum likelihood estimation following the first stage of samples. We do not use the discretization-free version of BQO in these problems because they were performed as part of an initial conference paper version of this work \citep{MCQMC}, and only the discretized version of BQO existed at that time.
Benchmark algorithms in each problem were implemented using the same approach to hyperparameter estimation as used by BQO.

% In those problems, we compare the IBO algorithm against the knowledge-gradient and expected improvement algorithms with the squared exponential kernel. The hyperparameters of the kernels,  $\sigma^{2}$ and the mean $\mu_{0}$ are calculated using maximum likelihood estimation following the first stage of samples.

\subsection{An Analytic Test Problem}
\label{sec:test}
\begin{figure}[!htb]
\centering
\includegraphics[width=0.50\linewidth]{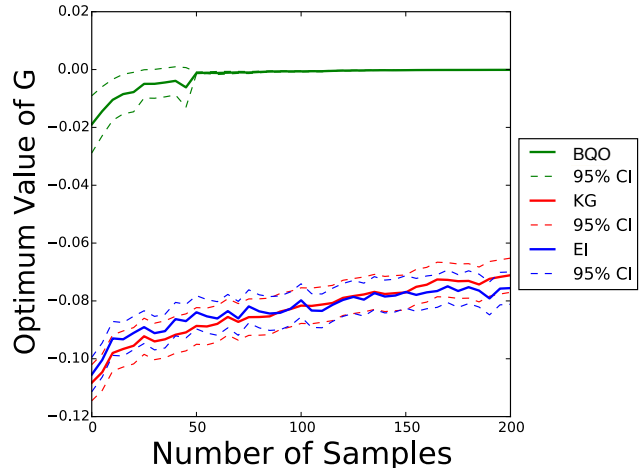}
\caption{Performance comparison between BQO and two Bayesian optimization benchmark, the KG and EI methods, on the analytic test problem \eqref{eq:test} from $\mathsection$\ref{SBO}, as described in $\mathsection$\ref{sec:test}.
\label{fig:analytic}} 
\end{figure}

In our first example, we consider the problem \eqref{eq:test} stated in $\mathsection$\ref{SBO}. BQO is well-suited to this problem because evaluations of $F(x,w)$ have much lower noise than those of $G(x)$. We do not compare against the multi-task algorithm \citep{swersky2013multi} and SDE algorithm \citep{williams2000sequential} because they can only be applied when the objective function is a finite sum. 
We do not compare against \citet{Xie:2012} because this problem has noisy evaluations.
Figure~\ref{fig:analytic} compares the performance of BQO, KG and EI on this problem, plotting the number of samples beyond the first stage on the $x$ axis, and the average true quality of the solutions provided, $G(\mathrm{argmax}_x E_n[G(x)])$, averaging over 3000 independent runs of the three algorithms. We see that BQO substantially outperforms both benchmark methods. This is because BQO reduces the noise in its observations by conditioning on $w$, allowing it to more swiftly localize the objective's maximum.

%\begin{figure}[!htb]
%	\includegraphics[width=0.9\linewidth]{comparisonSameConfigurationanal.pdf}
%    \caption{Performance comparison between SBO and two Bayesian optimization benchmark, the KG and EI methods, on the analytic test problem \eqref{eq:test} from $\mathsection$\ref{SBO}.  SBO performs significantly better than two benchmarks: knowledge-gradient (KG) and expected improvement (EI).
 %   \label{fig:tahi7}}
%\end{figure} 

\subsection{Branin Function}
\label{sec:branin}

In this example problem we compare BQO against the SDE \citep{williams2000sequential} and multi-task \citep[Section~3.2]{swersky2013multi} algorithms. We consider the Branin problem proposed in \citet{williams2000sequential} where $F\left(x_{1},x_{2},x_{3},x_{4}\right)=y_{b}\left(15x_{1}-5,15x_{2}\right)y_{b}\left(15x_{3}-5,15x_{4}\right)$, 

\[
y_{b}\left(u,v\right)=\left(v-\frac{5.1}{4\pi^{2}}u^{2}+\frac{5}{\pi}u-6\right)^{2}+10\left(1-\frac{1}{8\pi}\right)\cos\left(u\right)+10
\]
is the Branin function and $x_{1},x_{2},x_{3},x_{4}\in\left[0,1\right]$. We define $x:=\left(x_{1},x_{4}\right)$ and $w:=\left(x_{2},x_{3}\right)$. The joint distribution $p$ of $w$ is defined in \Cref{probability_dist_w}. We maximize the function $G\left(x_{1},x_{4}\right):=\sum_{\left(x_{2},x_{3}\right)\in \left\{ 0.25,0.5,0.75\right\} \times\left\{ 0.2,0.4,0.6,0.8\right\}  }p\left(x_{2},x_{3}\right)F\left(x_{1},x_{2},x_{3},x_{4}\right)$.

\begin{table}[htb]
\centering

\begin{tabular}{|l||r|r|r|c|}  
\hline
$x_{2} / x_{3}$ & 0.2 & 0.4 & 0.6 & 0.8\\
\hline\hline
0.25 & 0.0375 & 0.0875 & 0.0875 & 0.0375\\  
\hline
0.5 & 0.0750 & 0.1750 & 0.1750 & 0.0750\\
\hline
0.75 & 0.0375 & 0.0875 & 0.0875 & 0.0375\\
\hline
\end{tabular}

\caption{Probability distribution of $w=(x_{2},x_{3})$ for the Branin problem from $\mathsection$\ref{sec:branin}.}
\label{probability_dist_w}
\end{table}

\begin{figure}[!htb]
\centering
\includegraphics[width=0.50\linewidth]{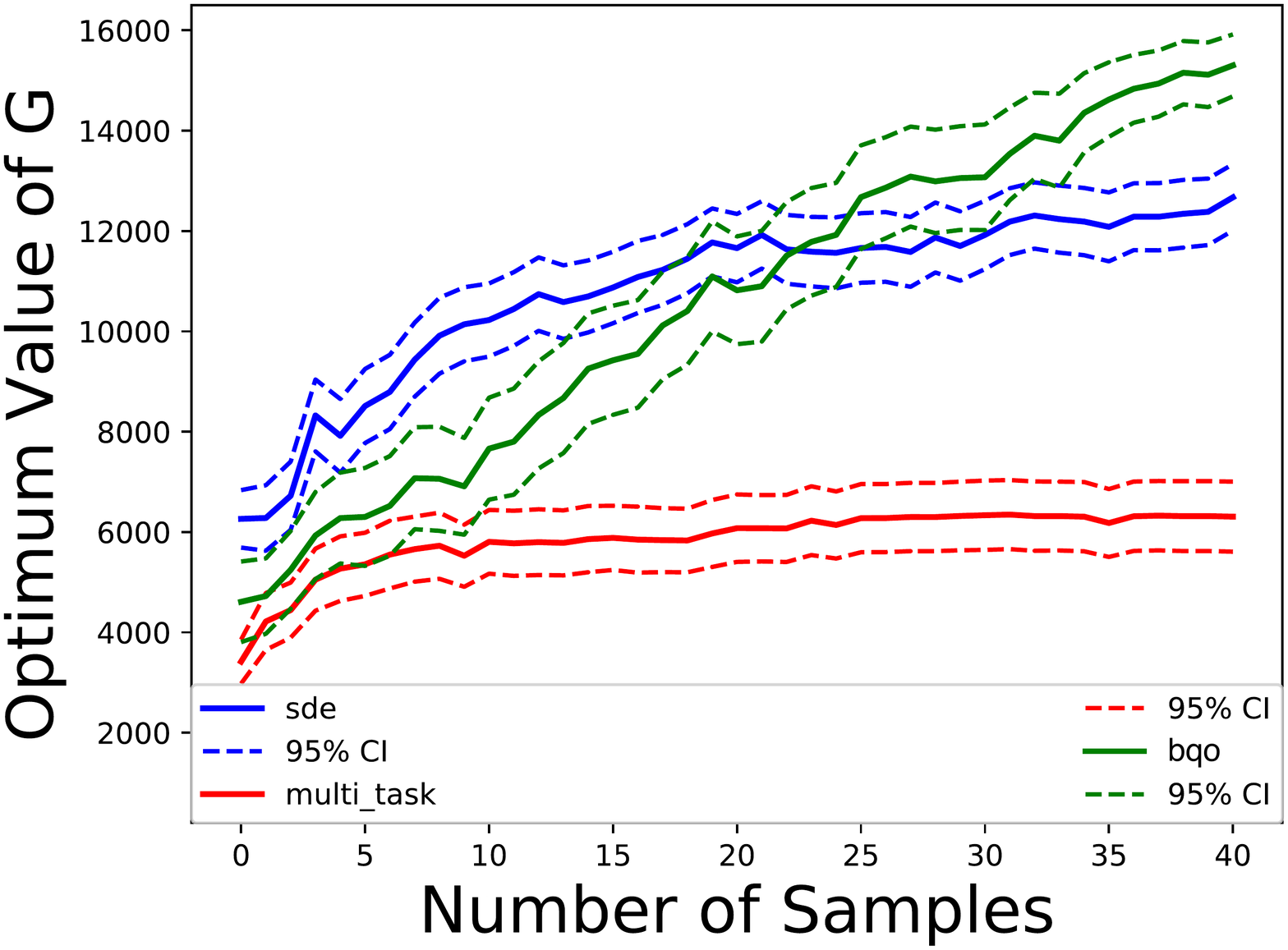}
\caption{Performance comparison between BQO, the SDE algorithm \citep{williams2000sequential}, and the multi-task algorithm \citep{swersky2013multi} on the Branin problem from $\mathsection$\ref{sec:branin}.\label{fig:branin}} 
\end{figure}

Figure~\ref{fig:branin} compares the performance of BQO, SDE and the multi-task algorithm on this problem, plotting the number of samples beyond the first stage on the $x$ axis, and the average true quality of the solutions provided, $G(\mathrm{argmax}_x E_n[G(x)])$. We average over 100 independent runs of the BQO algorithm, 126 independent runs of the  multi-task algorithm, and 230 independent runs of the SDE algorithm. We see that BQO substantially outperforms both the SDE and multi-task optimization benchmarks, despite the fact that these competing methods also model $F$. We believe this is because SDE and the multi-task optimization algorithm both choose points using a heuristic rule that performs poorly in certain settings, as explained in the introduction, rather than using a one-step optimality analysis like BQO.

\subsection{New York City's Citi Bike System}
\label{sec:citibike}

We now consider a queuing simulation based on New York City's Citi Bike system in which system users may remove an available bike from a station at one location within the city and ride it to a station with an available dock in some other location.  The optimization problem that we consider is the allocation of a constrained number of bikes (6000) to available docks within the city at the start of rush hour, so as to minimize, in simulation, the expected number of potential trips in which the rider could not find an available bike at their preferred origination station, or could not find an available dock at their preferred destination station.  We call such trips ``negatively affected trips.''

% In the last example, we simulated bike usage on the morning rush hours for New York City's bike sharing system. A bike sharing system is a group of bikes placed in different bike stations. People can rent these bikes and return them to any bike station after usage. 

We simulate the demand for bike trips on days from January 1st to December 31st between 7:00am and 11:00am. We use 329 actual bike stations, locations, and numbers of docks from the Citi Bike system. In our simulator, we choose a day at random from the 365 days of the year and then simulate the demand for trips between each pair of bike stations on that day using an independent Poisson process whose rate is given by historical data from that day in 2014 available from Citi Bike's website \citep{citibike}.
Travel times between pairs of stations are modeled using an exponential distribution with parameters estimated from this same dataset. If a potential trip's origination station has no available bikes, then that trip does not occur, and we increment our count of negatively affected trips.  If a trip does occur, and its preferred destination station does not have an available dock, then we also increment our count of negatively affected trips, and the bike is returned to the closest bike station with available docks.

% We simulate the demand for bike trips on days from January 1st to December 31st between 7:00am and 11:00am. We use 329 actual bike stations, locations, and numbers of docks from the Citi Bike system, and estimate demand and average trip time for each day in a year using data from 2014 available from Citi Bike's website \citep{citibike}. 

% This dataset  contains start bike station, end bike station, station latitude and longitude, and trip time for each bike trip. We considered $329$ bike stations and $600$ bikes. 

% We simulate the demand for trips between each pair of bike stations on a day using an independent Poisson process, and trip times between pairs of stations using an exponential distribution. 
% If a potential trip's origination station has no available bikes, then that trip does not occur, and we increment our count of negatively affected trips.  If a trip does occur, and its preferred destination station does not have an available dock, then we also increment our count of negatively affected trips, and the bike is returned to the closest bike station with available docks.

We divide the bike stations into $4$ groups using k-nearest neighbors, and let $x$ be the number of bikes in each group at 7:00 AM. We suppose that bikes are allocated uniformly among stations within a single group.  
%Then we consider a directed graph between the bike stations, where each pair of bike stations has two directed edges, and we divided these edges in $4$ groups. If the edge $(i,j)$ is in a group, then $(j,i)$ is also in that group. 
The random variable $\w$ is the total demand for bike trips during the period of our simulation, summed over all pairs of bike stations.  The distribution of $\w$ is a mixture of Poisson distributions. Evaluations of $F(x,\w)$ for $\w$ fixed are noisy due to additional sources  of randomness beyond $\w$ within our simulation.

We solve this problem with BQO, KG, EI and the multi-task algorithm. The multi-task algorithm cannot solve problems where the objective function is an infinite sum, as it is in this problem, so we modify the objective function it uses to a truncated expectation over finitely many values of $\w$. Because implementing the multi-task algorithm become computationally intractable when there are thousands of tasks, we restrict this truncated expectation to 181 values of $\w$. 

\begin{figure}[!htb]
    \centering
    \subcaptionbox{Performance comparison between BQO and two Bayesian optimization benchmark, the KG and EI methods, on the Citi Bike Problem from $\mathsection$\ref{sec:citibike} \label{fig:citibike}}[0.45\linewidth]{
      \includegraphics[width=0.45\linewidth]{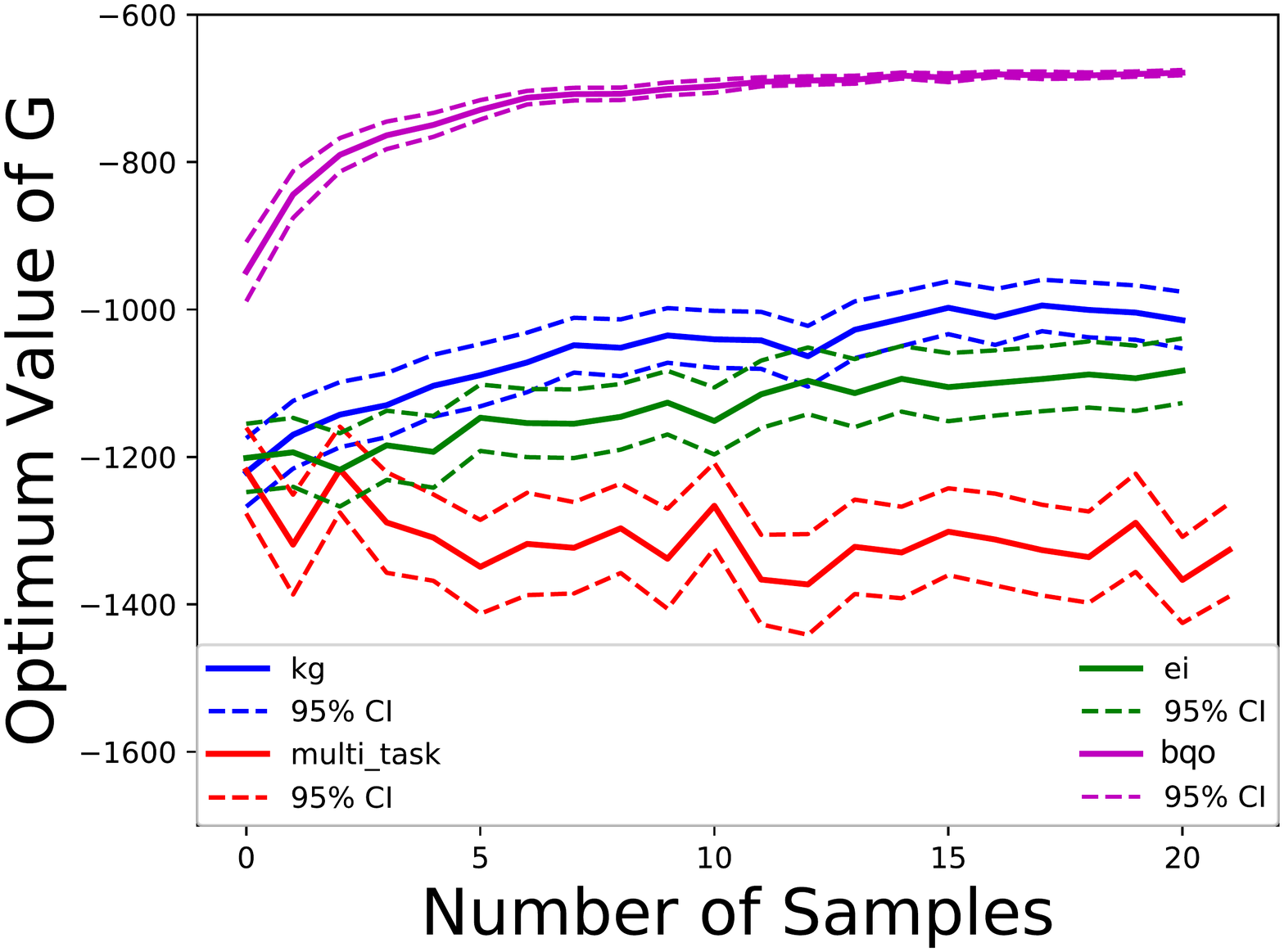}}
          \quad
     \subcaptionbox{Location of bike stations (circles) in New York City, where size and color represent the ratio of available bikes to available docks.}[0.45\linewidth]{
      \includegraphics[width=0.45\linewidth]{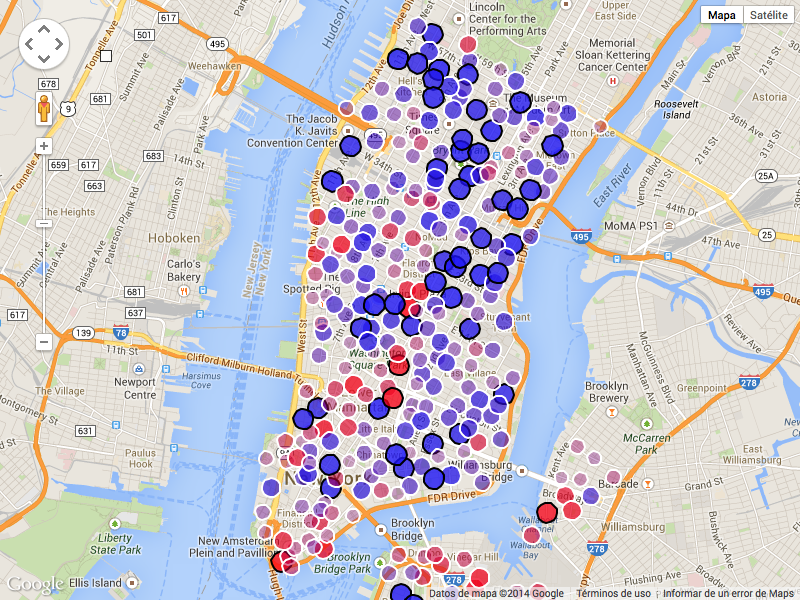}}
      \caption{Performance results for the Citi Bike problem (plot a), and a screenshot from our simulation of the Citi Bike problem (plot b), as described in $\mathsection$\ref{sec:citibike}.
    \label{fig:stuff}}

\end{figure}

Figure~\ref{fig:citibike} compares the performance of BQO, KG, EI and the multi-task algorithm, plotting the number of samples beyond the first stage on the $x$ axis, and the average true quality of the solutions provided, $G(\mathrm{argmax}_x E_n[G(x)])$, averaging over 300 independent runs of  BQO, EI and KG, and 100 independent runs of the multi-task algorithm. We see that BQO quickly finds an allocation of bikes to groups that attains a small expected number of negatively affected trips. We believe that multi-task does poorly because of the large number of tasks, and its inability to leverage information across related tasks. 

\subsection{Hyperparameter Tuning in Recommender Systems}
\label{sec:CVexample_filtering}

In this subsection and the following we consider optimization of a machine learning model's hyperparameters where error is evaluated using cross-validation. Cross-validation is a method for estimating a machine learning model's error.  In more detail, $n$-fold cross-validation randomly splits the training data into $n$ datasets of roughly equal size. 
Then, for each dataset (or ``fold''), it trains the machine learning model holding out that data, and evaluates the error of the resulting estimates on the held out data.
The average of these errors is called the cross-validation error, and is used as an objective in optimization of a machine learning model's hyperparameters.  In this approach, we minimize $\frac{1}{n}\sum_{i=1}^{n}L\left(x;D_{i}\right)$ over $x$, where $L\left(x;D_{i}\right)$ is the error of the model with hyperparameters $x$ evaluated on the $i$-th dataset $D_{i}$.

In this subsection we consider the problem of optimizing hyperparameters for probabilistic matrix factorization (PMF) models used in recommender systems \citep{mnih2008}. We apply this PMF model to a dataset from arxiv.org \citep{arxiv}, with information about downloads  from $2752$ users on $2018$ papers.
We treat a user as providing a positive binary rating for a paper if that user downloaded the paper, which creates $263,238$ positive binary ratings.

We use $5$-fold cross-validation to provide an estimate of the test error as a function of four PMF model hyperparameters: the learning rate, the $\ell_{2}$ regularizer, the number of epochs, and the matrix rank.  We then use  EI, BQO and the multi-task algorithm to choose these hyperparameters to minimize this cross-validation error.  EI simply selects a set of hyperparameters $x$ at each step and evaluates all 5 folds, while BQO and the multi-task algorithm select an $x$ and a fold $\w$.

\begin{figure}[!htb]
    \centering
    \subcaptionbox{Performance comparison between BQO and two Bayesian optimization benchmark, the multi-task and EI methods, on the recommender system problem $\mathsection$\ref{sec:CVexample_filtering}\label{fig:pmf}}[0.40\linewidth]{
      \includegraphics[width=0.40\linewidth]{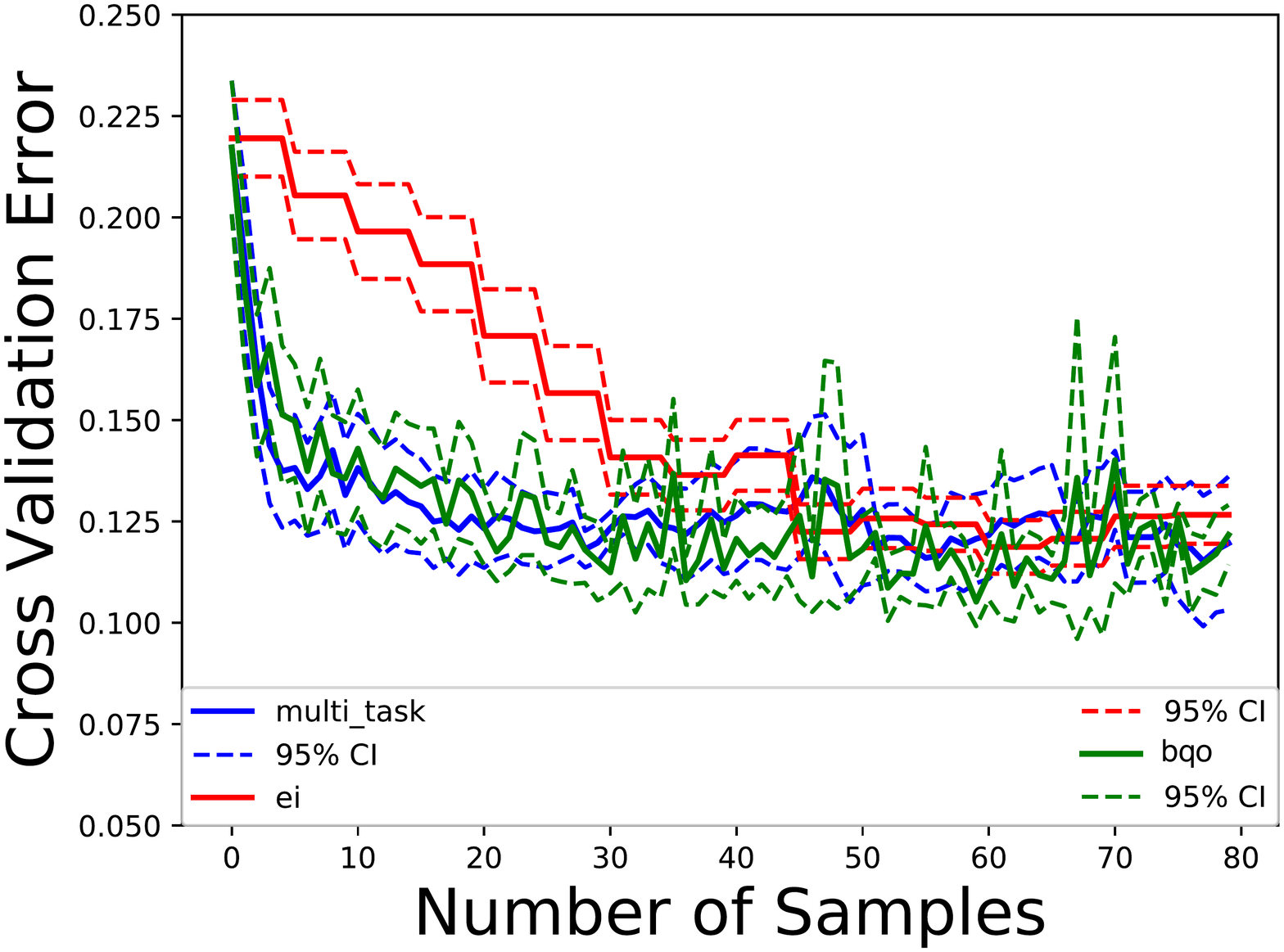}}
          \quad
     \subcaptionbox{Performance comparison between BQO and two Bayesian optimization benchmark, the multi-task and EI methods, on the convolutional neural network problem $\mathsection$\ref{sec:CVexample_cnn}\label{fig:cnn}}[0.40\linewidth]{
      \includegraphics[width=0.40\linewidth]{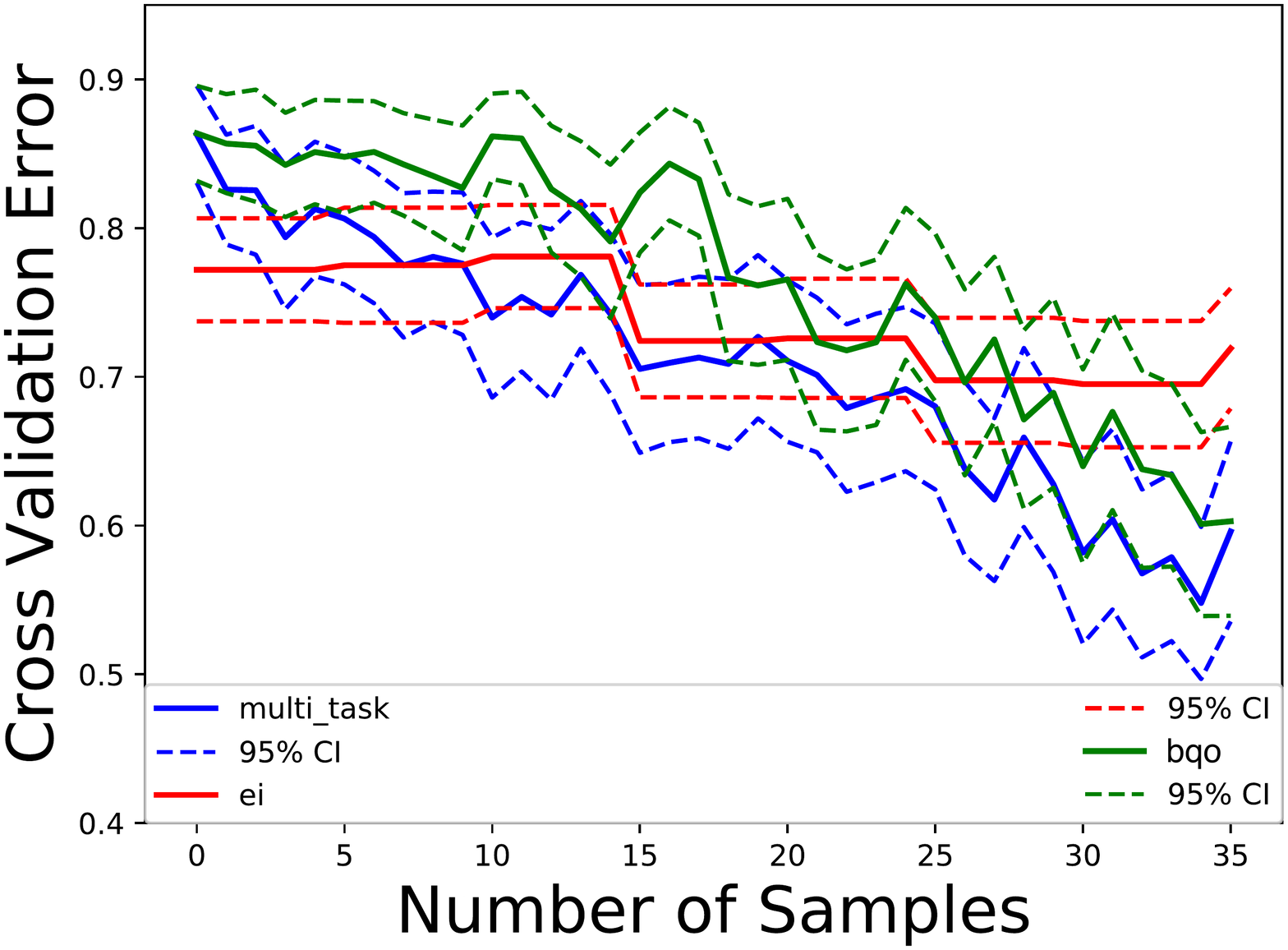}}
      \caption{Performance results for the recommender system (plot a) and convolutional neural network (plot b) problems. 
    \label{fig:stuff}}

\end{figure}

Figure~\ref{fig:pmf} compares the cross-validation error of these algorithms,  plotting the number of folds queried beyond the first stage on the $x$ axis, and the best error obtained, averaging over $35$ independent runs of BQO and multi-task, and $65$ of EI. We see that BQO and multi-task perform similarly, and both outperform EI.  We conjecture that multi-task's competitive performance in this problem is due to the small number of tasks and the homogeneity of the folds.

\subsection{Hyperparameter Tuning in Convolutional Neural Networks}
\label{sec:CVexample_cnn}

We consider the problem of training convolutional neural networks (CNNs) to classify images \citep{cnncifar}. We use $5$-fold cross validation on the CIFAR10 dataset \citep{cifar}, which consists of 10 classes and 50,000 training images. We choose the network architecture described in the pytorch tutorial \citep{pytorch}, which consists of two convolutional layers, two fully connected layers, and on top of them a softmax layer for final classification. We tune the following hyperparameters: the number of epochs, the batch size, the learning rate, the number of channels in the first convolutional layer, the size of the kernel in the convolutional layers, and the number of hidden units in the first fully connected layer. The number of channels in the second convolutional layer is the number of channels in the first convolutional layer plus 10,  and the number of hidden units in the second fully connected layer is 84.

Figure~\ref{fig:cnn} compares the performance of EI, BQO and the multi-task algorithm, plotting the number of folds queried beyond the first stage on the $x$ axis, and the best error obtained, averaging over $90$ independent runs of BQO and the multi-task algorithm, and $75$ of EI. We see that BQO and multi-task perform similarly, but better than EI.  As in the recommender system problem, we conjecture that multi-task is competitive because of the small number of tasks and their homogeneity.

\subsection{Newsvendor Problem under Dynamic Consumer Substitution}
\label{sec:IPexample}

The newsvendor problem under dynamic consumer substitution is adapted from \citet{ryzin2001Stocking}, and was considered in \citet{simopt}. In this problem, we choose the initial inventory levels of the products sold, each with given cost $c_{j}$ and price $p_{j}$.  Our goal is to optimize profit.

\newcommand{\cdf}{\Psi}
\newcommand{\anothercdf}{\Upsilon}

A sequence of $T$ customers indexed by $t$ arrive in order and either buy an in-stock product, or decide to not buy anything. Here, $T$ is known. Customer $t$ assigns a utility $U^{j}_{t}$ to each product $j$, and to the no-purchase option (indexed by $j=0$). Customer $t$ decides which product to buy, if any, by choosing the $j$ with the largest $U^j_t$ among the in-stock $j$ and the no-purchase option.
Utilities for products ($j>0$) are modeled with the multinomial logit model, where $U_{t}^{j}=u^{j}+\xi_{t}^{j}$, $u^{j}$ is constant, and
$\left\{ \xi_{t}^{j}\right\}$ are mutually independent Gumbel random
variables with distribution function $P\left(\xi_{t}^{j}\leq z\right)=\mbox{exp}\left(-e^{-\left(z/\mu+\gamma\right)}\right) =: \cdf^j_t(z)$
where $\gamma$ is Euler's constant.
The utility for the no-purchase option is $U_t^0 = 0$.
In this problem, the objective function $G(x)$ is defined as the expected overall profit considering the $T$ customers starting from a vector $x$ of initial inventory positions for each product. This profit is computed as sum of the prices of the products sold minus the cost of the initial inventory.

We consider the setting where there are $1000$ customers and $2$ products whose costs are $5$ and $10$ dollars respectively and prices are $8$ and $18$ dollars respectively. We assume that $u^{j}$ is equal to $1$ for all $j>0$.

We now describe how we use BQO in this problem. Fix a product $j$. Observe that since $\xi_{t}^{j}$ for $j>0$ follows a Gumbel distribution with cdf $\cdf_{t}^{j}$, then $\cdf_{t}^{j}\left(Z\right)$ is uniformly
distributed on $\left[0,1\right]$. Consequently, $\anothercdf^{-1}\left(\cdf_{t}^{j}\left(\xi_{t}^{j}\right)\right)$
follows a gamma distribution where $\anothercdf$ is the gamma cumulative distribution function.
Now define the vector $W:=\left(W_{1},W_{2}\right)$ where $W_{j}=\sum_{t=1}^{T}\anothercdf^{-1}\left(\cdf_{t}^{j}\left(\xi_{t}^{j}\right)\right)$.
It is straightforward to simulate $\xi = (\xi_t^j : t, j)$ given $W$, for example by noting that the distribution of $(\anothercdf^{-1}(\cdf^j_t(\xi^j_t)) / W_j : t \ge 1)$ is Dirichlet and independent of $W_j$ (see Theorem 2.1 of Section 2.1.2 of \citealt{dirichletbook}). Thus, simulating from this Dirichlet and multiplying by the given value of $W_j$ provides a sample of $\xi^j$ given $W_j$. Alternatively, we can use a simple modification of Example 10e in Section 10.2 of \citet{rossSimulation}.
We can also simulate $\xi^j$ given that $W^j$ resides in an interval by acceptance-rejection sampling, 
or by simulating $W^j$ from a truncated gamma distribution, and then simulating $\xi^j$ conditioned on $W^j$.

We then apply BQO with $F(x,w)$ equal to the conditional expectation of the profit given $W=\w$ and the initial inventory levels $x$ for each product.
To observe this conditional expectation we average results from $25$ independent simulations, where the collection of values for $\xi_t^j$ are simulated conditioned on $W$.

\newcommand{\median}{q_{1/2}}

We similarly apply the multi-task algorithm with $F(x,i)$ equal to the conditional expectation of the profit given the initial inventory level $x$ and that $W \in R_i$. 
Here, each $R_{i}$ is a rectangular region of values for $W$, given by  
$R_1 = [0,\median]^2$, $R_2 = (\median,\infty)^2$,
$R_3 = [0,\median]\times(\median,\infty)$, and
$R_4 = (\median,\infty)\times[0,\median]$, 
where $\median$ is the median of $W_j$ (this is the same for $j=1$ and $j=2$).
For each observation
of this conditional expectation we average $25$ independent simulations. The EI algorithm observes the profit without conditioning, averaging $25$ independent simulations.

\begin{figure}[!htb]
\centering
\includegraphics[width=0.50\linewidth]{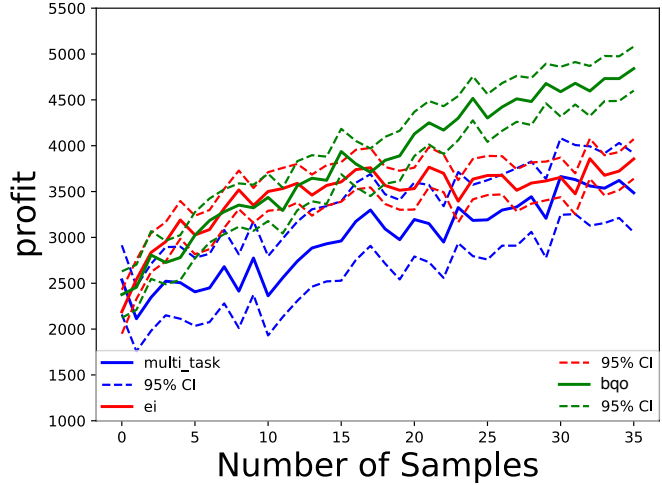}
\caption{Performance results for the vendor problem $\mathsection$\ref{sec:IPexample}\label{fig:vendor}} 
\end{figure}

In Figure~\ref{fig:vendor} we compare the performance of EI, BQO and the multi-task algorithm, plotting the number of samples beyond the first stage on the $x$ axis, and the best profit obtained, averaging over $100$ independent runs of BQO, $80$ of multi-task, and $230$ of EI. We see that BQO outperforms the benchmark algorithms, and the multi-task algorithm underperforms the other algorithms considered.

\subsection{Problems Simulated from Gaussian Process Priors}
\label{sec:GPexample}

We now compare the performance of BQO against a benchmark Bayesian optimization algorithm on synthetic problems drawn at random from Gaussian process priors.  By varying the parameters of the Gaussian process prior, we study how BQO's performance relative to a benchmark (the KG algorithm) varies with problem characteristics, offering insight into the types of real-world problems on which BQO is likely to provide the most substantial benefit in comparison with using a traditional Bayesian optimization method.

These experiments show that the most important factor influencing BQO's relative performance is the speed with which $F(x,\w)$ varies with $\w$.
BQO provides the most value when this variation is large enough to influence performance, and small enough to allow $F(x,\w)$ to be modeled with a Gaussian process.  
Thus, users of BQO should choose a $\w$ that plays a big role in overall performance, and whose influence on performance is smooth enough to support predictive modeling.
These experiments also show that when settings are favorable, BQO provides substantial benefit, in some cases offering an improvement of almost $1000\%$.  On those few problems in which BQO underperforms the benchmark, it underperforms by a much smaller margin of less than $50\%$. 

% the conditional expectation $E[f(x,w,z) | w]$ varies slowly enough with $w$ to be modeled productively by a Gaussian process, but has enough variation to introduce     
% we demonstrate that SBO should be used when the correlation $\mbox{corr}\left[f\left(x,w,z\right),f\left(x,w'z\right)|x,w,w'\right]$ is large enough. We also show that the variance reduction $\frac{\mbox{Var}\left(f\left(x,w,z\right)|x\right)-\mbox{Var}\left(f\left(x,w,z\right)|x,w\right)}{\mbox{Var}\left(f\left(x,w,z\right)|x\right)}$ benefits SBO.
 %the difference of variances $D(x,\w):=\mbox{Var}\left[f\left(x,\w,\z\right)|f,x\right]-\mbox{Var}\left[f\left(x,\w,\z\right)|f,x,\w\right]$ is sufficiently large for many different points $(x,\w)$, or/and% 
 
% class of synthetic problems provide a controlled environment in which we can understand how problem characteristics influence the benefit provided by IBO. We use the KG algorithm to provide a benchmark against which to compare IBO's performance.

We now construct these problems in detail. 
Let $f(x,\w,\z)=h(x,\w)+r(\z)$ on $\left[0,1\right]^{2}\times\mathbb{R}$, where:
$r(z)$ is drawn, for each $z$ in a fine discretization of $[0,1]$, independently from a normal distribution with mean $0$ and variance $\alpha_{d}$
 (we could have set $r$ to be an Orstein-Uhlenbeck process with large volatility, and obtained an essentially identical result); and
$h$ is drawn from a Gaussian Process with mean $0$ and Gaussian covariance function $\Sigma\left(\left(x,\w\right),\left(x',\w'\right)\right)=\alpha_{h}\exp\left(-\beta\left\Vert \left(x,\w\right)-\left(x',\w'\right)\right\Vert _{2}^{2}\right)$.
We then define $F$ by $F(x,w)=E[f(x,W,Z)\mid W=w]$ where the expectation is over $Z$, and $G$ by $G(x)=E[f(x,W,Z)]$, where the expectation is over both $W$ and $Z$,  
$W$ is drawn uniformly from $\left\{ 0,1/49,2/49,\ldots,1\right\}$ and  $Z$ is drawn uniformly from the discretization of $[0,1]$.
To observed $F$, we draw 1 sample of $W$ and $Z$ and average $f(x,W,Z)$.
(We also performed experiments, not shown here, that observed $F$ by averaging multiple samples, and found the same qualitative behavior.)  

We thus have a class of problems parametrized by $\alpha_h$, $\alpha_d$, $\beta$, and an outcome measure determined by the overall number of samples.
Before displaying results, 
we reparametrize the dependence on $\alpha_h$ and $\alpha_d$ in what will be a more interpretable way. We first set $\mathrm{Var}[f(x,W,Z)|W,Z] = \alpha_h + \alpha_d$ to 1, as multiplying both $\alpha_h$ and $\alpha_d$ by a scalar simply scales the problem. 
Then, the variance reduction ratio
$\mathrm{Var}[f(x,W,Z)|W] / \mathrm{Var}[f(x,W,Z)]$
achieved by BQO in conditioning on $W$ is approximately 
$\alpha_h / (\alpha_d + \alpha_h)$, with this estimate becoming exact as $\beta$ grows large and the values of $h(x,w)$ become uncorrelated across $w$.
We define $A = \alpha_h / (\alpha_d + \alpha_h)$ equal to this approximate variance reduction ratio.
Thus, our problems are parametrized by the approximate variance reduction ratio $A$, 
the overall number of samples, and by $\beta$, which measures the speed with which $F(x,w)$ varies with $\w$.

\begin{figure}[!htbp]
    \centering
    \subcaptionbox{Normalized performance difference as a function of $\beta$ and $A$, when the overall number of samples is 50.
    \label{fig:sim3}}[0.35\linewidth]{ 
    \includegraphics[width=0.35\linewidth]{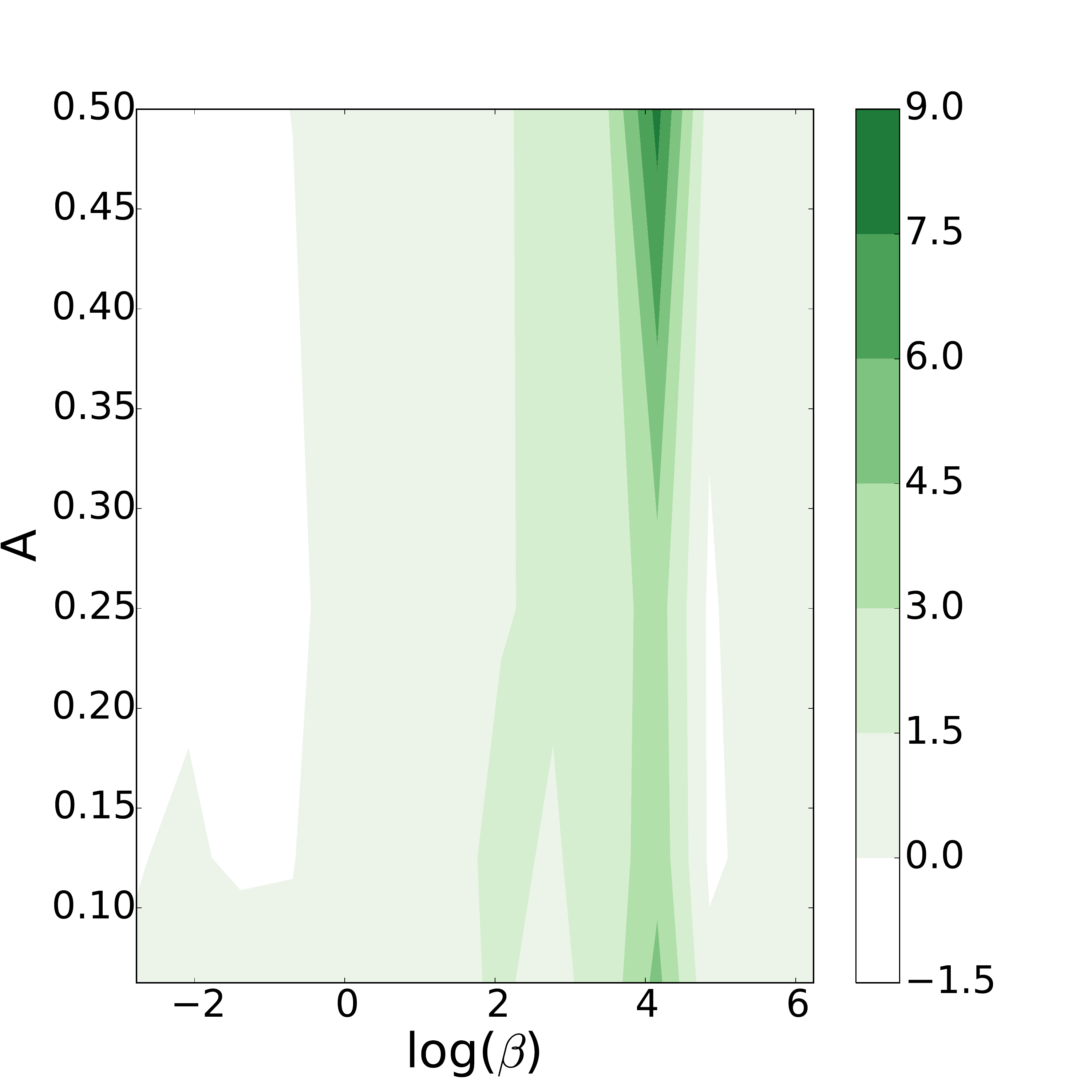}} 
    %\quad
    %\subcaptionbox{Performance between SBO and KG when the variance reduction is $1/16$. SBO is almost 9 times better when $\beta=64$. \label{fig:sim2}}[0.45\linewidth]{
    %\includegraphics[width=0.45\linewidth]{contourPlotbetahN8A2ver3.pdf}}
    \quad
    \subcaptionbox{Normalized performance difference as a function of $\beta$ and the overall number of iterations, when $A=1/2$.
    \label{fig:sim1}}[0.35\linewidth]{
    \includegraphics[width=0.35\linewidth]{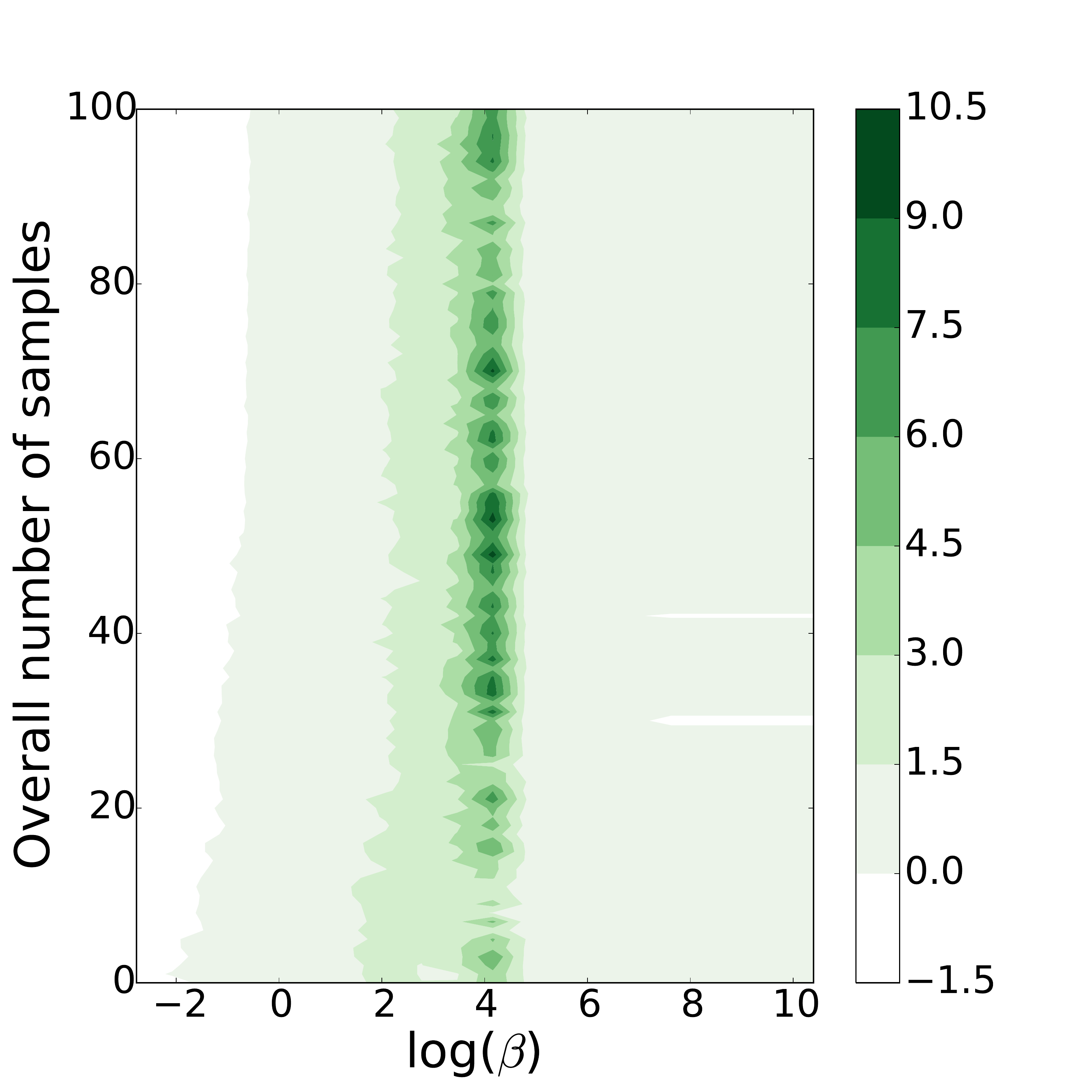}}
    %\quad
    %\subcaptionbox{Performance between SBO and KG when $n=8$. SBO is almost 9 times better when $\beta=64$. \label{fig:sim4}}[0.45\linewidth]{
    %\includegraphics[width=0.45\linewidth]{contourPlotbetahN8iter50ver3.pdf}}
\caption{ Normalized performance difference between BQO and KG in problems simulated from a Gaussian process, as a function of $\beta$, which measures how quickly $F(x,w)$ varies with $\w$, the approximate variance reduction ratio $A$, and the overall number of samples. BQO outperforms KG over most of the parameter space, and is approximately 10 times better when $\beta$ is near $\exp(4)$.
    \label{fig:simulated}}
\end{figure}

Given this parametrization, we sampled problems from Gaussian process priors using all combinations of 
$A\in\left\{ \frac{1}{2}, \frac14, \frac18, \frac1{16} \right\}$ and 
$\beta \in \left\{ 2^{-4}, 2^{-3},\ldots, 2^{9}, 2^{10} \right\}$.
We also performed additional simulations at $A=\frac12$ for $\beta \in \left\{ 2^{11},\ldots,2^{15}\right\}$.
%First, the observation variance $\mathrm{Var}[f(x,w,z)|f,x]$ from a single sample experienced by %benchmarks that do not condition on $w$ (KG, EI, Probability of improvement) is approximately %$(\alpha_{d}+\alpha_{h})=1$, with this estimate becoming exact as $\beta$ grows large and the %values of $h(x,w)$ become uncorrelated across $w$.
%We define $v = (\alpha_{d}+\alpha_{h})/n$
% For a fixed f, the variance is
% Var[f(x,w,z) | f] 
% = Var[h(x,w) + g(z) | h,g] 
% = Var[h(x,w) | h] + alphad
% Think of h(x,w) for a fixed x as being a long list of values that have marginal distribution N(0,alpha_h), and w as selecting from this list at random.  If the range of values taken by w is large, so that we can ignore correlation, then the distribution of the response is N(0,alpha_h).
% I should have said this: We define $C = (\alpha_{d}+\alpha_{h})/n$.
% In our experiments, we set $(\alpha_{d}+\alpha_{h})=1/n$
Figure~\ref{fig:simulated} shows Monte Carlo estimates of the normalized performance difference between BQO and KG for these problems, as a function of $\log(\beta)$ ($\log$ is the natural logarithm), $A$, and the number of samples taken overall.
% $E[(G(x^*_{\mathrm{SBO}}) -  G(x^*_{\mathrm{KG}}))) / |G(x^*_{\mathrm{KG}}|]$,
% where $x^*_\mathrm{SBO}$ is the final solution calculated by SBO, and similarly for $x^*_{\mathrm{KG}}$, and the expectation is taken both with respect to the randomness in the final solution quality for a given problem, and over the prior on problems.
The normalized performance difference is estimated for each set of problem parameters by taking a randomly sampled problem generated using those problem parameters, discretizing the domain into 2500 points, running each algorithm independently 500 times on that problem, and averaging $(G(x^*_{\mathrm{BQO}}) -  G(x^*_{\mathrm{KG}})) / |G(x^*_{\mathrm{KG}})|$ across these 500 samples, where $x^*_{\mathrm{BQO}}$ is the final solution calculated by BQO, and similarly for $x^*_{\mathrm{KG}}$.

The normalized performance difference is robust to $A$ and the overall number of samples, but is strongly influenced by $\beta$. BQO is always better than KG whenever $\beta \ge 1$. Moreover, it is substantially better than KG when $\mbox{log}(\beta)\in(3,5)$, with BQO outperforming KG by as much as a factor of $10$.   For larger $\beta$, BQO remains better than KG, but by a smaller margin. This unimodal dependence of the normalized performance difference on $\beta$ can be understood as follows: BQO provides value by modeling the dependence of $F(x,w)$ on $w$.   Modeling this dependence is most useful when $\beta$ takes moderate values because it is here where observations of $F(x,\w)$ at one value of $\w$ are most useful in predicting the value of $F(x,\w)$ at other values of $\w$.  When $F$ varies very quickly with $\w$ (large $\beta$), it is more difficult to generalize, and when $F$ varies very slowly with $\w$ ($\beta$ close to $0$), then modeling dependence on $\w$ is comparable with modeling $F$ as constant.

\section{Conclusions}
\label{conclusion}
We have presented a new Bayesian optimization algorithm, Bayesian Quadrature Optimization, designed for objectives that are sums or integrals of expensive-to-evaluate integrands.  This method is derived from a conceptual one-step optimality analysis for which we provide novel computational techniques that support efficient implementation.  We demonstrated that this method is consistent when the objective is a finite sum, and showed via extensive numerical experiments that it performs as well or better than the state of the art, providing substantial value when evaluations are noisy or the integrand varies smoothly in the integrated variables.  

\section*{Acknowledgments}

The authors were partially supported by NSF CAREER CMMI-1254298, NSF CMMI-1536895, AFOSR FA9550-15-1-0038,  AFOSR FA9550-16-1-0046, and DMR-1120296.

\putbib
\end{bibunit}

\appendix

\section*{Appendix}
%\label{appendix}

%\renewcommand{\thesubsection}{\Alph{subsection}}

\section{Proofs of Results in Section 4}
\label{proofs_monte_carlo_sect}

\begin{bibunit}

\textbf{Proof of \cref{postdist}.}
\proof
Recall that we assume throughout $\mathsection$\ref{sec:VOI} that 
$G$ has the integrated form \eqref{eq:goal2}. Results for \eqref{eq:goal1} are similar with the resulting expressions obtained by replacing integration over $w$ by a sum. 
By equation (\ref{eq:a_n}),
\begin{equation}
\begin{split}
a_{n+1}\left(x\right)
&=\int\mu_{n+1}\left(x,w\right)p\left(w\right)dw\\
&=\int\mu_{0}\left(x,w\right)p(w)dw
 +\left[B\left(x,1\right)\mbox{ }\cdots\mbox{ }B\left(x,n+1\right)\right]A_{n+1}^{-1}\left(\begin{array}{c}
y_{1}-\mu_{0}\left(x_{1},w_{1}\right)\\
\vdots\\
y_{n+1}-\mu_{0}\left(x_{n+1},w_{n+1}\right)\end{array}\right),
\end{split}
\label{eq:12-1}
\end{equation}
where $B(x,i):=\int\Sigma_{0}\left(x,w,x_{i},w_{i}\right)p\left(w\right)dw$ for
$1\leq i\leq n+1$.

Since $y_{n+1}$ conditioned on $H_{n},x_{n+1},w_{n+1}$
is normally distributed, then $a_{n+1}\left(x\right)\mid H_{n},x_{n+1},w_{n+1}$
is also normally distributed. By the tower property,
\begin{eqnarray*}
E_{n}\left[a_{n+1}\left(x\right)\mid x_{n+1},w_{n+1}\right] &= & E_{n}\left[E_{n+1}\left[G\left(x\right)\right]\mid x_{n+1},w_{n+1}\right]=  E_{n}\left[G\left(x\right)\right]=a_{n}\left(x\right),
\end{eqnarray*}
and by the law of total variance,
\begin{eqnarray*}
 \sigmatilde_{n}^{2}\left(x,x_{n+1},w_{n+1}\right) & := &\mbox{Var}_{n}[a_{n+1}(x)\mid x_{n+1},w_{n+1}] \\ & = & \mbox{Var}_{n}\left[E_{n+1}\left[G\left(x\right)\right]\mid x_{n+1},w_{n+1}\right]
  =  \mbox{Var}_{n}\left[G\left(x\right)\right]-E_{n}\left[\mbox{Var}_{n+1}\left[G\left(x\right)\right]\mid x_{n+1},w_{n+1}\right].
\end{eqnarray*}

Using the equation (\ref{eq:12-1}) and the previous expressions,
we get the following formula for $a_{n+1}$ 
\begin{eqnarray}
a_{n+1}\left(x\right)=a_{n}\left(x\right)+\sigmatilde_{n}\left(x,x_{n+1},w_{n+1}\right)Z \label{eq:post_dist_fut_0}
\end{eqnarray}
where $Z\sim N\left(0,1\right)$ conditioning on $H_{n}$, and $\sigmatilde_{n}\left(x,x_{n+1},w_{n+1}\right)=\pm\sqrt{\mbox{Var}_{n}[a_{n+1}(x)\mid x_{n+1},w_{n+1}]}$, which ends the proof of the first part of the lemma.

% Here, we give expressions for $a_n$  to compute the parameters of the posterior distribution of $a_{n+1}$. First,
% $a_{n}$ can be computed using the following formula,
% \begin{align*}
% a_{n}\left(x\right) &= E\left[\mu_{n}\left(x,\w\right)\right]\\
%  &= E\left[\mu_{0}\left(x,\w\right)\right]+\left[B\left(x,1\right)\mbox{ }\cdots\mbox{ }B\left(x,n\right)\right]A_{n}^{-1}\left(\begin{array}{c}
% y_{1}-\mu_{0}\left(x_{1},\w_{1}\right)\\
% \vdots\\
% y_{n}-\mu_{0}\left(x_{n},\w_{n}\right)
% \end{array}\right).
% \end{align*}
% In some cases it is possible to get a closed-form formula for $B$, for example if $\w$ follows a normal distribution, the components of $\w$ are independent and we use the squared exponential kernel, see \cref{gaussian_case}.

Finally, we only need to show the last claim of the lemma. We have that if $\lambda_{(x_{n+1},w_{n+1})} > 0$, or  $(x_{n+1},w_{n+1})$ is not in $H_{n}$, then $\Sigma_{n+1}(x_{n+1},w_{n+1},x_{n+1},w_{n+1})>0$, and then
\begin{eqnarray*}
\sigmatilde^2_{n}\left(x,x_{n+1},w_{n+1}\right) & = & \mbox{Var}_{n}\left[G\left(x\right)\right]-E_{n}\left[\mbox{Var}_{n+1}\left[G\left(x\right)\right]\mid x_{n+1},w_{n+1}\right]\\
 & = & \mbox{Var}_{n}\left[G\left(x\right)\right]-\mbox{Var}_{n+1}\left[G\left(x\right)\mid x_{n+1},w_{n+1}\right]\\
 & = & \int\int\Sigma_{n}\left(x,w,x,w'\right)p\left(w\right)p\left(w'\right)dwdw'
-\int\int\Sigma_{n+1}\left(x,w,x,w'\right)p\left(w\right)p\left(w'\right)dwdw'\\
 & = & \int\int\Sigma_{n}\left(x,w,x_{n+1},w_{n+1}\right)\frac{\Sigma_{n}\left(x,w',x_{n+1},w_{n+1}\right)}{\Sigma_{n}\left(x_{n+1},w_{n+1},x_{n+1},w_{n+1} \right)+\lambda_{(x_{n+1},w_{n+1})}}p\left(w\right)p\left(w'\right)dwdw'\\
%  & = & \left[\frac{\int\Sigma_{n}\left(x,w,x_{n+1},w_{n+1}\right)}{\sqrt{\Sigma_{n}\left(x_{n+1},w_{n+1},x_{n+1},w_{n+1}\right)}}p\left(w\right)dw\right]^{2}\\
 & = & \left[\frac{\int\Sigma_{n}\left(x,w,x_{n+1},w_{n+1}\right)p\left(w\right)dw}{\sqrt{\Sigma_{n}\left(x_{n+1},w_{n+1},x_{n+1},w_{n+1}\right)+\lambda_{(x_{n+1},w_{n+1})}}}\right]^{2}\\
 & = & \left[\frac{B\left(x,n+1\right)-\left[B\left(x,1\right)\mbox{ }\cdots\mbox{ }B\left(x,n\right)\right]A_{n}^{-1}\gamma}{\sqrt{\Sigma_{0}\left(x_{n+1},w_{n+1},x_{n+1},w_{n+1}\right)-\gamma^{T}A_{n}^{-1}\gamma+\lambda_{(x_{n+1},w_{n+1})}}}\right]^{2},
\end{eqnarray*}
where $\gamma^{T}:=(\Sigma_{0}(x_{n+1},w_{n+1},x_{1},w_{1}),\ldots,\Sigma_{0}(x_{n+1},w_{n+1},x_{n},w_{n}))$. Observe that in (\ref{eq:post_dist_fut_0}) 
the distribution of the left-hand side does not depend on the sign of $\sigmatilde_{n}$.  Thus, without loss of generality, 
we define $\sigmatilde_{n}\left(x,x_{n+1},w_{n+1}\right)$ equal to $\frac{B\left(x,n+1\right)-\left[B\left(x,1\right)\mbox{ }\cdots\mbox{ }B\left(x,n\right)\right]A_{n}^{-1}\gamma}{\sqrt{\Sigma_{0}\left(x_{n+1},w_{n+1},x_{n+1},w_{n+1}\right)-\gamma^{T}A_{n}^{-1}\gamma+\lambda_{(x_{n+1},w_{n+1})}}}$. If $\lambda_{(x_{n+1},w_{n+1})}=0$ and  $(x_{n+1},w_{n+1})$ is in $H_{n}$, then it is easy to see that $\sigmatilde^2_{n}\left(x,x_{n+1},w_{n+1}\right)=0$ because $\Sigma_{n}= \Sigma_{n+1}$.

\endproof

% \begin{lemma}

% We assume $\mu_{0}$ is constant, and the kernel $\Sigma_{0}$ of the prior distribution on $F$ is continuously differentiable and bounded. We also suppose there is a non-negative function $h$ such that $\int h(x,w',x')p(w')dw'$ is finite for all $x,x'\in A$, and $\left|\frac{\partial\Sigma_{0}\left(x,w',x',w\right)}{\partial w}\right|<h(x,w',x')$ for all $x,x'\in A$ and $w,w'\in W$. Then:
% \begin{enumerate}
% \item $a_{n}$ and $\sigmatilde_{n}\left(\cdot,x,w\right)$ are both continuously differentiable for any $x,w$ if $\Sigma_{n}(x,w,x,w) > 0$.
% \item For any $x'$, $\sigmatilde_{n}\left(x',x,w\right)$ is continuously differentiable with respect to  $x,w$ if $\Sigma_{n}(x,w,x,w) > 0$ and $\lambda_{(x,w)}$ is continuously differentiable.
% \end{enumerate}
% \end{lemma}

\vspace{5mm}

\textbf{Proof of \cref{smooth_ibo}.}
\proof
We first show that $a_{n}\left(\cdot\right)$ and $\sigmatilde_{n}\left(\cdot,x_{n+1},w_{n+1}\right)$ are continuously differentiable for each $\left(x_{n+1},w_{n+1}\right)$. In \cref{postdist},
we show that
\[
a_{n}\left(x\right)=\int\mu_{0}\left(x,w\right)p\left(w\right)dw-\left[B\left(x,1\right),\ldots,B\left(x,n\right)\right]A_{n}^{-1}\left[\begin{array}{c}
y_{1}-\mu_{0}\left(x_{1},w_{1}\right)\\
\vdots\\
y_{n}-\mu_{0}\left(x_{n},w_{n}\right)
\end{array}\right]
\]
and 
\[
\sigmatilde_{n}\left(x,x_{n+1},w_{n+1}\right) = \frac{B\left(x,n+1\right)-\left[B\left(x,1\right)\mbox{ }\cdots\mbox{ }B\left(x,n\right)\right]A_{n}^{-1}\gamma}{\sqrt{\Sigma_{0}\left(x_{n+1},w_{n+1},x_{n+1},w_{n+1}\right)-\gamma^{T}A_{n}^{-1}\gamma+\lambda_{(x_{n+1},w_{n+1})}}}.
\]
Thus we only need to show that $B\left(x,i\right)$ for all $1\leq i \leq n$ and $\int\mu_{0}\left(x,w\right)p\left(w\right)dw$ are continuously
differentiable on $x$. $\int\mu_{0}\left(x,w\right)p\left(w\right)dw$ is continuously
differentiable because $\mu_{0}$ is constant. 

We now show that $B\left(x,i\right)$ is continuously differentiable for any $1\leq i \leq n$.  $\Sigma_{0}$ is bounded, and so $x\mapsto\Sigma_{0}\left(x,w',y,w\right)p\left(w'\right)$
is integrable with respect to $w'$ for any $\left(y,w\right)$ because $p$ is integrable with
respect to $w'$. Moreover, $\Sigma_{0}\left(x,w',y,w\right)$ is differentiable with
respect to $x$, and $\frac{\partial\Sigma_{0}\left(x,w',y,w\right)}{\partial x}$ is continuous on $x$, and so it is bounded for any $y,w,w'$ fixed because $A$ is compact. Consequently, 
\begin{equation}
\frac{\partial B\left(x,i\right)}{\partial x}=\int\frac{\partial\Sigma\left(x,w',x_{i},w_{i}\right)}{\partial x}p\left(w'\right)dw'\label{eq:200}
\end{equation}
for all $i$ by Corollary 5.9 of \citet{bartle}. Moreover, by Corollary 5.8 of \citet{bartle}, $\frac{\partial B\left(x,i\right)}{\partial x}$
is continuous on $x$. This proves the first part of the lemma.

We now prove the second part of the lemma. Using a similar argument with the hypothesis that  $\left|\frac{\partial\Sigma_{0}\left(x,w',x',w\right)}{\partial w}\right|<h(x,w',x')$, and Corollaries 5.8 and 5.9 of \citet{bartle}, we can show that $(y,w)\mapsto \int\Sigma_{0}\left(x,w',y,w\right)p\left(w'\right)dw'$ is continuously differentiable for any x. Thus, using that $\lambda_{(x,w)}$ is continuously differentiable, we can conclude that $\sigmatilde_{n}\left(x',x,w\right)$ is continuously differentiable with respect to $(x,w)$. 

\endproof

% \begin{lemma}
% We assume that $\Sigma_{0}(\cdot,w,x',w
% ')$ is continuous for all $w,w'\in W$, $x'\in A$, $\Sigma_{0}$ is bounded, and $\mu_{0}$ is a constant. Suppose that we have an increasing sequence of finite discretizations $\left\{ A'_{L}\right\} _{L=1}^{\infty}$ of $A$, such that $\bigcup_{L=1}^{\infty}A'_{L}$ is dense in $A$. Then
% \begin{align*}
% V_n(x_{n+1},\w_{n+1})=\lim_{L\rightarrow\infty}\left(E_{n}\left[\max_{x\in A'_{L}} \left(a_{n}\left(x\right) + \sigmatilde(x, x_{n+1},\w_{n+1})Z_{n+1}\right)\right]-\max_{x \in A'_{L}} a_{n}\left(x\right)\right).
% \end{align*}
% \end{lemma}
\vspace{5mm}

\textbf{Proof of \cref{discretization_converge_past}.}
\proof{}
First, we show $a_{n}(x)$ and $\sigmatilde_{n}(\cdot,x_{n+1},\w_{n+1})$ are both uniformly continuous in $A$. By \cref{postdist}, 
\[
a_{n}\left(x\right)=\int\mu_{0}\left(x,w\right)p\left(w\right)dw-\left[B\left(x,1\right),\ldots,B\left(x,n\right)\right]A_{n}^{-1}\left[\begin{array}{c}
y_{1}-\mu_{0}\left(x_{1},w_{1}\right)\\
\vdots\\
y_{n}-\mu_{0}\left(x_{n},w_{n}\right)
\end{array}\right]
\]
and 
\[
\sigmatilde_{n}\left(x,x_{n+1},w_{n+1}\right) = \frac{B\left(x,n+1\right)-\left[B\left(x,1\right)\mbox{ }\cdots\mbox{ }B\left(x,n\right)\right]A_{n}^{-1}\gamma}{\sqrt{\Sigma_{0}\left(x_{n+1},w_{n+1},x_{n+1},w_{n+1}\right)-\gamma^{T}A_{n}^{-1}\gamma+\lambda_{(x_{n+1},w_{n+1})}}}.
\]
Thus we only need to show that $B\left(x,i\right)$ is continuous for all $1\leq i \leq n$ because $A$ is compact. The continuity follows from the Corollary 5.7 of \citet{bartle}.

$a_{n}(x)$ and $\sigmatilde_{n}\left(x,x_{n+1},w_{n+1}\right)$ are both bounded on $x$ because they are both continuous on a compact set, and so $E_{n}\left[\left|\mbox{sup}_{x\in A}a_{n+1}\left(x\right)\right|\right]<\infty$. Consequently by the monotone convergence theorem,
\begin{equation}
\lim_{L\rightarrow\infty}E_{n}\left[\max_{x\in A'_{L}} \left(a_{n}\left(x\right) + \sigmatilde(x, x_{n+1},\w_{n+1})Z_{n+1}\right)\right] = E_{n}\left[\lim_{L\rightarrow\infty}\max_{x\in A'_{L}} \left(a_{n}\left(x\right) + \sigmatilde(x, x_{n+1},\w_{n+1})Z_{n+1}\right)\right]. \label{disc:1}
\end{equation}

Furthermore, by a trivial modification to our \cref{approxdomain} using the uniform continuity of $a_{n}(x)$ and $\sigmatilde_{n}(\cdot,x_{n+1},\w_{n+1})$, we can see that  $\mbox{lim}_{L\rightarrow\infty}\mbox{max}_{x\in A'_{L}}a_{n}\left(x\right)=\mbox{max}_{x\in A}a_{n}\left(x\right)$, and
\begin{align}
\lim_{L\rightarrow\infty}\max_{x\in A'_{L}} \left(a_{n}\left(x\right) + \sigmatilde(x, x_{n+1},\w_{n+1})Z_{n+1}\right)= \max_{x\in A} \left(a_{n}\left(x\right) + \sigmatilde(x, x_{n+1},\w_{n+1})Z_{n+1}\right) \mbox{ a.s.}
\end{align}
Thus, using the two previous equations and (\ref{disc:1}), we conclude the proof of the lemma.\Halmos 
\endproof

\section{BQO's Time and Space Complexity}
\label{bqo_complexity}

In this section, we discuss BQO's time and space complexity, assuming we use it to select $N$ points $(x,\w)$ to sample. To select each point $(x,\w)$ to sample, we use \Cref{alg:SBO}, which runs the ADAM algorithm for $T$ iterations.  Each iteration requires a stochastic gradient computed using $J$ independent standard Gaussian random variables, $J$ independent samples from the posterior on $\theta$, and $J$ runs of LBFGS to maximize 
$\left(a_{n}\left(x'\right)+\sigmatilde_{n}\left(x',x_{n+1},w_{n+1}\right)Z_{i}\right)$. 
Let $K$ be the number of steps in a single run of LBFGS, where each step requires an evaluation of $a_n$, $\sigmatilde_n$, and their gradients. Let $O(L)$ be the complexity of computing the kernel and its gradient, and let $O(S)$ be the complexity of computing (\ref{eq:200}),$\nabla_{n+1}B\left(x,n+1\right)$, and $B(x,i)$ for all $i\leq n$.

To obtain a sample $\widehat{\theta}_{j}$ of $\theta$ from its posterior, we run slice sampling for at most $Q$ iterations. To obtain a sample $\widehat{\theta}_{j}$ of $\theta$ from its posterior, we run slice sampling for at most $Q$ iterations. Each iteration of slice sampling requires computing the likelihood of the data $H_{n}$ given the candidate parameters $\theta$ for the model.
Computing the likelihood requires first computing $A_n$, which requires $O(n^2)$ kernel evaluations, each of which has complexity $L$.
It then requires  computing the Cholesky decomposition of $A_{n}$ (which has complexity $O(n^3)$) and solving a triangular system of equations involving this Cholesky decomposition (which has complexity $O(n^2)$).    Thus, each iteration of slice sampling has time complexity $O(n^3+Ln^2)$ and space complexity $O(n^2)$, and the total complexity of sampling $\theta$ once using slice sampling is  $O(Qn^3+QLn^2)$. 
Since we obtain $J$ samples for each value of $n$ from $n=1$ to $n=N$, the total complexity due to sampling $\theta$ is $O(JQN^4 + JQLN^3)$ in time, and $O(N^2)$ in space.

Now, for each iteration of the ADAM algorithm, we optimize $a_{n+1}(x;\widehat{\theta}_{j})=a_{n}(x;\widehat{\theta}_{j})+\sigmatilde_n(x,x_{n+1},w_{n+1};\widehat{\theta}_{j})Z_{j}$ using LBFGS for at most $K$ iterations where  $\widehat{\theta}_{j}$ is a sample of the hyperparameters of the model and $Z_{j}$ is a sampled Gaussian random variable. 

For each point $x$ visited within this optimization, we compute $a_{n}(x;\widehat{\theta}_{j})$ and its gradient, which 
requires computing $B\left(x,n+1\right)-\left[B\left(x,1\right)\mbox{ }\cdots\mbox{ }B\left(x,n\right)\right]A_{n}^{-1}(y_{1:n}-\mu_{0}(x_{1:n})$ and its gradient (see \cref{postdist}).
This has time complexity $O(nS + n^2)$, because 
the computation of each of the $O(n)$ values of $B(x,i)$ and their gradients is $O(S)$, giving the $O(nS)$ term. We then compute the matrix product involving $A_n^{-1}$ using the pre-computed Cholesky decomposition of $A_n$ by solving a triangular linear system with complexity $O(n^2)$. 
(The complexity of computing the Cholesky decomposition of $A_{n}$ for this $\widehat{\theta}_{j}$ has already been accounted for above.)

For each point $x$ visited in the optimization we also compute
$\sigmatilde_n(x,x_{n+1},w_{n+1};\widehat{\theta}_{j})$ and its gradient with respect to $x$.  
To do this, we compute $B\left(x,n+1\right)-\left[B\left(x,1\right)\mbox{ }\cdots\mbox{ }B\left(x,n\right)\right]A_{n}^{-1}\gamma$ and its gradient with respect to $x$, and $\Sigma_{n+1}(x_{n+1},w_{n+1},x_{n+1},w_{n+1})$.
In doing so, we re-use the previously discussed computation of the $B(x,i)$ terms and their gradients, and computation of 
the matrix product of $\left[B\left(x,1\right)\mbox{ }\cdots\mbox{ }B\left(x,n\right)\right]$ and the inverse of the Cholesky decomposition of $A_n$.  The additional computation required is the matrix product of this same inverse of the Cholesky decomposition of $A_n$ with $\gamma$, which does not depend on $x$ and can be performed only once per optimization of $a_{n+1}$, saving a factor of $K$ below.
This extra computation introduces an additional time complexity incurred once per optimization of $O(nL+n^2)$.

% which has time complexity $O(nL+n^2)$ assuming that the $B$ terms and their gradients have already been computed, by an argument similar to above.  Moreover, this extra computation must be performed only once per optimization of $a_{n+1}(x;\widehat{\theta}_j)$ because it does not depend on $x$.

Consequently, the overall complexity due to optimizing $a_{n+1}(x;\widehat{\theta}_j)$ within the ADAM algorithm for a single $n$ (but not including the complexity of sampling $\widehat{\theta}_j$  and computing the Cholesky decomposition of $A_n$)
is $O(JT(K(nS + n^2)+nL))$ in time, and $O(n^2)$ in space. 
This computation is repeated for $n$ ranging from $1$ to $N$, and has overall time complexity $O(JTK(SN^2 + N^3)+JTLN^2)$ and space complexity $O(N^2)$.

By the previous two paragraphs, we conclude that the BQO algorithm has time complexity $O(JQN^4 + JQLN^3 + JTK(SN^2 + N^3)+JTLN^2)$ and space complexity $O(N^2)$.

\section{Closed-Form Expressions for the Gaussian and Squared Exponential Kernel Case}
\label{gaussian_case}

In this section we give closed-form expressions for $B$ and its gradient to compute BQO and its gradient when we use the squared exponential kernel $\Sigma\left(\left(x,z\right),\left(y,w\right)\right)=\sigma_{0}^{2}\mbox{exp}\left(-\sum_{k=1}^{n}\alpha_{1}^{\left(k\right)}\left[x_{k}-y_{k}\right]^{2}-\sum_{k=1}^{d}\alpha_{2}^{\left(k\right)}\left[w_{k}-z_{k}\right]^{2}\right)$, and 
\begin{equation}
p\left(w\right)=\Pi_{i=1}^{n}\varphi_{i}\left(w_{i}\right)\label{eq:101}
\end{equation}
where $w=\left(w_{1},\ldots,w_{n}\right)$, and $\varphi_{i}$ is
the density of a normal random variable with mean $\mu_{i}$ and variance
$\sigma_{i}^{2}$ . 

The previous assumptions are quite common, for example they are assumed in \citet{Xie:2012}. Moreover, we can always transform a problem of the form
$\int F\left(x,w\right)dp\left(w\right)dw$ to a new problem of the form $\int F'\left(x,w\right)p'\left(w\right)dw$ where $p'$ is of the form given in (\ref{eq:101}) under suitable conditions. The procedure and necessary conditions are:

\begin{itemize}
\item Denote the continuous density of $W$ by $p$, where $W$ is a random vector
such that $P\left[W=w\right]=p\left(w\right)$.
\item Assume that we can compute or estimate the marginals of the random vector $W$:
$F_{i}\left(w_{i}\right):=P\left(W_{i}\leq w_{i}\right)$ for all
$i$.
\item Denote the inverse of the distribution function of a standard Gaussian random variable by $h$. We
have that $h\left(F_{i}\left(W_{i}\right)\right)\sim N\left(0,1\right)$
for all $i$. Thus, 
\[
h\left(F\left(W\right)\right)=\left(h\left(F_{1}\left(W_{1}\right)\right),\ldots,h\left(F_{n}\left(W_{n}\right)\right)\right)\sim N\left(0,\Sigma\right).
\]
\item We assume that $\Sigma$ is
positive definite, and can be computationally estimated.
\item We have that $\Sigma^{-1/2}h\left(F\left(W\right)\right)\sim N\left(0,I\right)$. 
\item Define $Y=\Sigma^{-1/2}h\left(F\left(W\right)\right)=\left(h_{1}\left(W\right),\ldots,h_{n}\left(W\right)\right)$,
where $h_{i}\left(w\right)=\left[\Sigma^{-1/2}h\left(F\left(W\right)\right)\right]_{i}$
is the ith component of the vector $\Sigma^{-1/2}h\left(F\left(W\right)\right)$
for all $i$. Let $J_{i}$ denote the Jacobian computed from $h_{i}$.
Assume that $J_{i}$ does not vanish identically. By the change of variables theorem, we have that 
\[
\int F\left(x,w\right)p\left(w\right)dw=\int F'\left(x,y\right)p'\left(y\right)dy
\]
where $F'\left(x,y\right)=F\left(x,\left(F_{1}^{-1}\left(h^{-1}\left[\Sigma^{1/2}y\right]_{1}\right),\ldots,F_{n}^{-1}h^{-1}\left[\Sigma^{1/2}y\right]_{n}\right)\right)\left|J_{i}\right|$
and $\left[\Sigma^{1/2}y\right]_{i}$ is the ith entry of $\Sigma^{1/2}y$.
\end{itemize}
 
This shows how to transform a general problem to a problem with density given by (\ref{eq:101}), and the conditions under which this is possible. However, we do not always want to use this transformation because the correlation between $F'(x,\w)$ and $F'(x',\w')$ can be too small, which may not be optimal for BQO as we discuss in $\mathsection$\ref{sec:GPexample}.

We now compute the closed-form expressions of $B$ and its gradient. We have that

\begin{eqnarray*}
B\left(x,i\right) & = & \int\Sigma_{0}\left(x,w,x_{i},w_{i}\right)p\left(w\right)dw\\
 & = & \sigma_{0}^{2}\mbox{exp}\left(-\sum_{k=1}^{n}\alpha_{1}^{\left(k\right)}\left[x_{k}-x{}_{ik}\right]^{2}\right)\prod_{k=1}^{d}\int\mbox{exp}\left(-\alpha_{2}^{\left(k\right)}\left[w_{k}-w_{ik}\right]^{2}\right)\varphi_{k}\left(w_{k}\right)d\left(w_{k}\right)
\end{eqnarray*}

for $i=1,\ldots,n$. Thus, we only need to compute $\int\mbox{exp}\left(-\alpha_{2}^{\left(k\right)}\left[w_{k}-w_{ik}\right]^{2}\right)\varphi_{k}\left(w_{k}\right)d\left(w_{k}\right)$
for any $k$ and $i$, which is given by the following equations,

\begin{eqnarray*}
\int\mbox{exp}\left(-\alpha_{2}^{\left(k\right)}\left[w_{k}-w_{ik}\right]^{2}\right)\varphi_{k}\left(w_{k}\right)d\left(w_{k}\right) & = & \frac{1}{\sqrt{2\pi}\sigma_{k}}\int\mbox{exp}\left(-\alpha_{2}^{\left(k\right)}\left[z-w_{ik}\right]^{2}-\frac{\left[z-\mu_{k}\right]^{2}}{2\sigma_{k}^{2}}\right)dz\\
 & = & \frac{1}{\sqrt{2\pi}\sigma_{k}}\mbox{exp}\left(-\frac{\mu_{k}^{2}}{2\sigma_{k}^{2}}-\alpha_{2}^{\left(k\right)}\left(w_{ik}\right)^{2}-\frac{\left(\frac{\mu_{k}}{\sigma_{k}^{2}}+2\alpha_{2}^{\left(k\right)}w_{ik}\right)^{2}}{4\left(-\alpha_{2}^{\left(k\right)}-\frac{1}{2\sigma_{k}^{2}}\right)}\right)\\
 &  & \times\int\mbox{exp}\left(-\left(\alpha_{2}^{\left(k\right)}+\frac{1}{2\sigma_{k}^{2}}\right)\left[z-\frac{\frac{\mu_{k}}{\sigma_{k}^{2}}+2\alpha_{2}^{\left(k\right)}w_{ik}}{2\left(b+\frac{1}{2\sigma_{k}^{2}}\right)}\right]^{2}\right)dz\\
 & = & \frac{1}{\sqrt{2}\sigma_{k}}\frac{1}{\sqrt{\alpha_{2}^{\left(k\right)}+\frac{1}{2\sigma_{k}^{2}}}}\\
 &  & \times\mbox{exp}\left(-\frac{\mu_{k}^{2}}{2\sigma_{k}^{2}}-\alpha_{2}^{\left(k\right)}\left(w_{ik}\right)^{2}-\frac{\left(\frac{\mu_{k}}{\sigma_{k}^{2}}+2\alpha_{2}^{\left(k\right)}w_{ik}\right)^{2}}{4\left(-\alpha_{2}^{\left(k\right)}-\frac{1}{2\sigma_{k}^{2}}\right)}\right).
\end{eqnarray*}

This shows how to compute $B$. We now compute the gradient of $B$. Observe that 
\begin{eqnarray*}
\nabla_{x_{n+1},j}\Sigma_{0}\left(x_{n+1},w_{n+1},x_{i},w_{i}\right) & = & \begin{cases}
0, & i=n+1\\
-2\alpha_{1}^{\left(j\right)}\left[x_{n+1,j}-x_{i,j}\right]\Sigma_{0}\left(x_{n+1},w_{n+1},x_{i},w_{i}\right), & i<n+1
\end{cases}\\
\nabla_{w_{n+1},j}\Sigma_{0}\left(x_{n+1},w_{n+1},x_{i},w_{i}\right) & = & \begin{cases}
0, & i=n+1\\
-2\alpha_{2}^{\left(j\right)}\left[w_{n+1.j}-w_{i,j}\right]\Sigma_{0}\left(x_{n+1},w_{n+1},x_{i},w_{i}\right), & i<n+1
\end{cases}
\end{eqnarray*}

where $\nabla_{x_{n+1},j}$ is the derivative respect to the jth entry
of $x_{n+1}$. Consequently,

\begin{eqnarray*}
\nabla_{x_{n+1,j}}B\left(x,n+1\right) & = & -2\alpha_{1}^{\left(j\right)}\left(x{}_{j}-x_{n+1,j}\right)B\left(x,n+1\right)\\
\nabla_{w_{n+1},k}B\left(x,n+1\right) & = & \sigma_{0}^{2}\mbox{exp}\left(-\sum_{i=1}^{n}\alpha_{1}^{\left(i\right)}\left[x_{i}-x{}_{n+1,i}\right]^{2}\right)\prod_{j\neq k}\int\mbox{exp}\left(-\alpha_{2}^{\left(j\right)}\left[w_{j}-w_{n+1,j}\right]^{2}\right)\varphi_{j}\left(w_{j}\right)d\left(w_{j}\right)\\
 &  & \times\int\left(-2\alpha_{2}^{\left(k\right)}\left(w_{k}-w_{n+1,k}\right)\right)\mbox{exp}\left(-\alpha_{2}^{\left(k\right)}\left[w_{k}-w_{n+1,k}\right]^{2}\right)\varphi_{k}\left(w_{k}\right)d\left(w_{k}\right),
\end{eqnarray*}
and 
\begin{eqnarray*}
\int w_{k}\mbox{exp}\left(-\alpha_{2}^{\left(k\right)}\left[w_{k}-w_{n+1,k}\right]^{2}\right)\varphi_{k}\left(w_{k}\right)d\left(w_{k}\right) & = & \frac{1}{\sqrt{2\pi}\sigma_{k}}\int z\mbox{exp}\left(-\alpha_{2}^{\left(k\right)}\left[z-w_{ik}\right]^{2}-\frac{\left[z-\mu_{k}\right]^{2}}{2\sigma_{k}^{2}}\right)dz\\
 & = & \frac{1}{\sqrt{2\pi}\sigma_{k}}\mbox{exp}\left(-\frac{\mu_{k}^{2}}{2\sigma_{k}^{2}}-\alpha_{2}^{\left(k\right)}\left(w_{ik}\right)^{2}-\frac{\left(\frac{\mu_{k}}{\sigma_{k}^{2}}+2\alpha_{2}^{\left(k\right)}w_{ik}\right)^{2}}{4\left(-\alpha_{2}^{\left(k\right)}-\frac{1}{2\sigma_{k}^{2}}\right)}\right)\\
 &  & \times\int z\mbox{exp}\left(-\left(\alpha_{2}^{\left(k\right)}+\frac{1}{2\sigma_{k}^{2}}\right)\left[z-\frac{\frac{\mu_{k}}{\sigma_{k}^{2}}+2\alpha_{2}^{\left(k\right)}w_{ik}}{2\left(b+\frac{1}{2\sigma_{k}^{2}}\right)}\right]^{2}\right)dz\\
 & = & \frac{1}{\sqrt{2}\sigma_{k}}\mbox{exp}\left(-\frac{\mu_{k}^{2}}{2\sigma_{k}^{2}}-\alpha_{2}^{\left(k\right)}\left(w_{ik}\right)^{2}-\frac{\left(\frac{\mu_{k}}{\sigma_{k}^{2}}+2\alpha_{2}^{\left(k\right)}w_{ik}\right)^{2}}{4\left(-\alpha_{2}^{\left(k\right)}-\frac{1}{2\sigma_{k}^{2}}\right)}\right)\\
 &  & \times\left(\frac{\frac{\mu_{k}}{\sigma_{k}^{2}}+2\alpha_{2}^{\left(k\right)}w_{ik}}{2\left(b+\frac{1}{2\sigma_{k}^{2}}\right)}\right),
\end{eqnarray*}
which shows how to compute the gradient of $B$.

% Moreover, if $p$ is not the Gaussian density, and
% $w$ is one-dimensional (or $p\left(w\right)=\Pi_{i=1}^{n}p_{i}\left(w_{i}\right)$),
% in many cases we can transform the integral $\int F\left(x,w\right)p\left(w\right)dw$
% to a new integral $\int F\left(x,w\right)p'\left(w\right)dw$ where
% $p'$ is proportional to the form given in (\ref{eq:101}).
\section{Illustration of Poor Performance of the Multi-Task Algorithm}
\label{bad_example_mt}

In this section we give an example where the average multi-task algorithm presented in Section 3.2 of \citet{swersky2013multi} is inefficient, and we show that the BQO algorithm does not have that problem.

Let $A=\left\{ 1,2\right\} $, $W=\left\{ 1,\ldots,M\right\} $, and
$w_{1}=0.5$, $w_{i}=\frac{0.5}{M-1}$ if $i>1$. We want to maximize
\[
G\left(x\right)=\sum_{i=1}^{M}w_{i}F\left(x,i\right).
\]
We assume that $F\left(1,1\right)=\cdots=F\left(1,M\right)=L>0$, and $F\left(1,1\right)=L>0$ has been evaluated. In addition, we suppose that $F\left(x,i\right)\sim N\left(0,M-1\right)$ for $i>1$, $F\left(2,1\right)\sim N\left(0,v^{2}\right)$
for some $v^{2}>1$, and their correlation is equal to zero. By equation 15 in \citet{jones1998efficient}, we have that
\[
EI_{n}\left(x\right)=\left(a_{n}\left(x\right)-\mbox{max}_{i\leq n}a_{n}\left(x_{i}\right)\right)\Phi\left(z_{n}\right)+\sqrt{\Sigma_{n}\left(x,x\right)}\phi\left(z_{n}\right)
\]
if $\Sigma_{n}\left(x,x\right)>0$, where $z_{n}=\left(a_{n}\left(x\right)-\mbox{max}_{i\leq n}a_{n}\left(x_{i}\right)\right)/\sqrt{\Sigma_{n}\left(x,x\right)}$,
and $\Sigma_{n}\left(x,x\right)$ is the posterior variance of $G\left(x\right)$.
Consequently,
\[
EI_{1}\left(1\right)=\frac{1}{\sqrt{2\pi}}\sqrt{\left(\frac{1}{4\left(M-1\right)^{2}}\right)M-1}=\frac{1}{\sqrt{2\pi}}\sqrt{\left(\frac{1}{4\left(M-1\right)}\right)}
\]
and
\[
EI_{1}\left(2\right)=\phi\left(z_{1,2}\right)\sqrt{\left(\frac{1}{4\left(M-1\right)}+\frac{v^{2}}{4}\right)}+\Phi\left(z_{1,2}\right)\left(-\frac{L}{2}\right)
\]
where $z_{1,2}=\frac{-\frac{L}{2}}{\sqrt{\left(\frac{1}{4\left(M-1\right)}+\frac{v^{2}}{4}\right)}}$. 

Suppose that in the next iteration we choose again a point of the form
$\left(1,i\right)$, and $F\left(1,i\right)=L_{i}$, thus we have
that
\[
EI_{2}\left(1\right)=\frac{1}{\sqrt{2\pi}}\sqrt{\left(\frac{1}{4\left(M-1\right)^{2}}\right)\left(M-2\right)}
\]
and
\[
EI_{2}\left(2\right)=\phi\left(z_{2,2}\right)\sqrt{\left(\frac{1}{4\left(M-1\right)}+\frac{v^{2}}{4}\right)}+\Phi\left(z_{2,2}\right)\left(-\frac{L}{2}-\frac{1}{2\left(M-1\right)}L\right)
\]
where $z_{2,2}=\frac{-\frac{L}{2}-\frac{1}{2\left(M-1\right)}L}{\sqrt{\left(\frac{1}{4\left(M-1\right)}+\frac{v^{2}}{4}\right)}}$.
Similarly, we can see that if we keep choosing points of the form
$\left(1,i\right)$, we are going to have that
\[
EI_{n}\left(1\right)=\frac{1}{\sqrt{2\pi}}\sqrt{\left(\frac{1}{4\left(M-1\right)^{2}}\right)\left(M-n\right)}
\]
and
\[
EI_{n}\left(2\right)=\phi\left(z_{n,2}\right)\sqrt{\left(\frac{1}{4\left(M-1\right)}+\frac{v^{2}}{4}\right)}+\Phi\left(z_{n,2}\right)\left(-\frac{L}{2}-\frac{n-1}{2\left(M-1\right)}L\right)
\]
where $z_{n,2}=\frac{-\frac{L}{2}-\frac{n-1}{2\left(M-1\right)}L}{\sqrt{\left(\frac{1}{4\left(M-1\right)}+\frac{v^{2}}{4}\right)}}$
if $n\leq M$.

Observe that 
\[
\phi\left(\frac{-\frac{L}{2}-\frac{n-1}{2\left(M-1\right)}L}{\sqrt{\left(\frac{1}{4\left(M-1\right)}+\frac{v^{2}}{4}\right)}}\right)\rightarrow0
\]
uniformly on $n<M$ if $L\rightarrow\infty$. Consequently, for $L$
large, we have that
\[
EI_{n}\left(2\right)<EI_{n}\left(1\right)
\]
if $n<M$, and so we can only choose a point of the form $\left(2,j\right)$
until we have evaluated all the $M$ different points of the form $\left(1,i\right)$,
which is clearly inefficient when $M$ is large. 

We now show that the BQO algorithm does not have that problem. Observe
that if $i>1$,

\begin{eqnarray*}
\left(\frac{\sum_{j}w_{j}\Sigma\left(1,i,1,j\right)-\sum_{j}w_{j}\Sigma\left(1,i,2,j\right)}{\Sigma\left(1,i,1,i\right)}\right)^{2} & = & \left(\frac{\frac{1}{2}}{M-1}\right)^{2}\\
 & < & \left(\frac{\frac{v^{2}}{2}}{M-1}\right)^{2}\\
 & = & \left(\frac{\sum_{j}w_{j}\Sigma\left(2,1,1,j\right)-\sum_{j}w_{j}\Sigma\left(2,1,2,j\right)}{\Sigma\left(2,1,2,1\right)}\right)^{2}
\end{eqnarray*}
and by \cref{inequality_ibo} we have that $BQO\left(\left(1,i\right)\right)<BQO\left(\left(2,1\right)\right)$
for all $i$, and thus the BQO algorithm chooses to evaluate $\left(2,1\right)$.

\begin{theorem}
\label{inequality_ibo}
We have that 
\[
BQO_{n}\left(t,s\right) \geq BQO_{n}(z,w')
\]
if for all $x,y$ 
\[
\left(\frac{\int\Sigma_{n}\left(x,w,z,w'\right)dp\left(w\right)-\int\Sigma_{n}\left(y,w,z,w'\right)dp\left(w\right)}{\sqrt{\Sigma_{n}\left(z,w',z,w'\right)}}\right)^{2}\leq\left(\frac{\int\Sigma_{n}\left(x,w,t,s\right)dp\left(w\right)-\int\Sigma_{n}\left(y,w,t,s\right)dp\left(w\right)}{\sqrt{\Sigma_{n}\left(t,s,t,s\right)}}\right)^{2}
\]

\end{theorem}

\proof{}
The proof is a direct consequence of \cref{postdist} and the Vitale's extension of the Sudakov-Fernique inequality \citep{vitale2000}.

\endproof

\section{Consistency of BQO}
In this section, we prove the two consistency results: Theorem~\ref{continuum_th_intuitive} and Theorem~\ref{main_theorem_simplified} of $\mathsection$\ref{sec:asymptotic}. To support the proofs of our theorems, we embed BQO within a controlled Markov process framework. We denote the probability space by $\left(\Omega,\mathcal{F},P\right)$, and assume it is complete. The action space is the domain of $F$, which is $A\times W$, where $W$ is a subset of $\mathbb{R}^{\ell}$. The state space is the set of possible parameters of the Gaussian process on $F$.  More formally, if we denote $A\times W$ by $A'$, the state space is defined by $\mathcal{H}:=D\left(A'\right)\times D_{\mbox{kernel}}\left(A'\times A'\right)$,
where $D\left(A'\right)$ is the set of functions defined on $A'$, and $D_{\mbox{kernel}}\left(A'\times A'\right)$ is the set of positive semidefinite functions defined on $A'\times A'$.

The discrete-time dynamic system, which shows how the posterior parameters change when a new observation is obtained, is given by
\[
S^{n+1}:=\left(\mu_{n+1},\Sigma_{n+1}\right)=f_{n}\left(\left(\mu_{n},\Sigma_{n}\right),u_{n},Z_{n}\right)
\]
where $u_{n}$ is the chosen point to measure at stage $n$, $Z_{n}\sim N\left(0,1\right)$,
and $f_{n}$ is defined by
\[
\mu_{n+1}\left(x', w'\right)=\mu_{n}\left(x', w'\right)+\frac{\Sigma_{n}\left((x',w'), u_{n}\right)}{\sqrt{\Sigma_{n}\left(u_{n},u_{n}\right)+\lambda_{u_{n}}}}Z_{n}
\]
and
\[
\Sigma_{n+1}\left((y,r),(z,s)\right)=\Sigma_{n}\left((y,r),(z,s)\right)+\frac{\Sigma_{n}\left((y,r),u_{n}\right)\Sigma_{n}\left((z,s),u_{n}\right)}{\Sigma_{n}\left(u_{n},u_{n}\right) + \lambda_{u_{n}}}
\]
where $\lambda_{u_{n}}$ is the variance of the sample $y_{u_{n}}$ of $F(u_{n})$, e.g. $y_{u_{n}}=F(u_{n})+\epsilon_{n}$ where $\epsilon_{n}\sim N(0, \lambda_{u_{n}})$.

We now define a sequence of value functions $V^{n}:\mathcal{H}\rightarrow\mathbb{R}$
by
\[
V^{n}\left(s\right):=\mbox{sup}_{\pi\in\Pi}E^{\pi}\left[\mbox{max}_{x\in A}a_{N}\left(x\right)\mid S^{n}=s\right]
\]
where $N$ is the number of times that we can evaluate $F$, $\Pi$ is the class of admissible policies (a policy is a sequence of maps $\pi=(\pi_{1},\ldots,\pi_{N})$, where each $\pi_{i}$ maps parameters of the Gaussian process into the domain of $F$), and conditioning on $S^{n}=s$ means that $F\sim GP(s)$ where $s=(\mu,\Sigma)$.

Similarly, we define the value of a policy $\pi\in\Pi$ as
\[
V^{\pi,n}\left(s\right):=E^{\pi}\left[V^{N}\left(S^{N}\right)\mid S^{n}=s\right]=E^{\pi}\left[\mbox{max}_{x\in A}a_{N}\left(x\right)\mid S^{n}=s\right].
\]
An optimal policy given $s$ optimizes $V^{\pi,0}\left(s\right)$.

 We denote the weighted sum of a function over $W$ by its Lebesgue integral respect to $p$. Given $s$, we denote the parameters of the induced GP on $G$ by $T(s)$, where the mean is $a\left(x\right)=\int_{w}\mu\left(x,w\right)dp\left(w\right)$
and the covariance is $\alpha\left(x,y\right)=\int_{w}\int_{w'}\Sigma\left(\left(x,w\right),\left(y,w'\right)\right)dp\left(w\right)dp\left(w'\right)$. Observe that the integrals are sums because $W$ is finite.

We define the set $\mathcal{H_{0}} \subset \mathcal{H}$ as
\[
\left\{ \left(\mu,\Sigma\right):\mu\equiv0, \Sigma_{w,w'}\left(x,y\right):=\Sigma\left(x,w,y,w'\right)\mbox{ is in }C^{1}\left(\mathbb{R}^{n}\times\mathbb{R}^{n}\right) \mbox{ for
all } w,w'\in W, \mbox{ and }\Sigma\mbox{ is isotropic}\right\}.
\]

We summarize the notation that is used in this section:

\begin{eqnarray*}
V^{n}\left(s\right) & = & \mbox{sup}_{\pi\in\Pi}E^{\pi}\left[\mbox{max}_{x\in A}a_{N}\left(x\right)\mid S^{n}=s\right]=:V^{n}\left(s;N\right)\\
U\left(s\right) & = & E\left[\mbox{max}_{x\in A}G\left(x\right)\mid S^{0}=s\right]\\
V^{\pi,n}\left(s;N\right) & = & E^{\pi}\left[\mbox{max}_{x\in A} a_{N}\left(x\right)\mid S^{n}=s\right]:=V^{\pi,n}\left(s\right)\\
R^{n}\left(s;(x,w)\right) & = & E\left[V^{n+1}\left(S^{n+1}\right)\mid S^{n}=s,(x_{n+1},w_{n+1})=(x,w)\right]\\
\mathcal{F}^{n} &=& \mbox{the smallest sigma-algebra generated by the observations obtained by time } n\\
\mathcal{F}^{\infty} &=& \mbox{the smallest sigma-algebra generated by } \left\{ \mathcal{F}^{n} \right\}_{n}\\
\mu_{n}\left(x,w\right)&:=& E\left[F\left(x,w\right)\mid\mathcal{F}^{n}\right]
\mbox{ for } 0\leq n\leq\infty\\
\Sigma_{n}\left(x,w,x',w'\right)&:=&\mbox{Cov}\left[F\left(x,w\right),F\left(x',w'\right)\mid\mathcal{F}^{n}\right] \mbox{ for } 0\leq n\leq\infty.
\end{eqnarray*}
Observe that
\begin{eqnarray*}
V^{N}\left(s\right) & = & \mbox{max}_{x\in A} a_{N}\left(x\right) = \mbox{max}_{x\in A}\int_{w}\mu_{N}\left(x,w\right)dp\left(w\right),\\
R^{N-1}\left(s;(x,w)\right) & = & E\left[V^{N}\left(S^{N})\right)\mid S^{N-1}=s,(x_{N},w_{N})=(x,w)\right].
\end{eqnarray*}

We do not write the specific dependence on a policy $\pi$ when there is no confusion.

\subsection{Consistency of BQO for Finite Domains}
\label{proof_finite_case}

In this subsection, we prove a slightly stronger result (Theorem~\ref{main_theorem}) whose proof implies Theorem~\ref{main_theorem_simplified}. All the results of this subsection assume that both $A$ and $W$ are finite.

\begin{theorem}
\label{main_theorem}
Suppose that $A$ and $W$ are finite. For each $s\in\mathcal{H}$, we have that
\[
\lim_{N\rightarrow\infty}V^{0}\left(s;N\right)=\lim_{N\rightarrow\infty}V^{BQO,0}\left(s;N\right).
\]
\end{theorem}

\proof{} 

We first note that the limit of $S^{n}$ exists by \cref{convergence}. Denoted it by $S^{\infty}$. Furthermore, $V^{BQO,0}\left(s;N\right)$ is non-decreasing in $N$ and bounded above by \cref{policyincreasing}, and so by Fatou's lemma we have that $\mbox{lim }_{N\rightarrow\infty}V^{BQO,0}\left(s;N\right)=V^{BQO,0}\left(s;\infty\right)$.
By \cref{equalitylim}, under the BQO policy $V^{N}\left(S^{\infty}\right)=U\left(S^{\infty}\right)$  a.s. for all $N$. Thus,
\begin{eqnarray*}
V^{BQO,0}\left(s;\infty\right) & = & E^{BQO}\left[V^{N}\left(S^{\infty}\right)\mid S^{0}=s\right]\\
 & = & E^{BQO}\left[U\left(S^{\infty}\right)\mid S^{0}=s\right]\\
 & = & E^{BQO}[E\left[\mbox{max}_{x\in A}G\left(x\right)\mid S^{0}=S^{\infty}\right]\mid S^{0}=s]\\
 & = & E\left[\mbox{max}_{x\in A}G\left(x\right)\mid S^{0}=s\right]\\
 & = & U\left(s\right)
\end{eqnarray*}
and $U\left(s\right)\geq V^{0}\left(s;\infty\right)$
by \cref{policyincreasing}. This ends the proof because $V^{0}\left(s;\infty\right)\geq V^{BQO,0}\left(s;\infty\right)$.

\endproof 

\begin{lemma}
\label{convergence}

For any $s=\left(\mu,\Sigma\right)\in\mathcal{H}$, and policy $\pi$, we have that
%finite
%set $Q\subset A$, and policy $\pi$. Let $S_{Q}^{n}=\left(\mu_{Q,n},\Sigma_{Q,n}\right)$ where $\mu_{Q,n}:Q\times %W\rightarrow\mathbb{R}$
%and $\mu_{Q,n}\left(q,w\right):=\mu_{n}\left(q,w\right)$; similarly, $\Sigma_{Q,n}:\left(Q\times W\right)\times\left(Q\times %W\right)\rightarrow\mathbb{R}$,
%and $\Sigma_{Q,n}\left(\left(q,w\right),\left(q',w'\right)\right)=\Sigma_{n}\left(\left(q,w\right),\left(q',w'\right)\right)$
$\left(S^{n}\right)$
converges almost surely pointwise to $S^{\infty}=\left(\mu_{\infty},\Sigma_{\infty}\right) \in \mathcal{H}$, where $\mu_{\infty}\left(x,w\right)=\lim_{n}\mbox{ }\mu_{n}\left(x,w\right)$
and $\Sigma_{\infty}\left(\left(x,w\right),\left(x',w'\right)\right)=\lim_{n}\mbox{ }\Sigma_{n}\left(\left(x,w\right),\left(x',w'\right)\right)$.

\end{lemma}

\proof{}

This is proof is strongly based on \citet{frazier2009knowledge}. We first show that the limit exists for any pair of points $\left(x,w\right),\left(x',w'\right)$.
Let $M^{n}=\left(\mu^{n},\Sigma^{n}+\mu^{n}\left(\mu^{n}\right)^{\prime}\right)$
where $\mu^{n}=\left(\mu_{n}\left(x,w\right),\mu_{n}\left(x',w'\right)\right)^{\prime}$,
and 
\[
\Sigma^{n}:=\left(\begin{array}{cc}
\Sigma_{n}\left(\left(x,w\right),\left(x,w\right)\right) & \Sigma_{n}\left(\left(x,w\right),\left(x',w'\right)\right)\\
\Sigma_{n}\left(\left(x',w'\right),\left(x,w\right)\right) & \Sigma_{n}\left(\left(x',w'\right),\left(x',w'\right)\right)
\end{array}\right).
\]

We only need to show that $M^{n}$ converges a.s. since $\left(\mu^{n},\Sigma^{n}\right)$
is a linear transformation of $M^{n}$. We may write the components
of $M^{n}$ as the conditional expectation of an integrable random
variable with respect to $\mathcal{F}^{n}$ (the smallest $\sigma-$algebra
generated by the information at time $n$) by $\mu^{n}=E_{n}\left[\left(F\left(x,w\right),F\left(x',w'\right)\right)\right]^{\prime}$
and $\Sigma^{n}+\mu^{n}\left(\mu^{n}\right)^{\prime}=E_{n}\left[\left(F\left(x,w\right),F\left(x',w'\right)\right)^{\prime}\left(F\left(x,w\right),F\left(x',w'\right)\right)\right]$.
Thus $M^{n}$ is a uniformly integrable martingale and hence converges a.s. (see \citealt{Kallenberg1997}).
Thus the limit exists a.s. because the domain of $F$ is finite. Furthermore, since the limit of kernels is a kernel, we must have that $\Sigma_{\infty}$ is semi-positive definite. 

\endproof

\begin{lemma}
\label{equalitylim}

Under the BQO policy, if $s=\left(\mu_{0},\Sigma_{0}\right)\in\mathcal{H}_{1}$ and $A$ is finite,
then $V^{N}\left(S^{\infty}\right)=U\left(S^{\infty}\right)$
a.s. where $S^{\infty}$ is the limit of $\left(S^{n}\right)$,
and 
\begin{eqnarray*}
V^{n}\left(s\right) & = & \mbox{sup}_{\pi\in\Pi}E^{\pi}\left[\mbox{max}_{x\in A} a_{N}\left(x\right)\mid S^{n}=s\right]\\
U\left(s\right) & = & E\left[\mbox{max}_{x\in A}G\left(x\right)\mid S^{0}=s\right].
\end{eqnarray*}
\end{lemma}

\proof{}
Define the events $H_{(x,w)}:=\left\{ R^{N-1}\left(S^{\infty};(x,w)\right)>V^{N}\left(S^{\infty}\right)\right\} $
for any $x\in A$, $w\in W$, where $S^{\infty}$ is the limit of the parameters of the GP model, and define the Q-factors
\[
R^{N-1}\left(s;(x,w)\right):=E\left[V^{N}\left(S^{N}\right)\mid S^{N-1}=s,(x_{N},w_{N})=(x,w)\right].
\]

For any $B\subset A\times W$, we define the events:
\[
H_{B}:=\left[\bigcap_{(x,w)\in B}H_{(x,w)}\right]\bigcap\left[\bigcap_{(x,w)\notin B}H_{(x,w)}\right].
\]

By \cref{moreinfpolicy}, $R^{N-1}\left(s;(x,w)\right)\geq V^{N}\left(s\right)$,
and so $H_{B}$ is the event that $R^{N-1}\left(S^{\infty};(x,w)\right)>V^{N}\left(S^{\infty}\right)$
for $(x,w)\in B$, and $R^{N-1}\left(S^{\infty};(x,w)\right)=V^{N}\left(S^{\infty}\right)$
for $(x,w)\notin B$. We will show that $P\left[H_{B}\right]=0$ if $B\neq\emptyset$,
and so we will have that $P\left[H_{\emptyset}\right]=1$.

Let $B\neq\emptyset$. Assume that $P[H_{B}\bigcap\left\{ S^{n}\rightarrow S^{\infty}\right\}]>0$. By contraposition of \cref{measureio}, there exists a measurable set $\mathcal{L}$ such that $P[\mathcal{L}]>0$, and for each $\omega \in \mathcal{L}\bigcap \left\{ S^{n}\rightarrow S^{\infty}\right\}$, we have that there exists $K_{(x,w)}\left(\omega\right)\in\mathbb{N}$
for each $(x,w)\in B$ such that the BQO policy does not sample $(x,w)$ for
$n>K_{(x,w)}\left(\omega\right)$. 

Fix $\omega \in \mathcal{L}\bigcap \left\{ S^{n}\rightarrow S^{\infty}\right\}$, and define $K\left(\omega\right):=\mbox{max}_{(x,w)\in B}K_{(x,w)}\left(\omega\right)$. Given that $S^{n}\rightarrow S^{\infty}$ and $R^{N-1}\left(S^{\infty};(x,w)\right)>V^{N}\left(S^{\infty}\right)=R^{N-1}\left(S^{\infty};(y,r)\right)$ for $(x,w)\in B$ and $(y,r)\notin B$, there exists $n>K\left(\omega\right)$, such that

\[
\mbox{min}_{\left(x,w\right)\in B}R^{N-1}\left(S^{n}\left(\omega\right);\left(x,w\right)\right)>\mbox{max}_{\left(x,w\right)\notin B}R^{N-1}\left(S^{n}\left(\omega\right);\left(x,w\right)\right).
\]
Thus the BQO policy must sample from $\left(x,w\right)\in B$ at time
$n$, which contradicts the statement that the BQO policy never samples from $\left(x,w\right)\in B$
at time $n$. Consequently, $P\left[H_{\emptyset}\right]=1$, and
so $R^{N-1}\left(S^{\infty};\left(x,w\right)\right)=V^{N}\left(S^{\infty}\right)$
for all $(x,w)$ almost surely. By \cref{important_lemma}, we conclude that $V^{N}\left(S^{\infty}\right)=U\left(S^{\infty}\right)$ almost
surely, as we wanted to prove. 

\endproof

\begin{lemma}
\label{policyincreasing}
$\mbox{sup}_{\pi}E^{\pi}\left[\mbox{max}_{x\in A} a_{N}\left(x\right)\mid S^{0}=s\right]$
is non-decreasing in $N$ and is bounded above. For any policy $\pi$,
$E^{\pi}\left[\mbox{max}_{x\in A}a_{N}\left(x\right)\mid S^{0}=s\right]$
is non-decreasing in $N$ and is bounded above. Furthermore,  $V^{0}\left(s;N\right)\leq U\left(s\right)$.
\end{lemma}

\proof{}

First, we prove that $V^{0}\left(s;N\right)=\mbox{sup}_{\pi}E^{\pi}\left[\mbox{max}_{x\in A}a_{N}\left(x\right)\mid S^{0}=s\right]$
is non-decreasing in $N$ and bounded above. Observe that $V^{0}\left(s;N-1\right)=\mbox{sup}_{\pi}E^{\pi}\left[\mbox{max}_{x\in A} a_{N-1}\left(x\right)\mid S^{0}=s\right]=V^{1}\left(s;N\right),$
then
\[
V^{0}\left(s;N\right)-V^{0}\left(s;N-1\right)=V^{0}\left(s;N\right)-V^{1}\left(s;N\right),
\]
and this difference is not negative by \cref{moreinf}. This shows that
$V^{0}$ is non-decreasing. 

Now, we show that $V^{0}\left(s;N\right)\leq U\left(s\right)$.
We have that for any policy $\pi$,
\begin{eqnarray*}
E^{\pi}\left[\mbox{max}_{x\in A}a_{N}\left(x\right)\mid S^{0}=s\right] & = & E^{\pi}\left[\mbox{max}_{x\in A}E_{N}^{\pi}\left[G\left(x\right)\right]\mid S^{0}=s\right]\\
 & \leq & E^{\pi}\left[E_{N}^{\pi}\left[\mbox{max}_{x\in A}G\left(x\right)\right]\mid S^{0}=s\right]\\
 & = & E^{\pi}\left[\mbox{max}_{x\in A}G\left(x\right)\mid S^{0}=s\right]\\
 & = & E\left[\mbox{max}_{x\in A}G\left(x\right)\mid S^{0}=s\right]\\
 & = & U\left(s\right),
\end{eqnarray*}
and then $V^{0}\left(s;N\right)\leq U\left(s\right)$.

We now show that $V^{\pi,0}\left(s;N\right)=E^{\pi}\left[\mbox{max}_{x\in A}a_{N}\left(x\right)\mid S^{0}=s\right]$
is non-decreasing in $N$ for any stationary policy. We have that
$V^{\pi,0}\left(s;N-1\right)=E^{\pi}\left[\mbox{max}_{x\in A}a_{N-1}\left(x\right)\mid S^{0}=s\right]=V^{\pi,1}\left(s;N\right),$
and then
\[
V^{\pi, 0}\left(s;N\right)-V^{\pi, 0}\left(s;N-1\right)=V^{\pi, 0}\left(s;N\right)-V^{\pi, 1}\left(s;N\right),
\]
and this difference is not negative by \cref{moreinfpolicy}.

\endproof

\begin{proposition}
\label{moreinf}

For $s=\left(\mu,\Sigma\right)\in\mathcal{H}$
and $x\in A$,$w\in W$, we have that $R^{n-1}\left(s;(x,w)\right)\geq V^{n}\left(s\right)$ for all $0\leq n<N$. Furthermore, $V^{n+1}\left(s;N\right)\leq V^{n}\left(s;N\right)$ for all states $s$. 

\end{proposition}

\proof{}

This proof is based on a similar proposition in \citet{frazier2009knowledge}. We first show that $R^{n-1}\left(s;(x,w)\right)\geq V^{n}\left(s\right)$. We proceed by induction on $n$. For $n=N-1$, we have that $V^{N}\left(s\right)=\mbox{max}_{x\in A}a\left(x\right)$,
and by Jensen's inequality
\begin{eqnarray*}
R^{N-1}\left(s;(x,w)\right) & = & E\left[V^{N}\left(S^{N}\right)\mid S^{N-1}=s,(x_{N},w_{N})=(x,w)\right]\\
 & = & E\left[\mbox{max}_{z\in A}a_{N}\left(z\right)\mid S^{N-1}=s,(x_{N},w_{N})=(x,w)\right]\\
 & \geq & \mbox{max}_{z\in A}E\left[a_{N}\left(z\right)\mid S^{N-1}=s,(x_{N},w_{N})=(x,w)\right]\\
 & = & \mbox{max}_{z\in A}E[\int\mu\left(z,r\right)dp\left(r\right)+\sigmatilde(z,x,w)Z]\\
 &=&V^{N}\left(s\right),
\end{eqnarray*}
where the penultimate equality follows by \cref{postdist}.

Now, we prove the induction step. For $0\leq n<N-1$, 

\begin{eqnarray*}
R^{n}\left(s;\left(x,w\right)\right) & = & E\left[V^{n+1}\left(S^{n+1}\right)\mid S^{n}=s,(x_{n+1},w_{n+1})=\left(x,w\right)\right]\\
 & = & E\left[\mbox{sup}_{\left(x',w'\right)}R^{n+1}\left(S^{n+1};\left(x',w'\right)\right)\mid S^{n}=s,(x_{n+1},w_{n+1})=\left(x,w\right)\right]\\
 & \geq & \mbox{sup}_{\left(x',w'\right)}E\left[R^{n+1}\left(S^{n+1};\left(x',w'\right)\right)\mid S^{n}=s,(x_{n+1},w_{n+1})=\left(x,w\right)\right]\\
 & = & \mbox{sup}_{\left(x',w'\right)}E\left[V^{n+2}\left(S^{n+2}\right)\mid S^{n}=s,(x_{n+1},w_{n+1})=\left(x,w\right),(x_{n+2},w_{n+2})=\left(x',w'\right)\right]
\end{eqnarray*}
where the second equality follows from the dynamic programming principle.

In the last equation both points $\left(x,w\right)$ and $\left(x',w'\right)$ are inerchangeable, in the sense that it does not matter the order in which $\left(x,w\right)$ and $\left(x,w\right)$ are chosen. Thus, we have that
\begin{eqnarray*}
R^{n}\left(s;\left(x,w\right)\right) & \geq & \mbox{sup}_{\left(x',w'\right)}E\left[V^{n+2}\left(S^{n+2}\right)\mid S^{n}=s,(x_{n+1},w_{n+1})=\left(x',w'\right),(x_{n+2},w_{n+2})=\left(x,w\right)\right]\\
 & = & \mbox{sup}_{\left(x',w'\right)}E\left[E\left[V^{n+2}\left(S^{n+2}\right)\mid S^{n}=s,(x_{n+1},w_{n+1})=\left(x',w'\right),(x_{n+2},w_{n+2})=\left(x,w\right)\right]\mid Z_{2}\right]\\
 & = & \mbox{sup}_{\left(x',w'\right)}E\left[R^{n+1}\left(f_{n}\left(s,\left(x',w'\right),Z_{2}\right);\left(x,w\right)\right)\right]
\end{eqnarray*}
where $Z_{2}\sim N\left(0;\Sigma_{n}\left(\left(x',w'\right),\left(x',w'\right)\right) + \lambda_{(x',w')}\right)$,
and $f_{n}\left(s,\left(x',w'\right),Z_{2}\right)$ are the new parameters of the posterior Gaussian process after observing $Z_{2}$ and choosing $\left(x',w'\right)$.

By the induction hypothesis,
\begin{eqnarray*}
R^{n}\left(s;\left(x,w\right)\right) & \geq & \mbox{sup}_{\left(x',w'\right)}E\left[V^{n+2}\left(f_{n}\left(s,\left(x',w'\right),Z_{2}\right)\right)\right]\\
 & = & \mbox{sup}_{\left(x',w'\right)}R^{n+1}\left(s;\left(x',w'\right)\right)\\
 & = & V^{n+1}\left(s\right)
\end{eqnarray*}
as we wanted to prove.

Finally, take the extra measurement $(x,w)$ to be the measurement made by the optimal policy in state s, and thus by the first part of the proposition we conclude that $V^{n+1}\left(s;N\right)\leq V^{n}\left(s;N\right)$.

\endproof

\begin{proposition}
\label{moreinfpolicy}

Let $\pi$ be a stationary policy, and $s=(\mu, \Sigma)$. We have that $V^{\pi,n}\left(s;N\right)\geq V^{\pi,n+1}\left(s;N\right)$.

\end{proposition}

\proof{}
We proceed by induction on $n$. Consider the base case, $n=N-1.$
By Jensen's inequality we have
\begin{eqnarray*}
V^{\pi,N-1}\left(s;N\right) & = & E\left[\mbox{max}_{x\in A}\left(\int\mu\left(x,w\right)dp(w)+\frac{\int\Sigma\left(\left(x,w\right),\pi\left(s\right)\right)dp\left(w\right)}{\Sigma\left(\pi\left(s\right),\pi\left(s\right)\right) + \lambda_{\pi\left(s\right)}}Z\right)\right]\\
 & \geq & \mbox{max}_{x\in A}E\left[\left(\int\mu\left(x,w\right)dp(w)+\frac{\int\Sigma\left(\left(x,w\right),\pi\left(s\right) \right)dp\left(w\right)}{\Sigma\left(\pi\left(s\right),\pi\left(s\right)\right) + \lambda_{\pi\left(s\right)}}Z\right)\right]\\
 & = & \mbox{max}_{x\in A}a\left(x\right)\\
 & = & V^{\pi,N}\left(s;N\right)
\end{eqnarray*}
where $Z\sim N\left(0,\Sigma\left(\pi\left(s\right),\pi\left(s\right)\right) + \lambda_{\pi\left(s\right)}\right)$, and $\lambda_{\pi\left(s\right)}$ is the variance of the observations of $F(\pi\left(s\right))$.

Now consider the induction step. For $n<N-1$, we assume that if $n \leq m$, we have that $V^{\pi,m}\left(s;N\right)\geq V^{\pi,m+1}\left(s;N\right)$ for all parameters $s$. Thus,
\begin{eqnarray*}
V^{\pi,n}\left(s;N\right) & = & E^{\pi}\left[\mbox{max}_{x\in A}a_{N}\left(x\right)\mid S^{n}=s\right]\\
 & = & E\left[E^{\pi}\left[\mbox{max}_{x\in A}a_{N}\left(x\right)\mid S^{n}=s,Z_{n+1},\left(x_{n+1},w_{n+1}\right)=\pi\left(s\right)\right]\right]\\
 & \geq & E\left[E^{\pi}\left[\mbox{max}_{x\in A}a_{N}\left(x\right)\mid S^{n+1}=s,Z_{n+2},\left(x_{n+2},w_{n+2}\right)=\pi\left(s\right)\right]\right]\\
 & = & E^{\pi}\left[\mbox{max}_{x\in A}a_{N}\left(x\right)\mid S^{n+1}=s\right]\\
 & = & V^{\pi,n+1}\left(s;N\right),
\end{eqnarray*}
 as we wanted to prove.

\endproof

\begin{lemma}
\label{measureio}
If the policy $\pi$ samples from $(x,w)$ infinitely often almost surely, then $R^{N-1}\left(S^{\infty};\left(x,w\right)\right) = V^{N}\left(S^{\infty}\right)$ almost surely under $\pi$.

\end{lemma}

\proof{}
Note that there exists a sequence of i.i.d. normal random variables $\left\{\epsilon_{k}\right\}_{k}$ with mean zero and variance $\lambda_{(x,w)}$ such that we observe $F(x,w)+\epsilon_{k}$ the $k$-th time that $F(x,w)$ is queried. By the strong law of large numbers, we have that there exists a measurable set $\mathcal{L}$ of probability one such that $\frac{1}{m}\sum_{k=1}^{m}\epsilon_{k}$ converges to the zero random variable as $m$
goes to infinity. By hypothesis, there exists a measurable set $\mathcal{G}$ with probability one such that $\pi$ samples from $(x,w)$ infinitely often. Consequently, $\mathcal{G} \bigcap \mathcal{L}$ has probability one. 

Fix $\omega \in \mathcal{G} \bigcap \mathcal{L}$. Consider the observations $\left\{y_{n_{m}}\right\}_{m>0}$  where $(x_{n_{m}},w_{n_{m}})=(x,w)$, and  $y_{n_{m}}:= F(x,w) + \epsilon_{m}$. So, we have that $y_{n_{m}}$ converges to $F(x,w)$, and then the posterior distribution of $G$ given $\mathcal{F}^{\infty}$ does not depend on the noisy observations of $F(x,w)$.  Consequently, we have that if $\epsilon\sim N\left(0,\lambda_{\left(x,w\right)}\right)$, then

\begin{eqnarray*}
R^{N-1}\left(S^{\infty};\left(x,w\right)\right) & = & E\left[\mbox{max}_{x'}E\left[G\left(x'\right)\mid\mathcal{F}^{\infty},F\left(x,w\right)+\epsilon\right]\mid\mathcal{F}^{\infty}\right]\\
 & = & E\left[\mbox{max}_{x'}E\left[G\left(x'\right)\mid\mathcal{F}^{\infty}\right]\mid\mathcal{F}^{\infty}\right]\\
 & = & \mbox{max}_{x'}E\left[G\left(x'\right)\mid\mathcal{F}^{\infty}\right]\\
 & = & V^{N}\left(S^{\infty}\right).
\end{eqnarray*}
This ends the proof.

\endproof

\begin{lemma}
\label{important_lemma}

Let $S=\left(\mu,\Sigma\right)\in\mathcal{H}$. If $A$ is finite and $R^{N-1}\left(S;(x,w)\right)=V^{N}\left(S\right)$
for all $(x,w)$, then
$V^{N}\left(S\right)=U\left(S\right)$.
\end{lemma}

\proof{}

Suppose that $A=\{x_{1},\ldots,x_{M}\}$ and $W=\{w_{1},\ldots,w_{q}\}$. Fix any $x\in A$ and $w\in W$. We will show that $\sigma_{i}\left(\Sigma,(x,w)\right)=\sigma_{1}\left(\Sigma,(x,w)\right)$
for every $i$, where
\[
\sigma_{i}\left(\Sigma,(x,w)\right)=\frac{\int\Sigma\left(\left(x_{i},r\right),\left(x,w\right)\right)dp\left(r\right)}{\sqrt{\Sigma\left((x,w),(x,w)\right)+\lambda_{(x,w)}}}.
\]
We reorder the index set $\left\{ 1,\ldots,M\right\} $ such that
$a\left(x_{1}\right)=\mbox{max}_{i}\mbox{ }a\left(x_{i}\right)=V^{N}\left(s\right)$.
For a standard univariate normal random variable $Z$, we have that
\begin{eqnarray*}
0 & = & R^{N-1}\left(S;(x,w)\right)-V^{N}\left(S\right)\\
 & = & E\left[\mbox{max}_{i}\left(a\left(x_{i}\right)+\sigma_{i}\left(\Sigma,(x,w)\right)Z\right)\right]-a\left(x_{1}\right)\\
 & = & E\left[\mbox{max}_{i}\left[a\left(x_{i}\right)-a\left(x_{1}\right)+\left(\sigma_{i}\left(\Sigma,(x,w)\right)-\sigma_{1}\left(\Sigma,(x,w)\right)\right)Z\right]\right]+E\left[\sigma_{1}\left(\Sigma,(x,w)\right)Z\right]\\
 & = & E\left[\mbox{max}_{i}\left[a\left(x_{i}\right)-a\left(x_{1}\right)+\left(\sigma_{i}\left(\Sigma,(x,w)\right)-\sigma_{1}\left(\Sigma,(x,w)\right)\right)Z\right]\right].
\end{eqnarray*}

Thus, using that $\mbox{max}_{i}\left[a\left(x_{i}\right)-a\left(x_{1}\right)+\left(\sigma_{i}\left(\Sigma,(x,w)\right)-\sigma_{1}\left(\Sigma,(x,w)\right)\right)Z\right] \geq 0$, we have that
\[
\int_{0}^{\infty}P\left[\mbox{max}_{i}\left[a\left(x_{i}\right)-a\left(x_{1}\right)+\left(\sigma_{i}\left(\Sigma,(x,w)\right)-\sigma_{1}\left(\Sigma,(x,w)\right)\right)Z\right]\geq u\right]du=0
\]
which implies that $P\left[\mbox{max}_{i}\left[a\left(x_{i}\right)-a\left(x_{1}\right)+\left(\left(\sigma_{i}\left(\Sigma,(x,w)\right)-\sigma_{1}\left(\Sigma,(x,w)\right)\right)\right)Z\right]\geq u\right]=0$
for almost every $u$ in $\left[0,\infty\right).$ Taking the limit
as $u\rightarrow0$, by the bounded convergence theorem, we have that
\begin{eqnarray*}
0 & = & \mbox{lim}_{u\rightarrow0}P\left[\mbox{max}_{i}\left[a\left(x_{i}\right)-a\left(x_{1}\right)+\left(\left(\sigma_{i}\left(\Sigma,(x,w)\right)-\sigma_{1}\left(\Sigma,(x,w)\right)\right)\right)Z\right]\geq u\right]\\
 & = & P\left[\mbox{max}_{i}\left[a\left(x_{i}\right)-a\left(x_{1}\right)+\left(\left(\sigma_{i}\left(\Sigma,(x,w)\right)-\sigma_{1}\left(\Sigma,(x,w)\right)\right)\right)Z\right]>0\right]
\end{eqnarray*}
and so $\mbox{max}_{i}\left[a\left(x_{i}\right)-a\left(x_{1}\right)+\left(\left(\sigma_{i}\left(\Sigma,(x,w)\right)-\sigma_{1}\left(\Sigma,(x,w)\right)\right)\right)Z\right]=0$
a.s., which implies that $\sigma_{i}\left(\Sigma,(x,w)\right)=\sigma_{1}\left(\Sigma,(x,w)\right)$.

For all $y,w,x_{i},x_{j}$, we have that
\[
\int\Sigma\left(\left(y,w\right),\left(x_{i},r\right)\right)dp\left(r\right)=\int\Sigma\left(\left(y,w\right),\left(x_{j},r\right)\right)dp\left(r\right)
\]
and so
\begin{eqnarray*}
\int\Sigma_{1}\left(\left(y,w\right),\left(x_{i},r\right)\right)dp\left(r\right) & = & \int\Sigma\left(\left(y,w\right),\left(x_{i},r\right)\right)dp\left(r\right)-\frac{\Sigma\left(\left(y,w\right),\left(x_{1},w_{1}\right)\right)\int\Sigma\left(\left(x_{1},w_{1}\right),\left(x_{i},r\right)\right)dp\left(r\right)}{\sqrt{\Sigma\left(\left(x_{1},w_{1}\right),\left(x_{1},w_{1}\right)\right)+\lambda_{\left(x_{1},w_{1}\right)}}}\\
 & = & \int\Sigma\left(\left(y,w\right),\left(x_{j},r\right)\right)dp\left(r\right)-\frac{\Sigma\left(\left(y,w\right),\left(x_{1},w_{1}\right)\right)\int\Sigma\left(\left(x_{1},w_{1}\right),\left(x_{j},r\right)\right)dp\left(r\right)}{\sqrt{\Sigma\left(\left(x_{1},w_{1}\right),\left(x_{1},w_{1}\right)\right)+\lambda_{\left(x_{1},w_{1}\right)}}}\\
 & = & \int\Sigma_{1}\left(\left(y,w\right),\left(x_{j},r\right)\right)dp\left(r\right).
\end{eqnarray*}
By recursion, we conclude that for all $n\leq N$, $y,w,x_{i},x_{j}$,
we have that 
\[
\int\Sigma_{n}\left(\left(y,w\right),\left(x_{i},r\right)\right)dp\left(r\right)=\int\Sigma_{n}\left(\left(y,w\right),\left(x_{j},r\right)\right)dp\left(r\right).
\]
Consequently, we have that for all $y,x_{i},x_{j}$
\[
\int\int\Sigma_{n}\left(\left(y,w\right),\left(x_{i},r\right)\right)dp\left(r\right)dp\left(w\right)=\int\int\Sigma_{n}\left(\left(y,w\right),\left(x_{j},r\right)\right)dp\left(r\right)dp\left(w\right).
\]
Denote $\left(x_{N},w_{N}\right)$ by $\left(x,w\right)$, we
have that for all $x_{i},x_{j},x'$,
\[
\int\int\Sigma_{n}\left(\left(x,w\right),\left(x_{i},r\right)\right)dp\left(r\right)dp\left(w\right)=\int\int\Sigma_{n}\left(\left(x,w\right),\left(x_{j},r\right)\right)dp\left(r\right)dp\left(w\right),
\]
\begin{eqnarray*}
\int\int\Sigma_{n}\left(\left(x',w\right),\left(x_{i},r\right)\right)dp\left(r\right)dp\left(w\right) & = & \int\int\Sigma_{n}\left(\left(x,w\right),\left(x_{j},r\right)\right)dp\left(r\right)dp\left(w\right)\\
 & = & \int\int\Sigma_{n}\left(\left(x,w\right),\left(x,r\right)\right)dp\left(r\right)dp\left(w\right)
\end{eqnarray*}

Consequently, the vector $\left(G\left(z\right):z\in A\right)$ has
a covariance matrix with all entries equal to 

\begin{equation*}
\int\int\Sigma_{N}\left(\left(x,w\right),\left(x,r\right)\right)dp\left(r\right)dp\left(w\right).
\end{equation*}

Now define a normal random vector $W\left(y\right)=a_{N}\left(y\right)-a_{N}\left(x\right)+G\left(x\right)$. Conditioned on $\mathcal{F}^{N}$,
it has mean vector $a_{N}$, and covariance matrix with all entries
equal to $\int\int\Sigma_{N}\left((x,u),(x,s)\right)dp(u)dp(s)$. Consequently, $W$ is
equal in distribution to $\left(G\left(y\right):y\in A \right)$.

We then have that
\begin{eqnarray*}
U\left(S^{N}\right) & = & E_{N}\left[\mbox{max}_{x\in A}G\left(x\right)\right]=E_{N}\left[\mbox{max}_{y\in A}W\left(y\right)\right]=E_{N}\left[\mbox{max}_{y\in A}\left(a_{N}\left(y\right)-a_{N}\left(x\right)+G\left(x\right)\right)\right]\\
 & = & E_{N}\left[\mbox{max}_{y\in A} a^{N}\left(y\right)\right]=\mbox{max}_{y\in A}a^{N}\left(y\right)=V^{N}\left(S^{N}\right).
\end{eqnarray*}
Then
\begin{eqnarray*}
V^{N}\left(S\right) & = & R^{N-1}\left(S;(x,w)\right)=E\left[V^{N}\left(S^{N}\right)\mid x_{N}=x,w_{N}=w,S^{N-1}=S\right]\\
 & = & E\left[U\left(S^{N}\right)\mid x_{N}=x,w_{N}=w, S^{N-1}=S\right]\\
 & = & E\left[E_{N}\left[\mbox{max}_{x\in A}G\left(x\right)\right]\mid x_{N}=x,w_{N}=w, S^{N-1}=S\right]\\
 & = & E[\mbox{max}_{x\in A}G\left(x\right)\mid x_{N}=x,w_{N}=w,S^{N-1}=S]\\
 & = & U\left(S\right),
\end{eqnarray*}
which ends the proof. 
\endproof

\subsection{Consistency of BQO for Continuum Domains}
\label{proof_continuum_case}

In this subsection, we prove a slightly stronger result (Theorem~\ref{continuum_th}) whose proof implies  Theorem~\ref{continuum_th_intuitive}.

\begin{theorem}
\label{continuum_th}
Suppose that $A=\left[a_{1},b_{1}\right]\times\cdots\times\left[a_{\dimension},b_{\dimension}\right]\subset\mathbb{R}^{\dimension}$,
$a_{i}<b_{i}$ for all $i$, and $\left|W\right|=m$. Let $s=\left(\mu,\Sigma\right)\in\mathcal{H_{0}}$. We assume that the function
$g_{w}\left(x\right):=\lambda_{\left(w,x\right)}$ is continuous in
$A$ for all $w\in W$, and there exist $k_{\lambda},K_{\lambda}>0$
such that $k_{\lambda}<\lambda_{\left(w,x\right)}<K_{\lambda}$ for
all $w\in W$ and $x\in A$. We then have that

\[
\lim_{N\rightarrow\infty}V^{0}\left(s;N\right)=\lim_{N\rightarrow\infty}V^{BQO,0}\left(s;N\right)
\]

\end{theorem}

As we did in the appendix $\mathsection$\ref{proof_finite_case}, we analyze the problem with a dynamic programming framework, and use the same notation defined there. In addition, we assume without loss of generality that $\int p(w)dw=1$. All the results of this subsection assume that $W$ is finite.

We introduce the following notation:

\begin{eqnarray*}
Q & \mbox{is} & \mbox{a finite set of A}\\
V_{Q}^{n}\left(s\right) & = & \mbox{sup}_{\pi\in\Pi}E^{\pi}\left[\mbox{max}_{x\in Q}a_{N}\left(x\right)\mid S^{n}=s\right]=:V_{Q}^{n}\left(s;N\right)\\
U_{Q}\left(s\right) & = & E\left[\mbox{max}_{x\in Q}G\left(x\right)\mid S^{0}=s\right]\\
V_{Q}^{\pi,n}\left(s;N\right) & = & E^{\pi}\left[\mbox{max}_{x\in Q}a_{N}\left(x\right)\mid S^{n}=s\right]:=V_{Q}^{\pi,n}\left(s\right)\\
R_{Q}^{n}\left(s;(x,w)\right) & = & E\left[V_{Q}^{n+1}\left(S^{n+1}\right)\mid S^{n}=s,x_{n+1}=x,w_{n+1}=w\right]\\
\mu_{Q}  & \mbox{ is} & \mbox{the function }\mu \mbox{ where the first entry is restricted to } Q\\
% \Sigma_{Q} & \mbox{ is} & \mbox{the function }\Sigma \mbox{ where the corresponding entries to } A \mbox{ are restricted to }Q\\
S_{Q} &=& \left(\mu_{Q},\Sigma\right)\\
L_{n}\left(x,y,z\right)&=&\int\Sigma_{n}\left(x,w,y,z\right)dp\left(w\right)
\end{eqnarray*}
Observe that
\begin{eqnarray*}
V_{Q}^{N}\left(s\right) & = & \mbox{max}_{x\in Q}a_{N}\left(x\right)\\
V_{Q}^{\pi,N}\left(s;N\right) & = & V_{Q}^{\pi,N}\left(s_{Q};N\right)\\
R_{Q}^{N-1}\left(s;(x,w)\right) & = & E\left[V_{Q}^{N}\left(S^{N}\right)\mid S^{N-1}=s,x_{N}=x,w_{N}=w\right].
\end{eqnarray*}

We should note that \cref{policyincreasing} proved in the appendix $\mathsection$\ref{proof_finite_case} under the assumption that we can only choose a finite number of alternatives, holds under the assumption that we are optimizing over a finite set and we can choose any alternative in a compact set.
\\
\\
\textbf{Proof of \cref{continuum_th}.}
\proof{}
Assume without loss of generality that $A=\left[0,1\right]$. Let
$\left\{ Q_{m}\right\} _{m\geq1}$ be an increasing sequence of sets
defined by 
\[
Q_{m}=\bigcup_{n\geq1}^{m}\bigcup_{l=0}^{n}\left\{ \frac{l}{n}\right\} \subset A.
\]
It is clear that $\bigcup_{m\geq1}Q_{m}=\mathbb{Q}\bigcap\left[0,1\right]$,
and $\left|Q_{m}\right|<\infty.$ Let $\left\{ b\left(m,N,\pi\right)\right\} _{N\geq0,m\geq1,\pi\in\Pi}$
be a sequence of real numbers defined by
\[
b\left(m,N,\pi\right)=E^{\pi}\left[\mbox{max}_{x\in Q_{m}}a_{N}\left(x\right)\mid S^{0}=s\right].
\]

By \cref{approxdomain} and the Monotone Convergence Theorem applied to $\left\{ \mbox{max}_{x\in Q_{m}}a_{N}\left(x\right)-\mbox{max}_{x\in Q_{1}}a_{N}\left(x\right)\right\} $ we have that (note that the expectation of $\mbox{max}_{x\in Q_{m}}a_{N}\left(x\right)$ is finite by \cref{intpostmean}) 
\begin{eqnarray*}
\mbox{lim}_{N}\mbox{sup}_{\pi}E^{\pi}\left[\mbox{max}_{x\in A}a_{N}\left(x\right)\mid S^{0}=s\right] & = & \mbox{lim}_{N}\mbox{sup}_{\pi}E^{\pi}\left[\mbox{lim}_{m}\mbox{max}_{x\in Q_{m}}a_{N}\left(x\right)\mid S^{0}=s\right]\\
 & = & \mbox{lim}_{N}\mbox{sup}_{\pi}\mbox{lim}_{m}E^{\pi}\left[\mbox{max}_{x\in Q_{m}}a_{N}\left(x\right)\mid S^{0}=s\right]\\
 & = & \mbox{lim}_{N}\mbox{sup}_{\pi}\mbox{sup}_{m}E^{\pi}\left[\mbox{max}_{x\in Q_{m}}a_{N}\left(x\right)\mid S^{0}=s\right]\\
 & = & \mbox{lim}_{N}\mbox{sup}_{m}\mbox{sup}_{\pi}E^{\pi}\left[\mbox{max}_{x\in Q_{m}}a_{N}\left(x\right)\mid S^{0}=s\right]\\
 & = & \mbox{lim}_{N}\mbox{lim}_{m}\mbox{sup}_{\pi}b\left(m,N,\pi\right)
\end{eqnarray*}

By \cref{policyincreasing}, we have that
\[
\mbox{lim}_{N}\mbox{lim}_{m}\mbox{sup}_{\pi}b\left(m,N,\pi\right)=\mbox{sup}_{N}\mbox{sup}_{m}\mbox{sup}_{\pi}b\left(m,N,\pi\right)
\]
and then
\begin{equation}
\mbox{lim}_{N}\mbox{lim}_{m}\mbox{sup}_{\pi}b\left(m,N,\pi\right)=\mbox{lim}_{m}\mbox{lim}_{N}\mbox{sup}_{\pi}b\left(m,N,\pi\right),\label{eq:1}
\end{equation}
thus by \cref{finitethm} and \cref{policyincreasing}, 
\begin{eqnarray*}
\mbox{lim}_{N}\mbox{sup}_{\pi}E^{\pi}\left[\mbox{max}_{x\in A}a_{N}\left(x\right)\mid S^{0}=s\right] & = & \mbox{lim}_{m}\mbox{lim}_{N}\mbox{sup}_{\pi}b\left(m,N,\pi\right)\\
 & = & \mbox{lim}_{m}\mbox{lim}_{N}b\left(m,N,BQO\right)\\
 & = & \mbox{sup}_{m}\mbox{sup}_{N}b\left(m,N,BQO\right)\\
 & = & \mbox{lim}_{N}\mbox{lim}_{m}b\left(m,N,BQO\right).
\end{eqnarray*}
By the Monotone Convergence Theorem applied to $\left\{ \mbox{max}_{x\in Q_{m}}a_{N}\left(x\right)-\mbox{max}_{x\in Q_{1}}a_{N}\left(x\right)\right\} $ 
and by \cref{approxdomain}, we have that
\begin{eqnarray*}
\mbox{lim}_{N}\mbox{sup}_{\pi}E^{\pi}\left[\mbox{max}_{x\in A}a_{N}\left(x\right)\mid S^{0}=s\right] & = & \mbox{lim}_{N}E^{BQO}\left[\mbox{lim}_{m}\mbox{max}_{x\in Q_{m}}a_{N}\left(x\right)\mid S^{0}=s\right]\\
 & = & \mbox{lim}_{N}E^{BQO}\left[\mbox{max}_{x\in A}a_{N}\left(x\right)\mid S^{0}=s\right].
\end{eqnarray*}

\endproof 

\begin{lemma}
\label{approxdomain}
Assume that $A=[0,1]$, and $\left|W\right|<\infty$.
$E^{\pi}\left[\lim_{m}\max_{x\in Q_{m}}a_{N}\left(x\right)\mid S^{0}=s\right]=E^{\pi}\left[\max_{x\in A}a_{N}\left(x\right)\mid S^{0}=s\right]$, 
where $\{Q_{m}\}_{m}$ are the sets defined in the proof of \cref{continuum_th}.

\end{lemma}

\proof{}

Let $\omega\in\varOmega$ and $\epsilon>0$. There exists $x_{0}\in A$
such that
\[
\mbox{max}_{x\in A}a_{N}\left(x\right)=a_{N}\left(x_{0}\right)
\]
because $a_{N}$ is a continuous function in a compact set. Let
$\delta>0$ such that $\left|a_{N}\left(x_{0}\right)-a_{N}\left(x\right)\right|<\epsilon$
whenever $\left|x_{0}-x\right|<\delta$. Using the completeness of
the rationals, we have that there exist $M$ ,$\left\{ q_{m}\right\} _{m>0}\subset\mathbb{Q}\bigcap[0,1]$,
and $\left\{ r_{m}\right\} _{m\geq1}\subset\mathbb{N}$ such that
$q_{m}\in Q_{r_{m}}$, $r_{m}\leq r_{m+1}$, and if $m\geq M$,
\[
\left|q_{m}-x_{0}\right|<\delta,
\]
and then
\[
\left|a_{N}\left(x_{0}\right)-a_{N}\left(q_{m}\right)\right|<\epsilon,
\]
which implies that if $m\geq r_{M}$, 
\[
\left|\mbox{max}_{x\in Q_{m}}a_{N}\left(x\right)-\mbox{max}_{x\in A}a_{N}\left(x\right)\right|<\epsilon,
\]
and then $\mbox{max}_{x\in Q_{m}}a_{N}\left(x\right)\rightarrow\mbox{max}_{x\in A}a_{N}\left(x\right)$
almost surely.

\endproof

\begin{lemma}
\label{intpostmean}

Let $\pi$ be an admissible policy, and $s_{0}\in \mathcal{H}_{1}$. We have that, given a finite or compact set $Q$ of $A$ and $\left|W\right|<\infty$,
\[
E^{\pi}\left[\sup_{x\in Q}a_{N}\left(x\right)\right]<\infty.
\]
\end{lemma}
\proof{}
Let $\phi_{i}\left(z_{i}\right)$ be the density of
a normal random variable with mean zero and variance $\Sigma_{i-1}\left(x_{i},w_{i},x_{i},w_{i}\right) + \lambda_{\left(x_{i},w_{i}\right)}$
where $(x_{i},w_{i})$ is the point chosen by $\pi$ at iteration $i$, we have that

\begin{eqnarray}
E^{\pi}\left[\mbox{sup}_{x\in Q}\left|a_{N}\left(x\right)\right|\right] & \leq & E^{\pi}\left[\mbox{sup}_{x\in Q}\left|a_{N-1}\left(x\right)\right|\right]+\nonumber \\
 &  & \int_{z_{1},\ldots z_{N}}\mbox{sup}_{x\in Q}\left|\frac{L_{N-1}\left(x,x_{N},w_{N}\right)}{\Sigma_{N-1}\left(x_{N},w_{N},x_{N},w_{N}\right)+\lambda_{\left(x_{N},w_{N}\right)}}\right|\Pi_{i=1}^{N-1}\phi_{i}\left(z_{i}\right)\left|z_{N}\right|\phi_{N}\left(z_{N}\right)dz_{1}\cdots d_{z_{N}}\nonumber \\
 & \leq & E^{\pi}\left[\mbox{sup}_{x\in Q}\left|a_{N-1}\left(x\right)\right|\right]+\nonumber \\
 &  & \int_{z_{1},\ldots z_{N-1}}\mbox{sup}_{x\in Q}\left|\frac{L_{N-1}\left(x,x_{N},w_{N}\right)}{\Sigma_{N-1}\left(x_{N},w_{N},x_{N},w_{N}\right)+\lambda_{\left(x_{N},w_{N}\right)}}\right|\Pi_{i=1}^{N-1}\phi_{i}\left(z_{i}\right)dz_{1}\cdots d_{z_{N-1}} \nonumber \\
 & & \int_{z_{N}}\left|z_{N}\right|\phi_{N}\left(z_{N}\right)d_{z_{N}}\nonumber \\
 & \leq & E^{\pi}\left[\mbox{sup}_{x\in Q}\left|a_{N-1}\left(x\right)\right|\right]+\nonumber \\
 &  & \sqrt{\frac{2}{\pi}}\int_{z_{1},\ldots z_{N-1}}\mbox{sup}_{x\in Q}\left|\frac{L_{N-1}\left(x,x_{N},w_{N}\right)}{\sqrt{\Sigma_{N-1}\left(x_{N},w_{N},x_{N},w_{N}\right)+\lambda_{\left(x_{N},w_{N}\right)}}}\right|\Pi_{i=1}^{N-1}\phi_{i}\left(z_{i}\right)dz_{1}\cdots d_{z_{N-1}}\nonumber \\
 & \leq & E^{\pi}\left[\mbox{sup}_{x\in Q}\left|a_{N-1}\left(x\right)\right|\right]+\nonumber \\
 &  & \sqrt{\frac{2}{\pi}}\int_{z_{1},\ldots z_{N-1}}\mbox{sup}_{x\in Q,w\in W}\sqrt{\Sigma_{N-1}\left(x,w,x,w\right)}\Pi_{i=1}^{N-1}\phi_{i}\left(z_{i}\right)dz_{1}\cdots d_{z_{N-1}}\nonumber \\
 & \leq & E^{\pi}\left[\mbox{sup}_{x\in Q}\left|a_{N-1}\left(x\right)\right|\right]+\sqrt{\frac{2}{\pi}}\sqrt{\mbox{sup}_{x\in Q,w\in W}\Sigma_{0}\left(x,w,x,w\right)}\label{eq:}\\
 & \vdots\nonumber \\
 & \leq & \mbox{sup}_{x\in Q}a_{0}(x) + N\sqrt{\frac{2}{\pi}}\sqrt{\mbox{sup}_{x\in Q, w\in W}\Sigma_{0}\left(x,w,x,w\right)}\label{eq:2},
\end{eqnarray} 
where $L_{n}(x,y,z)=\int\Sigma_{n}\left(x,w,y,z\right)dp\left(w\right).$ 
This shows that $E^{\pi}\left[\mbox{sup}_{x\in Q}a_{N}\left(x\right)\right]<\infty$. 

\endproof

\begin{theorem}
\label{finitethm}
Let $Q$ be a finite set of $A$. Suppose that $|W|<\infty$, and $s\in\mathcal{H}_{1}$. We have that
\[
\lim_{N\rightarrow\infty}\sup_{\pi}E^{\pi}\left[\max_{x\in Q}a_{N}\left(x\right)\mid S^{0}=s\right]=\lim_{N\rightarrow\infty}E^{BQO}\left[\max_{x\in Q}a_{N}\left(x\right)\mid S^{0}=s\right]
\]

\end{theorem}

\proof{}

%Let $S_{Q}^{n}$ be defined as in \cref{convergence}.
%\cref{convergence} shows that $S_{Q}^{\infty}$ exists. 
We first note that the limit of $S_{Q}^{n}$ exists a.s. by \cref{convergence_cont}, and we denote it by $S_{Q}^{\infty}$.
By \cref{equalitylim_cont}, under the BQO policy $V_{Q}^{N}\left(S_{Q}^{\infty}\right)=U_{Q}\left(S_{Q}^{\infty}\right)$
a.s. Furthermore, $V_{Q}^{BQO,0}\left(s;N\right)$ is non-decreasing in $N$ and bounded above by \cref{policyincreasing}, and so by Fatou's lemma we have that $\mbox{lim }_{N\rightarrow\infty}V^{BQO,0}\left(s;N\right)=V^{BQO,0}\left(s;\infty\right)$. Thus, for any $N>0$,
\begin{eqnarray*}
V_{Q}^{BQO,0}\left(s;\infty\right) & = & E^{BQO}\left[V_{Q}^{N}\left(S_{Q}^{\infty}\right)\mid S^{0}=s\right]\\
 & = & E^{BQO}\left[U_{Q}\left(S_{Q}^{\infty}\right)\mid S^{0}=s\right]\\
 & = & E^{BQO}[E\left[\mbox{max}_{x\in Q}G\left(x\right)\mid S^{0}=S_{Q}^{\infty}\right]\mid S^{0}=s]\\
 & = & E\left[\mbox{max}_{x\in Q}G\left(x\right)\mid S^{0}=s\right]\\
 & = & U_{Q}\left(s\right)
\end{eqnarray*}
and $U_{Q}\left(s\right)\geq V_{Q}^{0}\left(s;\infty\right)$
by \cref{policyincreasing}. This ends the proof because $V_{Q}^{0}\left(s;\infty\right)\geq V_{Q}^{BQO,0}\left(s;\infty\right)$. 

\endproof

\begin{lemma}
\label{rkhs}
Let $\mathcal{H}$ be the Reproducing Kernel Hilbert Space (RKHS)
associated to any isotropic kernel $\Sigma_{0}$ defined in $(\mathbb{R}^{d}\times W)\times(\mathbb{R}^{d}\times W)$, where $|W|<\infty$, we have the following:
\begin{enumerate}
\item Consider the operator $\bar{P}_{1:n}:\mathcal{H}\rightarrow\underset{t=1:n}{\mbox{span}}\left\{ \Sigma_{0}\left(x_{t},w_{t},\cdot,\cdot\right)\right\} $
$\subset\mathcal{H}$ defined by
\[
\bar{P}_{1:n}h:=\Sigma_{0}\left(\cdot,(x,w)_{1:n}\right)A_{n}^{-1}\left\langle \Sigma_{0}\left(\cdot,(x,w)_{1:n}\right),h\right\rangle 
\]
where $\Sigma_{0}\left(\cdot,(x,w)_{1:n}\right)=\left(\Sigma_{0}\left(\cdot,x_{1},w_{1}\right),\ldots,\Sigma_{0}\left(\cdot,x_{n},w_{n}\right)\right)$,
$A_{n}:=\left[\Sigma_{0}\left(x_{i},w_{i},x_{j},w_{j}\right)\right]_{i,j=1:n}$
and
\[
\left\langle \Sigma_{0}\left(\cdot,(x,w)_{1:n}\right),h\right\rangle :=\left[\begin{array}{c}
h\left(x_{1},w_{1}\right)\\
\vdots\\
h\left(x_{n},w_{n}\right)
\end{array}\right].
\]
We have that $\left\Vert \bar{P}_{1:n}\right\Vert \leq1$, and $\left\Vert 1-\bar{P}_{1:n}\right\Vert \leq1$
\item Suppose that $\Sigma_{w,w'}\left(x,y\right):=\Sigma_{0}\left(x,w,y,w'\right)$ is continuous for any $w,w'\in W$. Consider the operator $P_{1:n}:\mathcal{H}\rightarrow\underset{t=1:n}{\mbox{span}}\left\{ \Sigma_{0}\left(x_{t},w_{t},\cdot\right)\right\} $
$\subset\mathcal{H}$ defined by
\[
P_{1:n}h:=\Sigma_{0}\left(\cdot,(x,w)_{1:n}\right)A_{n}^{-1}\left\langle \Sigma_{0}\left(\cdot,(x,w)_{1:n}\right),h\right\rangle 
\]
where $\Sigma_{0}\left(\cdot,(x,w)_{1:n}\right)=\left(\Sigma_{0}\left(\cdot,x_{1},w_{1}\right),\ldots,\Sigma_{0}\left(\cdot,x_{n},w_{n}\right)\right)$,
$A_{n}:=[\Sigma_{0}\left(x_{i},w_{i},x_{j},w_{j}\right)+ 1_{\{(x_{i},w_{i})=(x_{j},w_{j})\}}\lambda_{\left(x_{i},w_{i}\right)}]_{i,j=1:n}$
and
\[
\left\langle \Sigma_{0}\left(\cdot,(x,w)_{1:n}\right),h\right\rangle :=\left[\begin{array}{c}
h\left(x_{1},w_{1}\right)\\
\vdots\\
h\left(x_{n},w_{n}\right)
\end{array}\right].
\]
We have that $\left\Vert P_{1:n}\Sigma_{0}\left(x',w',\cdot\right)\right\Vert ^{2}\leq\mbox{sup}_{x\in A,w\in W}\left[\Sigma_{0}\left(x,w,x,w\right)+\lambda_{\left(x,w\right)}\right]$, and $\left\Vert \left(1-P_{1:n}\right)\Sigma_{0}\left(x',w',\cdot\right)\right\Vert ^{2}\leq\mbox{sup}_{x\in A,w\in W}\left[\Sigma_{0}\left(x,w,x,w\right)+\lambda_{\left(x,w\right)}\right]+\mbox{sup}_{x\in A, w\in W}\Sigma_{0}\left(x,w,x,w\right)$ for all $x',w'$.
\item $\Sigma_{n}\left(\cdot,y,w\right)=\Sigma_{0}\left(\cdot,y,w\right) - P_{1:n}h_{y,w}$ where $h_{y,w}\left(z,w'\right):=\Sigma_{0}\left(z,w',y,w\right)$
\item $\Sigma_{n}\left(x,w,x,w\right)=\left\langle \Sigma_{0}\left(x,w,\cdot\right),\left(1-P_{1:n}\right)\Sigma_{0}\left(x,w,\cdot\right)\right\rangle $
\item $\left|\nabla_{x}P_{1:n}h\left(x,w\right)\right|^{2}\leq d\left\Vert P_{1:n}h\right\Vert^2 _{H}\left(\mbox{sup}_{i<d+1}\frac{\partial}{\partial x_{i}}\frac{\partial}{\partial r_{i}}\Sigma_{0}\left(x,w,r,w\right)\mid_{x=r}\right)$ for any $h\in H$  if $\Sigma_{0}\left(\cdot,w,\cdot,w\right)$ is in $C^{1}$.
\end{enumerate}
\end{lemma}

\proof{} 
\begin{enumerate}
\item Define $V=\left[\Sigma_{0}\left(x_{1},w_{1},\cdot\right),\ldots,\Sigma_{0}\left(x_{n},w_{n},\cdot\right)\right]\in\mathcal{H}^{n}$.
It is easy to see that:
\[
\bar{P}_{1:n}=V\left(V^{\intercal}V\right)^{-1}V^{\intercal}.
\]
Furthermore, it is well known that in that case $\bar{P}_{1:n}$ is a projection onto the space generated by 
\begin{equation*}
\{\Sigma_{0}\left(x_{1},w_{1},\cdot\right),\ldots,\Sigma_{0}\left(x_{n},w_{n},\cdot\right)\}, 
\end{equation*}
consequently we have that $\left\Vert \bar{P}_{1:n}\right\Vert \leq1$. Similarly,
we can see that $\left\Vert 1-\bar{P}_{1:n}\right\Vert \leq1$.
\item  Define the kernel $k\left(x,w,y,w'\right)=\Sigma_{0}\left(x,w,y,w'\right)+\lambda_{\left(x,w\right)}1_{\{x=y,w=w'\}}$
for $\lambda_{\left(x,w\right)}>0$, and let $\bar{H}$ be its RKHS. Define 

\begin{eqnarray*}
\bar{P}_{1:n}h & = & k\left(\cdot,(x,w)_{1:n}\right)A_{n}^{-1}\left\langle k\left(\cdot,(x,w)_{1:n}\right),h\right\rangle _{\bar{H}}\\
P_{1:n}h_{1} & = & \Sigma_{0}\left(\cdot,(x,w)_{1:n}\right)A_{n}^{-1}\left\langle \Sigma_{0}\left(\cdot,(x,w)_{1:n}\right),h_{1}\right\rangle _{H}
\end{eqnarray*}
where $h\in\bar{H},h_{1}\in H$ and $A_{n}:=\left[k\left(x_{i},w_{i},x_{j},w_{j}\right)\right]_{i,j=1:n}$. 

Observe that if $h_{1}=\Sigma_{0}\left(x,w,\cdot\right)+\lambda_{\left(x,w\right)}1_{\{(x,w)=\cdot\}}$, $h=\Sigma_{0}\left(x,w,\cdot\right)$ and $(x,w)\neq (x_{i},w_{i})$ for all $i$, we have that
\begin{eqnarray*}
\left\Vert P_{1:n}h\right\Vert _{H}^{2} & = & \left\langle P_{1:n}h,P_{1:n}h\right\rangle _{H}\\
 & = & \sum_{i,j=1}^{n}a_{i}a_{j}\Sigma_{0}\left(x_{i},w_{i},x_{j},w_{j}\right)\\
 & \leq & \sum_{i,j=1}^{n}a_{i}a_{j}\left(\Sigma_{0}\left(x_{i},w_{i},x_{j},w_{j}\right)+\lambda_{\left(x_{i},w_{i}\right)}1_{\{x_{i}=x_{j},w_{i}=w_{j}\}}\right)\\
 & = & \left\langle \bar{P}_{1:n}h_{1},\bar{P}_{1:n}h_{1}\right\rangle _{\bar{H}}\\
 & \leq & \left\Vert h_{1}\right\Vert _{\bar{H}}^{2}\mbox{, by (1)}\\
 & = & \Sigma_{0}\left(x,w,x,w\right)+\lambda_{\left(x,w\right)}
\end{eqnarray*}
where $a_{i}$ is the ith entry of $A_{n}^{-1}\left[h\left(x_{1},w_{1}\right),\ldots,h\left(x_{n},w_{n}\right)\right]^{'}$. 

Now take $(x,w)=(x_{i},w_{i})$ for some $i$, and $h=\Sigma_{0}\left(x,w,\cdot\right)$.
Consider a sequence of points $\left\{ (y_{m},w)\right\} \subset A\times W\rightarrow (x_{i},w)$
such that $y_{m}\neq x_{j}$ for all $j$. Let $a_{k}^{m}$ be the
kth entry of $A_{n}^{-1}\left[\Sigma_{0}\left(y_{m},w,x_{1},w_{1}\right),\ldots,h\left(y_{m},w,x_{n},w_{n}\right)\right]^{'}$.
Thus, we have that $a_{k}^{m}\rightarrow a_{k}$ where $a_{k}$ is
the kth entry of $A_{n}^{-1}\left[\Sigma_{0}\left(x_{i},w_{i},x_{1},w_{1}\right),\ldots,\Sigma_{0}\left(x_{i},w_{i},x_{n},w_{n}\right)\right]^{'}$.
By the previous result for all $m$,
\[
\sum_{i,j=1}^{n}a_{i}^{m}a_{j}^{m}\Sigma_{0}\left(x_{i},w_{i},x_{j},w_{j}\right)\leq\mbox{sup}_{x\in A,w\in W}\left[\Sigma_{0}\left(x,w,x,w\right)+\lambda_{\left(x,w\right)}\right]
\]
and so 
\[
\left\Vert P_{1:n}h\right\Vert _{H}^{2}=\sum_{i,j=1}^{n}a_{i}a_{j}\Sigma_{0}\left(x_{i},w_{i},x_{j},w_{j}\right)\leq\mbox{sup}_{x\in A,w\in W}\left[\Sigma_{0}\left(x,w,x,w\right)+\lambda_{\left(x,w\right)}\right].
\]

Now observe that by the previous equation and using that $A_{n}$ is positive definite, 
\begin{eqnarray*}
\left\Vert \left(1-P_{1:n}\right)h\right\Vert _{H}^{2} & = & \Sigma_{0}\left(x,w,x,w\right)-2P_{1:n}h\left(x,w\right)+\left\Vert P_{1:n}h\right\Vert _{H}^{2}\\
 & \leq & \Sigma_{0}\left(x,w,x,w\right)+\mbox{sup}_{x\in A, w\in W}\left[\Sigma_{0}\left(x,w,x,w\right)+\lambda_{\left(x,w\right)}\right]\\
 & \leq & \mbox{sup}_{x\in A, w\in W}\left[\Sigma_{0}\left(x,w,x,w\right)+\lambda_{\left(x,w\right)}\right]+\mbox{sup}_{x\in A,w\in W}\Sigma_{0}\left(x,w,x,w\right),
\end{eqnarray*}
as we wanted to prove.

\item The proof is a simple consequence of the definition of $P_{1:n}h$.
\item We have that
\begin{eqnarray*}
\left\langle \Sigma_{0}\left(x,w,\cdot\right),\left(1-P_{1:n}\right)\Sigma_{0}\left(x,w,\cdot\right)\right\rangle  & = & \Sigma_{0}\left(x,w,x,w\right)-\left\langle \Sigma_{0}\left(x,w,\cdot\right),P_{1:n}\Sigma_{0}\left(x,w,\cdot\right)\right\rangle \\
 & = & \Sigma_{0}\left(x,w,x,w\right)-P_{1:n}\Sigma_{0}\left(x,w,\cdot\right)(x,w)\\
 & = & \Sigma_{n}\left(x,w,x,w\right).
\end{eqnarray*}
\item It follows from the proof of Corollary 4.36 of \citet{steinwart2008svm}.
\end{enumerate}

\endproof

\begin{lemma}
\label{convergence_cont}

Suppose that $|W|<\infty$. Given a finite set $Q\subset A$. Consider any $s=\left(\mu,\Sigma\right)\in\mathcal{H}_{1}$, and policy $\pi$. We have that the following happens almost surely:
\begin{enumerate}
\item $\left(S_{Q}^{n}\right)$
converges to $S_{Q}^{\infty}=\left(\mu_{\infty},\Sigma_{\infty}\right) \in \mathcal{H}$, and $\mu_{\infty}\left(x,w\right)=\lim_{n}\mu_{n}\left(x,w\right)$ for all $x\in A\bigcap\mathbb{Q}^{d}\bigcup Q$, $w\in W$. Furthermore, $\Sigma_{\infty}\left(x,w,x',w'\right)=\lim_{n}\Sigma_{n}\left(x,w,x',w'\right)$ for all $x,x'\in A$, and $w,w'\in W$. 
\item The convergence of $\{\Sigma_{n}\}$ is uniform.
\item Let $w,w'\in W$, and define $g_{w,w'}\left(x,x'\right):=\Sigma_{\infty}\left(x,w,x',w'\right)$.
We have that $g_{w,w'}$ is continuous.
\item $E_{\infty}[F(x,w)]=\mu_{\infty}(x,w)$ exists for all $x\in A, w\in W$.
\item  The limit of $E_{n}\left(F^{2}\left(x,w\right)\right)$ exists for all $x\in A\bigcap\mathbb{Q}^{d}\bigcup Q$, $w\in W$, and it is equal to $\mu_{\infty}\left(\left(x,w\right)^2\right):=E_{\infty}\left(F^{2}\left(x,w\right)\right)$. Furthermore, $\Sigma_{\infty}\left(x,w,x,w\right)=\mu_{\infty}\left(\left(x,w\right)^2\right)-\mu_{\infty}^{2}\left(x,w\right)$ for all $x,w$.
% \item $F\left(x,w\right)\mid\mathcal{F}^{\infty}$ is Gaussian for
% each $x,w$.
\end{enumerate}

\end{lemma}

\proof{}
By the convergence of conditional expectations property implied by the Doob's martingale convergence theorem and the fact that $F(x,w)$ has second moments, we must have that for any points $x,w,x',w'$, it happens almost surely that $\mu_{n}(x,w)\rightarrow E\left[F\left(x,w\right)\mid\mathcal{F}_{\infty}\right]$, $\Sigma_{n}\left(x,w,x',w'\right)\rightarrow\mbox{cov}\left(F\left(x,w\right),F\left(x',w'\right)\mid\mathcal{F}_{\infty}\right)$
and $\Sigma_{n}\left(x,w,x,w\right)\rightarrow\mbox{var}\left(F\left(x,w\right)\mid\mathcal{F}_{\infty}\right)$. 

Consequently,  $\Sigma_{n}\left(x,w,x',w'\right)\rightarrow\Sigma_{\infty}\left(x,w,x',w'\right),\mu_{n}\left(x,w\right)\rightarrow\mu_{\infty}\left(x,w\right)$
for all $x,x'\in\mathbb{Q}^{d}\bigcap A\bigcup Q$, $w,w'\in W$ a.s. (the previous affirmation is true because
for any points $x,x'\in\mathbb{Q}^{d}\bigcap A\bigcup Q$, $w,w'\in W$
the event $\{\Sigma_{n}\left(x,w,x',w'\right)\rightarrow\Sigma_{\infty}\left(x,w,x',w'\right),\mu_{n}\left(x,w\right)\rightarrow\mu_{\infty}\left(x,w\right)\}$
is false has probability zero, and so the event 
\begin{equation*}
\bigcup_{x,x'\in\mathbb{Q}^{d}\bigcap A\bigcup Q;w,w'\in W}\{\Sigma_{n}\left(x,w,x',w'\right)\rightarrow\Sigma_{\infty}\left(x,w,x',w'\right),\mu_{n}\left(x,w\right)\rightarrow\mu_{\infty}\left(x,w\right)\}^{c}
\end{equation*}
has probability zero). 

Let's now prove that the
sequence $\Sigma_{n} $ is equicontinuous. Fix any $x\in A, w\in W$, observe that for any $x',y'\in A$,
by the Mean Value Theorem and \cref{rkhs}, we have that if $h=\Sigma_{0}\left(x,w,\cdot,\cdot\right),$
\begin{eqnarray*}
\left|\Sigma_{n}\left(x,w,x',w\right)-\Sigma_{n}\left(x,w,y',w\right)\right| & \leq & \mbox{sup}_{r\in A}\left|\nabla_{r}\left(1-P_{1:n}\right)h\left(r,w\right)\right|\left|x'-y'\right|\\
 & \leq & \left|x'-y'\right|d\left\Vert \left(1-P_{1:n}\right)h\right\Vert _{H}^{2}\mbox{sup}_{r\in A}\mbox{sup}_{i<n+1}\frac{\partial}{\partial x_{i}}\frac{\partial}{\partial r_{i}}\Sigma_{0}\left(x,w,r,w\right)\mid_{x=r}\\
 & = & \left|x'-y'\right|d\left(\mbox{sup}_{x\in A,w\in W}\left[\Sigma_{0}\left(x,w,x,w\right)+\lambda_{x,w}\right]+\mbox{sup}_{x\in A,w\in W}\Sigma_{0}\left(x,w,x,w\right)\right)\\
 & &\times\left(\mbox{sup}_{r\in A, w\in W}\mbox{sup}_{i<n+1}\frac{\partial}{\partial x_{i}}\frac{\partial}{\partial r_{i}}\Sigma_{0}\left(x,w,r,w\right)\mid_{x=r}\right)
\end{eqnarray*}
so $\left\{ \Sigma_{n}\left(x,w,\cdot,\cdot\right)\right\} _{n}$ is equicontinuous
for all $x,w$.

Now, let $\epsilon>0$ and $\delta<\epsilon$ such that if $\left|\left(x,w\right)-\left(y,z\right)\right|<\delta$, then $z=w$, and
$\left|a-\Sigma_{0}\left(y,z,x,z\right)\right|<\epsilon$ where $a=\Sigma_{0}\left(x,w,x,w\right)=\Sigma_{0}\left(y,z,y,z\right)$
. Observe that by \cref{rkhs}, 
\begin{eqnarray*}
\left|\Sigma_{n}\left(x,w,x,w\right)-\Sigma_{n}\left(y,w,y,w\right)\right| & \leq & \left|\Sigma_{0}\left(x,w,x,w\right)-\Sigma_{0}\left(y,w,y,w\right)\right|\\
& &+\left|\langle\Sigma_{0}\left(x,w,\cdot\right),P_{1:n}\Sigma_{0}\left(x,w,\cdot\right)\rangle-\langle\Sigma_{0}\left(y,w,\cdot\right),P_{1:n}\Sigma_{0}\left(y,w,\cdot\right)\rangle\right|\\
 & \leq & \epsilon+\left|\left\langle \Sigma_{0}\left(x,w,\cdot\right)-\Sigma_{0}\left(y,w,\cdot\right),P_{1:n}\Sigma_{0}\left(x,w,\cdot\right)\right\rangle \right|\\
 &  & +\left|\left\langle \Sigma_{0}\left(x,w,\cdot\right)-\Sigma_{0}\left(y,w,\cdot\right),P_{1:n}\Sigma_{0}\left(y,w,\cdot\right)\right\rangle \right|\\
 & \leq & \epsilon+\left(\sqrt{\left|\Sigma_{0}\left(x,w,x,w\right)+\Sigma_{0}\left(y,w,y,w\right)-2\Sigma_{0}\left(x,w,y,w\right)\right|}\right)\\
 &  & \times\left(\left|P_{1:n}\Sigma_{0}\left(x,w,\cdot\right)\right|+\left|P_{1:n}\Sigma_{0}\left(y,w,\cdot\right)\right|\right)\\
 & \leq & \epsilon+2\sqrt{2\epsilon}\sqrt{\left(\mbox{sup}_{x\in A,w\in W}\left[\Sigma_{0}\left(x,w,x,w\right)+\lambda_{\left(x,w\right)}\right]\right)}
\end{eqnarray*}
thus, $\left\{ \Sigma_{n}\left(\cdot,\cdot,\cdot,\cdot\right)\right\} $ is equicontinuous. 

Now, take any $x,x'\in A$, $w,w'\in W$. Let $\delta>0$ and $q,q'\in\mathbb{Q}^{d}\bigcap A$
such that $\left|q-x\right|<\delta,\left|q'-x'\right|<\delta$. Thus,
\begin{eqnarray*}
\left|\Sigma_{n}\left(x,w,x',w'\right)-\Sigma_{m}\left(x,w,x',w'\right)\right|&\leq&\left|\Sigma_{n}\left(x,w,x',w'\right)-\Sigma_{n}\left(q,w,q',w'\right)\right|+\left|\Sigma_{n}\left(q,w,q',w'\right)-\Sigma_{m}\left(q,w,q',w'\right)\right|\\
& & +\left|\Sigma_{m}\left(q,w,q',w'\right)-\Sigma_{m}\left(x,w,x',w'\right)\right|,
\end{eqnarray*}
which implies that $\left\{ \Sigma_{n}\left(x,w,x',w'\right)\right\} $ is
a Cauchy sequence if $\left\{ \Sigma_{n}\left(q,w,q',w'\right)\right\} $
converges, and thus $\left\{ \Sigma_{n}\left(x,w,x',w'\right)\right\} $ converges
too. Consequently, $\Sigma_{n}\left(x,w,x',w'\right)\rightarrow\Sigma_{\infty}\left(x,w,x',w'\right)$ for
all $x,x'\in A$, $w,w'\in W$. Furthermore, the convergent is uniform because $\left\{ \Sigma_{n}\right\} $ is
equicontinuous,which implies
that $\Sigma_{\infty}$ is continuous. Observe, that since $\Sigma_{\infty}(x,w,x,w)$ exists for all $x,w$, we must have that $\mu_{\infty}(x,w)$ exists for all $x,w$. 
Observe that the limit of kernels is a kernel, and so $\Sigma_{\infty}$
is a kernel. This proves (1), (2), (3) and (4). Similarly, we can prove (5). 

\endproof

\begin{lemma}
\label{uniformly_kg}

Suppose that $|W|<\infty$. Given $s_{0}\in\mathcal{H}_{1}$, $Q$ a finite set of $A\bigcap \mathbb{Q}^{d}$,
then $R_{Q}^{N-1}\left(S^{n};x,w\right)$ converges uniformly
to $R_{Q}^{N-1}\left(S^{\infty};x,w\right)$ a.s.
\end{lemma}

\proof{}

By \cref{convergence_cont}, we have that $\Sigma_{n}\left(x,w,r,w'\right)$
converges uniformly to $\Sigma_{\infty}\left(x,w,r,w'\right)$
for all $x,r,w,w'$ a.s.
Since we assume that the noise is bounded, we have that 
\[
\Sigma_{0}\left(x,w,x,w\right)+K_{\lambda}\geq\Sigma_{n}\left(x,w,x,w\right)+\lambda_{\left(x,w\right)}\geq k_{\lambda}
\]
where $K_{\lambda},k_{\lambda} > 0$ are the bounds of the variance of the noise. Thus, for all $x,w$
\[
\frac{1}{\Sigma_{n}\left(x,w,x,w\right)+\lambda_{\left(x,w\right)}}\Rightarrow\frac{1}{\Sigma_{\infty}\left(x,w,x,w\right)+\lambda_{\left(x,w\right)}}\mbox{ a.s.}
\]

Now,
\begin{eqnarray*}
\left|R_{Q}^{N-1}\left(S^{n};x,w\right)-R_{Q}^{N-1}\left(S^{\infty};x,w\right)\right| & = & \left|E\left[\mbox{max}_{z\in Q}\left[a_{n}\left(z\right)+\sigmatilde_{n}\left(z,x,w\right)Z\right]\right]-E\left[\mbox{max}_{z\in Q}\left[a_{\infty}\left(z\right)+\sigmatilde_{\infty}\left(z,x,w\right)Z_{1}\right]\right]\right|\\
 & \leq & E\left[\mbox{max}_{z\in Q}\left|\left[a_{n}\left(z\right)+\sigmatilde_{n}\left(z,x,w\right)Z-a_{\infty}\left(z\right)-\sigmatilde_{\infty}\left(z,x,w\right)Z\right]\right|\right]\\
\end{eqnarray*}
thus it is clear that $R_{Q}^{N-1}\left(S^{n};x,w\right)\Rightarrow R_{Q}^{N-1}\left(S^{\infty};x,w\right)$ for all $x,w$ a.s.,
because $a_{n}\left(z\right)\rightarrow a_{\infty}\left(a\right)$,
$\sigmatilde_{n}\left(z,x,w\right)\Rightarrow\sigmatilde_{\infty}\left(z,x,w\right)$ a.s., and $Q$ is finite. 

\endproof

\begin{lemma}
\label{equalitylim_cont}

Suppose that $|W|<\infty$. Under the BQO policy, for any finite set $Q$ of $A\bigcap \mathbb{Q}^{d}$, if $s=\left(\mu_{0},\Sigma_{0}\right)\in\mathcal{H}_{1}$,
then $V_{Q}^{N}\left(S_{Q}^{\infty}\right)=U_{Q}\left(S_{Q}^{\infty}\right)$
a.s. where $S_{Q}^{\infty}$ is the limit of $\left(S_{Q}^{n}\right)$,
and 
\begin{eqnarray*}
V_{Q}^{n}\left(s\right) & = & \sup_{\pi\in\Pi}E^{\pi}\left[\max_{x\in Q}\mu_{N}\left(x\right)\mid S^{n}=s\right]\\
U_{Q}\left(s\right) & = & E\left[\max_{x\in Q}G\left(x\right)\mid S^{0}=s\right]
\end{eqnarray*}
\end{lemma}

\proof{}
By \cref{uniformly_kg} and \cref{measureio_cont}, there exists a measurable set $F$ of probability one, such that if $\omega\in F$, $R_{Q}^{N-1}\left(S^{n};x,w\right)$ converges uniformly to $R_{Q}^{N-1}\left(S^{\infty};x,w\right)$, and the statement of the \cref{measureio_cont} is true. Fix a $\omega\in F$. 

Let $\epsilon>0$. By \cref{uniformly_kg}, $R_{Q}^{N-1}\left(S^{\infty};x,w\right)$ is continuous, and so it is uniformly continuous because $A$ is compact. Consequently, there exists $\delta>0$ such that
if $\left|(x,w)-(y,w')\right|<\delta$, then $w=w'$ and
\begin{equation}
\left|R_{Q}^{N-1}\left(S^{\infty};x,w\right)-R_{Q}^{N-1}\left(S^{\infty};y,w'\right)\right|<\epsilon.
\label{eq:r_unif}
\end{equation}

By \cref{measureio_cont}, the same $\delta>0$ can be chosen such that that if the policy chooses an infinite number of points in $B_{\delta}(x)\times \{w\}$, thus $\left|R_{Q}^{N-1}\left(S^{\infty};x,w\right)-V_{Q}^{N}\left(S^{\infty}\right)\right|<\epsilon$.

In order to simplify the notation, we assume that $A=[0,1]$. We build
a collection of finite sets of number of rationals $Q_{n}$ such that
$\bigcup_{n}Q_{n}=A\bigcap\mathbb{Q}$ (for example, see proof of \cref{continuum_th}).
We order the elements of $Q_{n}=\left\{ q_{m}\right\} _{n\geq m\geq1}$
such that $q_{m}\leq q_{m+1}$. We define the intervals
$I_{m}=[q_{m},q_{m+1})$ if $m < n-1$, and $I_{n-1}=[q_{n-1},q_{n}]$. By \cref{propsition_approx_rationals}, we
know that there exists $Q_{n}$ such that $\mbox{sup}_{m}\left|q_{m}-q_{m+1}\right|<\delta$.

Take any finite set $Q$ of $A\bigcap \mathbb{Q}^{d}$. Let $I=\{1,\ldots,n-1\}$. Let $B':=\bigcup_{w\in W}B_{w}\times\left\{ w\right\}$ where $B_{w}\subset I$ for all $w$,  and $B_{w}$ satisfies that there exists $x_{w}\in I_{i}$ for all $i\in B_{w}$ such that 
$R_{Q}^{N-1}\left(S_{Q}^{\infty};x_{w},w\right)>\epsilon+V_{Q}^{N}\left(S_{Q}^{\infty}\right)$. We then have that $V_{Q}^{N}\left(S_{Q}^{\infty}\right)\leq R_{Q}^{N-1}\left(S^{\infty};x,w\right)\leq\epsilon+V_{Q}^{N}\left(S_{Q}^{\infty}\right)$
for all $(x,w)$ such that $x\in I_{i}$ for all $i\notin B_{w}$ and $w\in W$. Observe that by \cref{moreinf}, $R_{Q}^{N-1}\left(s;x,w\right)\geq V_{Q}^{N}\left(s\right)$ for all $x,w$.

Assume that  $B'\neq\emptyset$. By \cref{measureio_cont}, there exists $N$ such that if $n>N$, $(x,w)$ is not chosen if $x\in I_{i}$ where $i\in B_{w}$ and $w\in W$.
Take any $(x_{w},w)$, such that $x_{w}\in I_{i}$ for some $i\in B_{w}$, $w\in W$, and define $\epsilon':=R_{Q}^{N-1}\left(S_{Q}^{\infty};x_{w},w\right)-V_{Q}^{N}\left(S_{Q}^{\infty}\right)-\epsilon > 0$. Take any $(y,w')$ such that $y\in I_{i}$ where $i\notin B_{w'}$ and $w'\in W$. Observe that $\epsilon'\leq R_{Q}^{N-1}\left(S_{Q}^{\infty};x_{w},w\right)-R_{Q}^{N-1}\left(S_{Q}^{\infty};y,w'\right)$. Define $\epsilon_{1}:=\epsilon'/2$. Since $R_{Q}^{N-1}\left(S_{Q}^{n};x,w\right)$
converges uniformly to $R_{Q}^{N-1}\left(S_{Q}^{\infty};x,w\right)$,
we have that there exists $n>N$ such that

\begin{eqnarray*}
R_{Q}^{N-1}\left(S^{n};x_{w},w\right) & > & R_{Q}^{N-1}\left(S_{Q}^{\infty};x_{w},w\right)-\epsilon_{1}\\
 & \geq & R_{Q}^{N-1}\left(S_{Q}^{\infty};y,w'\right)+\epsilon_{1}\\
 & > & R_{Q}^{N-1}\left(S^{n};y,w'\right)
\end{eqnarray*}
which contradicts that we would not choose a point $(x,w)$ if $x\in I_{i}$, $w\in W$, $i\in B_{w}$ and $n>N$. Consequently, $B'$ is the empty set. 

Thus, we have that for all $(x,w)$ and $\epsilon>0$, $R_{Q}^{N-1}\left(S_{Q}^{\infty};x,w\right)\leq \epsilon+V_{Q}^{N}\left(S_{Q}^{\infty}\right)$, and so $R_{Q}^{N-1}\left(S_{Q}^{\infty};x,w\right)= V_{Q}^{N}\left(S_{Q}^{\infty}\right)$. Finally, \cref{important} implies the result.

\endproof

 \begin{lemma}
\label{propsition_approx_rationals}
Let $Q_{m}=\bigcup_{n\geq1}^{m}\bigcup_{l=0}^{n}\left\{ \frac{l}{n}\right\} $
for $m\geq1$. If $\epsilon>0$, we have that there exists $M$ such
that if $m\geq M$, and the elements of $Q_{m}$ are sorted such that $q_{k}\leq q_{k+1}$,
then $\left|q_{k+1}-q_{k}\right|<\epsilon$ for all $k$.
 \end{lemma}
 
 \proof{}
 The proof is trivial. 
\endproof
\begin{lemma}
\label{measureio_cont}

Suppose that $|W|<\infty$. Let $s=\left(\mu,\Sigma\right)\in\mathcal{H}_{1}$. It almost surely happens that: for every $\epsilon>0$ and $w\in W$, there exists $\delta_{0}>0$ such
that if the policy $\pi$ measures an infinite number of alternatives 
in $B_{\delta}\left(x\right)\times \left\{ w\right\}$ for $\delta < \delta_{0}$, where $B_{\delta}\left(x\right)$
is the open ball of radio $\delta$ with center $x$, and $x\in A$, we then have that 
\begin{equation}
\left|R_{Q}^{N-1}\left(S^{\infty};x,w\right)-V_{Q}^{N}\left(S^{\infty}\right)\right|<\epsilon.\label{eq:-1}
\end{equation}

\end{lemma}

\proof{}

We can assume  without loss of generality that the Gaussian process is separable, because the probability space is complete and so there exists a separable version of every stochastic process defined in $A$ (\citealt{neveuBook, pollardMini}). Let $x\in A$ and $w\in W$. We define $f(y):=F(y,w)$. Since $\Sigma_{0}(\cdot,w,\cdot,w)$ is Lipschitz continuous, thus $f$ is continuous a.s., and so
by \citet{adlerRandomFields}, we have that for all $a>0$
\[
\int_{0}^{a}H^{1/2}\left(u\right)du<\infty
\]
where $H$ is the log-entropy function for $A$. 
By \cref{convergence_cont} and the fact that $f$ is continuous a.s., there exists a measurable set $F$ of probability one, such that if $\omega\in F$, the limits of the parameters of the GP exist and $f$ is continuous. Fix a $\omega\in F$. Let $\epsilon>0$. We then have that
\[
\mbox{lim}_{a\rightarrow0}\int_{0}^{a}H^{1/2}\left(u\right)du=0,
\]
thus there exists $\delta'>0$ such that $\epsilon>\delta'$, and if $a\leq\delta'$, then
$0\leq\int_{0}^{a}H^{1/2}\left(u\right)du<\epsilon$.

A simple extension of Lemma 1.3.1 of Chapter 1 of \citet{adlerRandomFields} shows that the canonical metric $d\left(s,t\right)=\left\{ E\left(\left(f\left(s\right)-f\left(t\right)\right)^{2}\right)\right\} ^{1/2}$
satisfies that there exists $\delta$ such that $\left|d\left(x,y\right)\right|<\delta'$
if $|y-x| < \delta$. 
% Since $\mu_{\infty}$ is continuous,
% we can take $\delta$ such that $\left|\mu_{\infty}\left(x\right)-\mu_{\infty}\left(y\right)\right|<\epsilon$
% if $y\in B_{\delta}\left(x\right)$.
Define $\epsilon\left(\delta\right):=\mbox{sup}_{|y-x|<\delta}\left|f\left(x\right)-f\left(y\right)\right|$,
which is finite because $f$ is bounded ($f$ is continuous in a compact set).

Observe that for all $n$ 
\begin{eqnarray*}
d^{2}\left(x,y\mid\mathcal{F}_{n}\right) & = & E\left[\left(f\left(x\right)-\mu_{n}\left(x,w\right)-f\left(y\right)+\mu_{n}\left(y,w\right)\right)^{2}\mid\mathcal{F}_{n}\right]\\
 & = & \Sigma_{n}\left(x,w,x,w\right)+\Sigma_{n}\left(y,w,y,w\right)-2\Sigma_{n}\left(x,w,y,w\right)\\
 & = & \Sigma_{n-1}\left(x,w,x,w\right)+\Sigma_{n-1}\left(y,w,y,w\right)-2\Sigma_{n-1}\left(x,w,y,w\right)\\
 & & - \frac{\Sigma_{n-1}^{2}\left(x,w,x_{n},w_{n}\right)+\Sigma_{n-1}^{2}\left(y,w,x_{n},w_{n}\right)+2\Sigma_{n-1}\left(x,w,x_{n},w_{n}\right)\Sigma_{n-1}\left(y,w,x_{n},w_{n}\right)}{\Sigma_{n-1}\left(x_{n},w_{n},x_{n},w_{n}\right)+\lambda_{x_{n},w_{n}}}\\
 & = & \Sigma_{n-1}\left(x,w,x,w\right)+\Sigma_{n-1}\left(y,w,y,w\right)-2\Sigma_{n-1}\left(x,w,y,w\right)\\
 & &-\frac{\left(\Sigma_{n-1}\left(x,w,x_{n},w_{n}\right)-\Sigma_{n-1}\left(y,w,x_{n},w_{n}\right)\right)^{2}}{\Sigma_{n-1}\left(x_{n},w_{n},x_{n},w_{n}\right)+\lambda_{x_{n},w_{n}}}\\
 & \leq & d^{2}\left(x,y\mid\mathcal{F}_{n-1}\right)\\
 & \vdots\\
 & \leq & d^{2}\left(x,y\right).
\end{eqnarray*}

Now, we will show that for all $n$, $H_{n}\left(u\right)\leq H\left(u\right)$
where $H_{n}$ is the log-entropy function for $A$ associated to the Gaussian
process on $f$ given $\mathcal{F}_{n}$, and $u>0$. Let $\left\{ B_{j}\left(x_{j};u\right)\right\} _{j=1}^{N\left(u\right)}$
be a cover of $A$, where $B_{j}\left(x_{j};u\right)$ is the ball
with center $x_{j}$ and radius $u$ using the metric $d$. Define
the sequence of balls $\left\{ B_{j}^{n}\left(x_{j};u\right)\right\} _{j=1}^{N\left(u\right)}$
, where $B_{j}^{n}\left(x_{j};u\right)$ is the ball with center $x_{j}$
and radius $u$ using the metric $d\left(\cdot,\cdot\mid\mathcal{F}_{n}\right)$.
Since $d^{2}\left(x,y\mid\mathcal{F}_{n}\right)\leq d^{2}\left(x,y\right)$,
we have that $\left\{ B_{j}^{n}\left(x_{j};u\right)\right\} _{j=1}^{N\left(u\right)}$
is a cover of $A$, and thus $H_{n}\left(u\right)\leq H\left(u\right)$.
So, for all $n$, we have that $\int_{0}^{a}H_{n}^{1/2}\left(u\right)du<\epsilon$ if $a\leq\delta'$.

Define $\epsilon_{n}\left(\delta\right)=\mbox{sup}_{|y-x|<\delta}\left|f\left(x\right)-\mu_{n}\left(x,w\right)-f\left(y\right)+\mu_{n}\left(y,w\right)\right|$.
We will prove by induction on $n$ that for all $x$, $\Sigma_{n}\left(x,w,x,w\right)>0$.
The result is true when $n=0$. Suppose that it is true if $m<n$.
Thus,
\begin{eqnarray*}
\Sigma_{n}\left(x,w,x,w\right) & = & \Sigma_{n-1}\left(x,w,x,w\right)-\frac{\Sigma_{n-1}^{2}\left(x,w,x_{n},w_{n}\right)}{\Sigma_{n-1}\left(x_{n},w_{n},x_{n},w_{n}\right)+\lambda_{x_{n},w_{n}}},
\end{eqnarray*}
and $\Sigma_{n-1}\left(x,w,x,w\right)\Sigma_{n-1}\left(x_{n},w_{n},x_{n},w_{n}\right)\geq\Sigma_{n-1}^{2}\left(x,w,x_{n},w_{n}\right)$,
consequently,
\[
\Sigma_{n-1}\left(x,w,x,w\right)\left(\Sigma_{n-1}\left(x_{n},w_{n},x_{n},w_{n}\right)+\lambda_{x_{n},w_{n}}\right)\geq\Sigma_{n-1}^{2}\left(x,w,x_{n},w_{n}\right)+\Sigma_{n-1}\left(x,w,x,w\right)\lambda_{x_{n},w_{n}}>\Sigma_{n-1}^{2}\left(x,w,x_{n},w_{n}\right),
\]
and then $\Sigma_{n}\left(x,w,x,w\right)>0$. By induction we have that
$\Sigma_{n}\left(x,w,x,w\right)>0$ for all $n$. Fix $n$, we then have that the GP
on $f$ given $\mathcal{F}_{n}$ is nondegenerate, and so by Theorem 1.5.4 and Corollary 1.3.4 of \citet{adlerRandomFields}, we have that there exists $K$ such that
\[
E[\epsilon_{n}\left(\delta\right)\mid\mathcal{F}_{n}]\leq K\int_{0}^{\delta'}H_{n}^{1/2}\left(\epsilon\right)d\epsilon<K\epsilon.
\]

Furthermore, in the proof of Theorem 1.3.3 of \citet{adlerRandomFields}, it is shown that for all
$u>1$, 
\[
P\left[\mbox{sup}_{d_{n}\left(y,x\right)\leq\delta'}\left(f\left(y\right)-\mu_{n}(y,w)+\mu_{n}(x,w)-f\left(x\right)\right)\geq uS\mid\mathcal{F}_{n}\right]\leq2^{1-u^{2}}
\]
where $S$ is a constant such that $0\leq S\leq\int_{0}^{\delta'}H_{n}^{1/2}\left(\epsilon\right)d\epsilon$, and $d_{n}\left(\cdot,\cdot\right):=d\left(\cdot,\cdot\mid\mathcal{F}_{n}\right)$.
Consequently,

\begin{eqnarray*}
E\left[\left(\epsilon_{n}\left(\delta\right)\right)^{2}\mid\mathcal{F}_{n}\right] & = & \int_{0}^{\infty}P\left[\left[\mbox{sup}_{d_{n}\left(y,x\right)\leq\delta'}\left(f\left(y\right)-\mu_{n}\left(y,w\right)-f\left(x\right)+\mu_{n}\left(x,w\right)\right)\right]^{2}\geq u\mid\mathcal{F}_{n}\right]du\\
 & = & \int_{0}^{\infty}P\left[\left[\mbox{sup}_{d_{n}\left(y,x\right)\leq\delta'}\left(f\left(y\right)-\mu_{n}\left(y,w\right)-f\left(x\right)+\mu_{n}\left(x,w\right)\right)\right]\geq\sqrt{u}\mid\mathcal{F}_{n}\right]du\\
 & = & 2S^{2}\int_{0}^{\infty}P\left[\left[\mbox{sup}_{d_{n}\left(y,x\right)\leq\delta'}\left(f\left(y\right)-\mu_{n}\left(y,w\right)-f\left(x\right)+\mu_{n}\left(x,w\right)\right)\right]\geq tS\mid\mathcal{F}_{n}\right]tdt\\
 & \leq & 2S^{2}+2S^{2}\int_{1}^{\infty}2^{1-u^{2}}udu\\
 & = & S^{2}\left[2+2\int_{1}^{\infty}2^{1-u^{2}}udu\right]\\
 & \leq & \left(\int_{0}^{\delta'}H_{n}^{1/2}\left(\epsilon\right)d\epsilon\right)^{2}\left[2+2\int_{1}^{\infty}2^{1-u^{2}}udu\right]\\
 & \leq & \epsilon^{2}K'
\end{eqnarray*}
where $K':=2+2\int_{1}^{\infty}2^{1-u^{2}}udu\geq0$. In particular, $\epsilon^{2}\left(\delta\right)$ is in $L^{1}$,
and so we must have that $E[\epsilon^{2}\left(\delta\right)\mid\mathcal{F}_{n}]\rightarrow E[\epsilon^{2}\left(\delta\right)\mid\mathcal{F}_{\infty}]$, and $E\left[\mbox{sup}_{|y-x|<\delta}\left|f\left(y\right)-f\left(x\right)\right|\mid\mathcal{F}_{n}\right]\rightarrow E\left[\mbox{sup}_{|y-x|<\delta}\left|f\left(y\right)-f\left(x\right)\right|\mid\mathcal{F}_{\infty}\right]$ a.s.

Note that, 
\[
\epsilon\left(\delta\right)\leq\epsilon'\left(\delta\right)+\mbox{sup}_{|y-x|<\delta}\left|\mu_{n}\left(x,w\right)-\mu_{n}\left(y,w\right)\right|
\]
where $\epsilon'\left(\delta\right)=\mbox{sup}_{d\left(x,y\right)<\delta'}\left|f\left(x\right)-\mu_{n}\left(x,w\right)-f\left(y\right)+\mu_{n}\left(y,w\right)\right|.$ Consequently,
\begin{eqnarray*}
E\left[\epsilon^{2}\left(\delta\right)\mid\mathcal{F}_{n}\right] & \leq & E\left[\epsilon_{n}^{2}\left(\delta\right)\mid\mathcal{F}_{n}\right]+\left(\mbox{sup}_{|y-x|<\delta}\left(\mu_{n}\left(y,w\right)-\mu_{n}\left(x,w\right)\right)\right)^{2}\\
& &+2E\left[\epsilon_{n}\left(\delta\right)\mid\mathcal{F}_{n}\right]\left(\mbox{sup}_{|y-x|<\delta}\left(\mu_{n}\left(y,w\right)-\mu_{n}\left(x,w\right)\right)\right)\\
 & \leq & \epsilon^{2}K'+\left(E\left[\mbox{\ensuremath{\mbox{sup}_{|y-x|<\delta}}}\left(f\left(y\right)-f\left(x\right)\right)\mid\mathcal{F}_{n}\right]\right)^{2}+2K\epsilon E\left[\mbox{\ensuremath{\mbox{sup}_{|y-x|<\delta}}}\left(f\left(y\right)-f\left(x\right)\right)\mid\mathcal{F}_{n}\right]
\end{eqnarray*}
and so 
\[
E\left[\epsilon^{2}\left(\delta\right)\mid\mathcal{F}_{\infty}\right]<\epsilon^{2}K'+\left(E\left[\mbox{\ensuremath{\mbox{sup}_{|y-x|<\delta}}}\left(f\left(y\right)-f\left(x\right)\right)\mid\mathcal{F}_{\infty}\right]\right)^{2}+2K\epsilon E\left[\mbox{\ensuremath{\mbox{sup}_{|y-x|<\delta}}}\left(f\left(y\right)-f\left(x\right)\right)\mid\mathcal{F}_{\infty}\right]
\] a.s. Similarly,
\[
E\left[\epsilon\left(\delta\right)\mid\mathcal{F}_{\infty}\right]<K\epsilon+E\left[\mbox{sup}_{|y-x|<\delta}\left(f\left(y\right)-f\left(x\right)\right)\mid\mathcal{F}_{\infty}\right]\mbox{ a.s.}
\]

We will prove that $\mbox{sup}_{|y-x|<\delta}\left|\mu_{\infty}\left(x,w\right)-\mu_{\infty}\left(y,w\right)\right|<\infty$
a.s. Suppose that $\mbox{sup}_{|y-x|<\delta}\left|\mu_{\infty}\left(x,w\right)-\mu_{\infty}\left(y,w\right)\right|=\infty$
in a set $B\subset\Omega$ such that $P\left[B\right]>0$. We then
have that 
\begin{eqnarray*}
\infty & = & E\left[\mbox{sup}_{|y-x|<\delta}\left|\mu_{\infty}\left(x,w\right)-\mu_{\infty}\left(y,w\right)\right|\right]\\
 & = & E\left[\mbox{sup}_{|y-x|<\delta}E\left[f\left(y\right)-f\left(x\right)\mid\mathcal{F}_{\infty}\right]\right]\\
 & \leq & E\left[E\left[\mbox{sup}_{|y-x|<\delta}f\left(y\right)-f\left(x\right)\mid\mathcal{F}_{\infty}\right]\right]\\
 & = & E\left[\mbox{sup}_{|y-x|<\delta}f\left(y\right)-f\left(x\right)\right].
\end{eqnarray*}
However, $E\left[\mbox{sup}_{|y-x|<\delta}f\left(y\right)-f\left(x\right)\right]<\infty$,
which is a contradiction. Consequently, the event 
\begin{equation*}
\mbox{sup}_{|y-x|<\delta}\left|\mu_{\infty}\left(x,w\right)-\mu_{\infty}\left(y,w\right)\right|<\infty
\end{equation*}
occurs a.s. Similarly, $E\left[\mbox{\ensuremath{\mbox{sup}_{|y-x|<\delta}}}\left(f\left(y\right)-f\left(x\right)\right)\mid\mathcal{F}_{\infty}\right]<\infty$ a.s. 

% Furthermore, we have that $E[\mbox{sup}_{x\in A}F\left(x,w\right)]<\infty$ using
% Theorem 1.3.3 and Theorem 1.5.4 of \citet{adlerRandomFields}. 
By using a similar argument than the previous one and the fact that $F$ is bounded a.s., we can then see that $\mbox{sup}_{x\in A}\left|\mu_{\infty}\left(x,w\right)\right|<\infty$ a.s.

Now observe that if $E\left[\mbox{\ensuremath{\mbox{sup}_{|y-x|<\delta}}}\left(f\left(y\right)-f\left(x\right)\right)\mid\mathcal{F}_{\infty}\right]<\infty$,
we then have that 
\begin{eqnarray*}
0 & \leq & E\left[\mbox{lim}_{\delta\rightarrow0^{+}}\mbox{sup}_{|y-x|<\delta}E\left[\left(f\left(y\right)-f\left(x\right)\right)\mid\mathcal{F}_{\infty}\right]\right]\\
 & = & \mbox{lim}_{\delta\rightarrow0^{+}}E\left[\mbox{sup}_{|y-x|<\delta}E\left[\left(f\left(y\right)-f\left(x\right)\right)\mid\mathcal{F}_{\infty}\right]\right]\mbox{ by the dominated convergence theorem,}\\
 & \leq & \mbox{lim}_{\delta\rightarrow0^{+}}E\left[\mbox{sup}_{|y-x|<\delta}\left(f\left(y\right)-f\left(x\right)\right)\right]\\
 & \leq & \mbox{lim}_{\delta'\rightarrow0^{+}}K\int_{0}^{\delta'}H^{1/2}\left(r\right)dr=0
\end{eqnarray*}
and thus $\mbox{lim}_{\delta\rightarrow0^{+}}E\left[\mbox{sup}_{|y-x|<\delta}\left(f\left(y\right)-f\left(x\right)\right)\mid\mathcal{F}_{\infty}\right]=0$
a.s. 

Observe that there exists a sequence i.i.d. standard normal random variables $\left\{ \epsilon_{k}\right\} _{k}$ such that we observe $F(x_{i},w_{i})+\sqrt{\lambda_{(x_{i},w_{i})}}\epsilon_{i}$ the $i$-th time that $F$ is queried. By the strong law of large numbers, there exists a measurable set $F_{1}$ of probability one, such that $\frac{1}{n}\sum_{j=1}^{n}\epsilon_{j}$ converges to the zero random variable.
% Take a numerable dense set $B=\left\{ b_{i}\right\} _{i}$ of $A$ ($B$ can be built using the rationals). Observe that for each $x\in B$, there exists a sequence i.i.d. normal random variables $\left\{ \epsilon_{k,x}\right\} _{k}$ with mean zero and variance $\lambda_{(x,w)}$ such that we observe $F(x,w)+\epsilon_{k,x}$ the $k$-th time that $F(x,w)$ is queried. We can consider the sequence of independent normal random variables $\left\{ \hat{\epsilon}_{j}\right\} _{j}:=\left\{\epsilon_{k,b_{i}}\right\}_{k,i}$. By the Liapunov condition, there exists a measurable set $F_{1}$ of probability one, such that $\frac{1}{n}\sum_{j=1}^{n}\hat{\epsilon}_{j}$ converges to the zero
% random variable.

Fix a $\omega\in F\bigcap F_{1}$, and take $\delta_{0}<\delta$  such that 
\begin{equation*}
E\left[\mbox{\ensuremath{\mbox{sup}_{|y-x|<\delta}}}\left(f\left(y\right)-f\left(x\right)\right)\mid\mathcal{F}_{\infty}\right]<\epsilon/\mbox{max}\left(\mbox{sup}_{x\in A}\left|\mu_{\infty}\left(x,w\right)\right|+L,1\right), 
\end{equation*}
and $\int_{0}^{\delta'}H^{1/2}\left(r\right)dr<\epsilon/\mbox{max}\left(\mbox{sup}_{x\in A}\left|\mu_{\infty}\left(x,w\right)\right|+L,1\right)$
where $\delta'=\mbox{sup}_{|y-x|<\delta_{0}}d\left(x,y\right)$ and $L$ is an upper bound of $|f|$ (e.g. $L\geq\left|f\left(x\right)\right|$ for all $x\in A$).

% We have that for each chosen point $(x_{i},w_{i})$, there exists a sequence of independent normal random variables with mean zero and variance $\lambda_{(x_{i},w_{i})}$  such that 

% Observe that for each $(x,w)$ there exists a sequence i.i.d. normal random variables $\left\{ \epsilon_{k}\right\} _{k}$ with mean zero and variance $\lambda_{(x,w)}$ such that we observe $F(x,w)+\epsilon_{k}$ the $k$-th time that $F(x,w)$ is queried. By the strong law of large numbers, we have that there exists a measurable set $F_{1}$ of probability one, such that $\frac{1}{m}\sum_{k=1}^{m}\epsilon_{k}$ converges to the zero random variable as $m$ goes to infinity.

% Let $\omega\in \Omega$ such that there exists $n$ such that $E[\epsilon^{2}\left(\delta\right)\mid\mathcal{F}_{\infty}]<\epsilon+E[\epsilon^{2}\left(\delta\right)\mid\mathcal{F}_{n}]$. 
% % Let $\delta_{0} < \delta$ such that $\mbox{sup}_{y\in B_{\delta_{0}}\left(x\right)}\left|\mu_{n}\left(x\right)-\mu_{n}\left(y\right)\right| < \epsilon$.

Suppose that the policy $\pi$ measures an infinite number of alternatives
in $B_{\delta_{0}}\left(x\right) \times {w}$. Consider the collection $C=\left\{ y_{n_{i}}\right\} _{i\geq1}$
where $y_{n_{i}}=f\left(x_{n_{i}}\right)+\lambda_{(x_{n_{i}},w)}\epsilon_{n_{i}}$ and $x_{n_{i}}\in B_{\delta_{0}}\left(x\right)$. We know that $\frac{1}{m}\sum_{j=1}^{m}\epsilon_{n_{j}}$ converges to zero. Furthermore, $\lambda_{(x,w)}$ is bounded for all $x,w$, and so $\frac{1}{m}\sum_{j=1}^{m}\sqrt{\lambda_{(x_{n_{i}},w)}}\epsilon_{n_{j}}$ is bounded for $m$ large, and converges to zero. Thus for $m$ large 
\begin{eqnarray*}
\left|\frac{1}{m}\sum_{i=1}^{m}y_{n_{i}}\right| & \leq & \frac{1}{m}\sum_{i=1}^{m}\left|f\left(x_{n_{i}}\right)\right|+\frac{1}{m}\sum_{i=1}^{m}\left|\sqrt{\lambda_{(x_{n_{i}},w)}}\epsilon_{i}\right|\\
 & \leq & L+\frac{1}{m}\sum_{i=1}^{m}\left|\sqrt{\lambda_{(x_{n_{i}},w)}}\epsilon_{i}\right|
\end{eqnarray*}
where $L\geq\left|f\left(x_{n_{i}}\right)\right|$ ($f$ is bounded
because $\omega\in F\bigcap F_{1}$), and thus $\frac{1}{m}\sum_{i=1}^{m}y_{n_{i}}$ is bounded for
$m$ large, and then there exists a convergent subsequence of $\left\{ \frac{1}{m}\sum_{i=1}^{m}y_{n_{i}}\right\} $.
Denote this convergent subsequence by $z_{m}$, and its limit
by $Z_{\delta_{0}}$. Observe that,
\[
z_{m}-\epsilon\left(\delta_{0}\right)\leq f\left(x\right)+\frac{1}{m}\sum_{i=1}^{m}\sqrt{\lambda_{(x_{n_{i}},w)}}\epsilon_{i}\leq z_{m}+\epsilon\left(\delta_{0}\right)
\]
and thus
\[
Z_{\delta_{0}}-\epsilon\left(\delta_{0}\right)\leq f\left(x\right)\leq Z_{\delta_{0}}+\epsilon\left(\delta_{0}\right).
\]
Now, by \cref{convergence} (remember that  $\omega\in F\bigcap F_{1}$), we have that
\begin{eqnarray*}
\Sigma_{\infty}\left(x,w,x,w\right) & = & E\left[\left(f\left(x\right)-\mu_{\infty}\left(x,w\right)\right)^{2}\mid\mathcal{F}_{\infty}\right]\\
 & = & E\left[f\left(x\right)^{2}\mid\mathcal{F}_{\infty}\right]-\left(\mu_{\infty}\left(x,w\right)\right)^{2}\\
 & \leq & E\left[\epsilon^{2}\left(\delta_{0}\right)+\left|Z_{\delta_{0}}\right|^{2}+2\epsilon\left(\delta_{0}\right)\left|Z_{\delta_{0}}\right|\mid\mathcal{F}_{\infty}\right]-\mu_{\infty}^{2}\left(x,w\right)\\
 & \leq & E\left[\epsilon^{2}\left(\delta_{0}\right)\mid\mathcal{F}_{\infty}\right]+2E\left[\epsilon\left(\delta_{0}\right)\mid\mathcal{F}_{\infty}\right]L+E\left[\epsilon\left(\delta_{0}\right)\mid\mathcal{F}_{\infty}\right](L+\left|\mu_{\infty}\left(x,w\right)\right|)\\
%  +2\left|\mu_{\infty}\left(x,w\right)\right|E\left[\epsilon\left(\delta_{0}\right)\mid\mathcal{F}_{\infty}\right]\\
%  & &+\left(E\left[\epsilon\left(\delta_{0}\right)\mid\mathcal{F}_{\infty}\right]\right)^{2}\\
& < & \hat \epsilon:=\epsilon^{2}K'+\epsilon^{2}+2K\epsilon^{2}+2\left(K\epsilon+\epsilon\right)+\left(K\epsilon+\epsilon\right)
\end{eqnarray*}

Since $Q$ and $W$ are finite, $\left|\sum_{w'}p(w')\left[\Sigma_{\infty}\left(z,w',x,w\right)\right]\right|<\hat \epsilon_{2}:=\left|W\right|\sup_{w}p\left(w\right)\sqrt{\hat{\epsilon}}\mbox{sup}_{x,w}\sqrt{\Sigma_{0}\left(x,w,x,w\right)}$
for all $z\in Q$.

Now observe that,
\begin{eqnarray*}
\left|R_{Q}^{N-1}\left(S^{\infty};x,w\right)-V_{Q}^{N}\left(S^{\infty}\right)\right| & = & \left|E\left[\mbox{max}_{z\in Q}\left[a_{\infty+1}\left(z\right)\right]\mid\mathcal{F}_{\infty}\right]-V_{Q}^{N}\left(S^{\infty}\right)\right|\\
 & \leq & E\left[\mbox{max}_{z\in Q}\left[\left|\frac{\sum_{w'}p(w')\left[\Sigma_{\infty}\left(z,w',x,w\right)\right]}{\sqrt{\Sigma_{\infty}\left(x,w,x,w\right)+\lambda_{x,w}}}Z\right|\right]\mid\mathcal{F}_{\infty}\right]\\
 & \leq & \sqrt{2/\pi}\frac{1}{\sqrt{k_{\lambda}}}\hat \epsilon_{2}
\end{eqnarray*}
where $Z\sim N\left(0,1\right)$. This ends the proof.
\endproof

% So, we have proved that given $w\in W$ almost surely for all $\epsilon>0$, there exists $\delta_{0}>0$
% such that if the policy $\pi$ measures an infinite number of alternatives
% in $B_{\delta}\left(x\right)\times \left\{ w\right\}$ for $\delta<\delta_{0}$, then 
% \[
% \left|\Sigma_{\infty}\left(x,w,x,w\right)\right|<\epsilon.
% \]

% Since $Q$ and $W$ are finite, there exists $\delta>0$ such that, $\left|\sum_{w'}p(w')\left[\Sigma_{\infty}\left(z,w',x,w\right)\right]\right|<\epsilon$
% for all $z\in Q$ whenever $\pi$ measures an infinite number of alternatives
% in $B_{\delta}\left(x\right)\times \left\{ w\right\}$.

\begin{lemma}
\label{important}

Suppose that $|W|<\infty$. Let $Q$ be a finite set of $A$. If
$R_{Q}^{N-1}\left(S_{Q}^{\infty};x,w\right)=V_{Q}^{N}\left(S_{Q}^{\infty}\right)$ for all $(x,w)$ a.s.,
then $V_{Q}^{N}\left(S_{Q}^{\infty}\right)=U_{Q}\left(S_{Q}^{\infty}\right)$ a.s.

\end{lemma}

\proof{}
It is essentially the same proof than the proof of \cref{important_lemma}.

\endproof

\putbib
\end{bibunit}

%%%%%%%%%%%%%%%%%
\end{document}